\documentclass[review]{elsarticle}
\usepackage{lineno,hyperref}
\usepackage{amsmath,amssymb}
\usepackage{url}
\usepackage{subcaption}
\usepackage{graphicx}
\usepackage{array,multirow} 
\usepackage{caption}
\usepackage[linesnumbered,ruled]{algorithm2e}
\usepackage{enumitem}
\usepackage{ragged2e}
\usepackage{color}
\usepackage{rotating}
\usepackage[autostyle]{csquotes}
\biboptions{numbers,sort&compress}

\setlist[enumerate]{label*=\arabic*.}

\modulolinenumbers[5]
\usepackage{array}
\newcolumntype{L}{>{\centering\arraybackslash}m{3cm}}
\newcolumntype{P}[1]{>{\centering\arraybackslash}p{#1}}
\makeatletter
\def\@author#1{\g@addto@macro\elsauthors{\normalsize%
    \def\baselinestretch{1}%
    \upshape\authorsep#1\unskip\textsuperscript{%
      \ifx\@fnmark\@empty\else\unskip\sep\@fnmark\let\sep=,\fi
      \ifx\@corref\@empty\else\unskip\sep\@corref\let\sep=,\fi
      }%
    \def\authorsep{\unskip,\space}%
    \global\let\@fnmark\@empty
    \global\let\@corref\@empty  
    \global\let\sep\@empty}%
    \@eadauthor={#1}
}
\makeatother
\journal{Neurocomputing}
\DeclareMathOperator*{\argmax}{argmax}
\begin{document}

\begin{frontmatter}

\title{Robust compressive tracking via online weighted multiple instance learning
	\tnoteref{mytitlenote}}
\author{Sandeep Singh Sengar}
\ead{sandeep.iitdhanbad@gmail.com}


\begin{abstract}
Developing a robust object tracker is a challenging task due to factors such as occlusion, motion blur, fast motion, illumination variations, rotation, background clutter, low resolution and deformation across the frames.  In the literature, lots of good approaches based on the sparse representation have already been presented to tackle the above problems. However, most of the algorithms do not focus on the learning of sparse representation. They only consider the modeling of target appearance and therefore drift away from the target with the imprecise training samples.  By considering all the above factors in mind, we have proposed a visual object tracking algorithm by integrating coarse-to-fine search strategy based sparse representation and the weighted multiple instance learning (WMIL) algorithm. Compared with the other trackers, our approach has more information of the original signal with less complexity due to the coarse-to-fine search method, and also has weights for important samples. Thus, it can easily discriminate the background features from the foreground. Furthermore, we have also selected the samples from the un-occluded sub-regions to efficiently develop the strong classifier. As a consequence, a stable and robust object tracker is achieved to tackle all the aforementioned problems. Experimental results with quantitative as well as qualitative analysis on challenging benchmark \textcolor{blue}{datasets} show the accuracy and efficiency of our method. 
\end{abstract}
\begin{keyword}
	\texttt Object tracking \sep multiple instance learning \sep compressive sensing \sep coarse-to-fine strategy \sep sparse representation 
\end{keyword}

\end{frontmatter}
\section{Introduction}
Visual tracking has remained an active research topic in the computer vision community as it is widely applied in the automatic object identification, automated surveillance, vehicle navigation and many others. Despite great progress in last two decades~\cite{Babenko,Kalal,Sengar,Bai,ZhangB,Rait}, many challenging problems still remain when designing a practical visual tracking system. For example, background clutter, rotation of object, fast motion, illumination changes and occlusions all may cause serious stability issues for a visual tracker~\cite{Smeulders,Lu,Zhong,sengar2017foreground,ZhangK,sengar2017moving,WuY}. Object appearance model, motion model and search strategy are the three main components of a tracking method among these robust object trackers can be designed by giving much attention for effective appearance model and search strategy~\cite{Babenko,Kalal}. Based on different appearance models, tracking methods can be categorized as generative~\cite{ZhouX} or discriminative~\cite{ChenY}. Generative approaches first design an appearance model to represent the target. Subsequently, tracking task based on integral histogram~\cite{AdanA}, template matching~\cite{OranS,LiuX}, incremental subspace learning~\cite{LiuR}, sparse representation~\cite{FengP,ZhouX1}, visual tracking decomposition~\cite{KwonJ} etc. is formulated to find the target appearance with minimal reconstruction error. An offline subspace model is learned by Black and Jepson~\cite{BlackM} to represent the object of interest.  However, it is difficult to adapt the appearance variations in this model.  Furthermore, online expectation maximization and principle component analysis are used by Jepson~\cite{JepsonA} and IVT method~\cite{RossD} respectively to deal with appearance variations. In~\cite{MeiX}, sparse representation is used for object tracking where an object is shown by a trivial templates and sparse linear combination of target. However, it has a problem of optimization and high processing time. To efficiently solve the optimization problem in~\cite{MeiX}, the orthogonal matching pursuit algorithm is adapted by Li et al.~\cite{LiH}. To improve the performance of~\cite{MeiX} in real time, accelerated proximal gradient (APG) approach is used by Bao et al.~\cite{BaoC}. With the help of multiple dynamic and observation models, particle filtering framework is extended by Kwon and Lee~\cite{KwonJ} to account for appearance variations due to scale, partial occlusion, and illumination as well as pose variations. 

\par Discriminative approaches address the tracking as a binary classification task which aims to discriminate the target from the background. These approaches are also known as tracking-by-detection approaches which take tracking as a detection task. Among such methods, the optical flow tracker and the SVM classifier in a Support Vector Tracking (SVT) mechanism are integrated by Avidan~\cite{AvidanS}. A tracking method based on the online multiple instance learning (MIL) method is proposed by Babenko et al.~\cite{BabenkoB1} to treat ambiguous negative and positive samples into bags for learning a discriminative classifier. Babenko et al.~\cite{Babenko} proposed one more method based on MIL to update the appearance model using a set of image patches. On-line feature ranking mechanism is given by Collins et al.~\cite{CollinsR} to select the top-ranked distribution features for separating the target from the background. On-line semi-supervised boosting method is used by Grabner et al.~\cite{GrabnerH}  to solve the drift problem in tracking applications by combining the decision of a given prior and an on-line classifier. Boosting and mean-shift techniques are used in~\cite{AvidanS1} to train a strong classifier and to find the location of the target respectively. Yao et al.~\cite{YaoR} proposed a model for weighted online learning with the help of weighted reservoir sampling for tracking. Furthermore, lots of approaches that take advantage of both generative and discriminative models have been presented~\cite{WangQ,ZhangS}.
\par Sparse random projection based dimensionality reduction approach is employed for target representation in object tracking~\cite{ZhangK1,ZhangK2,WuY1}. Based on the sparse representation based compressive sensing ideas, Zhang et al.~\cite{ZhangK1} proposed a compressive tracking (CT) framework. Here the Haar-like generated features in the compressed domain are classified via a naive Bayes classifier using online update. Real time compressive tracking~\cite{ZhangK1} and the coarse-to-fine search strategy based fast compressive tracking (FCT) algorithms~\cite{ZhangK2} are proposed for visual object tracking, these are good for real-time implementation due to the high processing speed. \textcolor{blue}{Furthermore, Both the CT and FCT approaches consider all the features for the classifier update procedure. If, some parts of the target region are occluded, then this part may not be clearly visible and extracted features from these parts are no more reliable. Consequently, the results obtained from both the above approaches will not be accurate and target will drift away from the original position. For that Yan et al.~\cite{YanJ} proposed a hybrid method of visual object tracking by integrating both the CT~\cite{ZhangK1} and the MIL~\cite{BabenkoB1} approaches. However, this method did not consider the concept of \textit{importance of sample} in its learning process; and again due to the less important positive samples, tracking results will not be reliable in some challenging situations such as illumination variations, rotation, deformation, and background clutter etc.} Wu et al.~\cite{WuY1} and Teng et al.~\cite{TengF} presented a multi-scale tracking based on compressive sensing (MSCT) and multi-scale tracking method via random projections (MSRP) respectively to reduce the effect of target appearance change problems in~\cite{ZhangK1,ZhangK2}, here rapid fern-based features have been employed instead of the Haar-like features. Sengar et al.~\cite{Wispnet} proposed a method for object tracking using Laplacian-DCT based perceptual hash. Here, features in the form of binary hash are extracted and compared with the features of successive frames to detect the similarity. Two random measurement matrices are used in~\cite{TengF1} to extract the complementary features, and the classifier is updated using an adaptive weighting approach for favoring the best features. A semi-supervised compressive coding scheme is proposed by Chen et al.~\cite{ChenS} for online sample labeling.
Here Fisher discrimination criterion and weighted random projection are employed for adaptive compressive sensing for appearance modeling to assess the discrimination capability of randomly generated feature.

\par Motivated by the work in~\cite{YanJ}, in this paper an effective and efficient object tracking method is proposed. Our technique is dependent on the fast compressive tracking for sparse measurement matrix and Haar-like  features to represent the target and online weighted multiple instance strategy to learn the appearance model and to select the high confidence features from it. Furthermore, to reduce the problems of tracking drift and to enhance the accuracy of results, we have incorporated the following ideas (i)  random selection of Haar-like rectangular features from the un-occluded sub-regions to effectively represent the appearance model for target region (ii) size of randomly generated rectangular features are constrained by some parameters for not selecting the features with high dissimilarity (iii) coarse-to-fine search strategy is adopted to reduce the computational complexity (iv) weighted  scheme is employed  to give the importance for positive samples. Extensive experimental results with quantitative and qualitative evaluations on challenging benchmark sequences show the superiority of our tracker to the high performing recent approaches as well as other algorithms in terms of efficiency, stability, accuracy and processing speed. 
\par 
After this introductory section, the rest of this paper is organized as follows. Brief summary of sparse representation based compressive and fast compressive tracer as well as weighted multiple instance learning approach are given in Section~\ref{related}.  It is followed by the proposed work and its implementational  details in Section~\ref{proposed}.  Section~\ref{difference} describes the difference of our approach with the related works. Experimental results with detailed quantitative and qualitative evaluations are presented in Section~\ref{experiment}.  Finally, Section~\ref{conclusion} concludes our work.

\section{Related work}
\label{related}
The main focus of this article is to increase the accuracy of the popular real-time object tracking approaches, namely the CT, its extension FCT, and the weighted multi instance learning.  Review of these approaches are given below which is the base tracker in \textcolor{blue}{the} proposed method.  As known, the FCT is an extended version of the CT, thus prior to presenting the proposed method we will first illustrate the CT and its extension FCT for a better understanding. 

\subsection{Compressive tracker}
Random projection based CT is the most popular approach and is related to \textcolor{blue}{our} method. There are following steps in this approach:
\subsubsection{Sparse representation}

\begin{figure}[h!]
	\centering
	\includegraphics[width=1\linewidth]{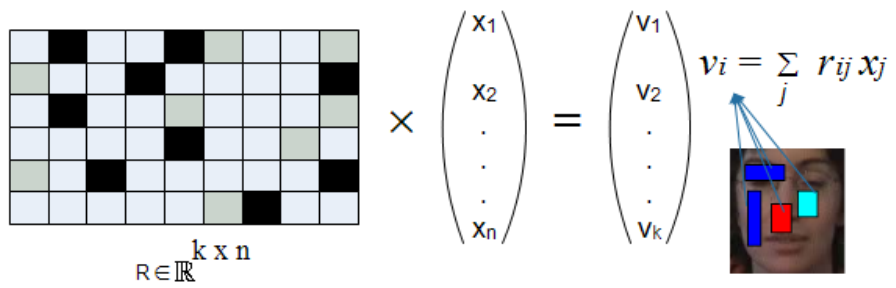}
	\caption{Sparse representation approach used by compressive tracking~\cite{ZhangK1}.}
	\label{sparse0}
\end{figure}

A highly sparse random measurement matrix and random projection are two key concepts for the CT. The relation between them can be shown using following equation:
\begin{equation}
v=Rx
\label{V}
\end{equation}
Here the random projection matrix $R\epsilon \mathbb{R}^{k\times n}$ is used to project the data from high dimensional space $(x \epsilon\mathbb{R}^n)$ to a lower dimensional subspace $(v \epsilon\mathbb{R}^k)$ and $k\ll n$. A restricted isometry  property (RIP) in compression sensing theory is satisfied by Johnson-Lindenstrauss lemma (JL)~\cite{AchlioptasD}, and assumes that random projection matrix \textit{R} satisfies the JL lemma. Thus \textit{x} can be reconstructed with minimum error. A sparse random matrix is adopted to save storage space and enhance the computational efficiency, which generates an embedding effect analogous to the conventional Gaussian random project matrix. The elements of matrix $R$ can be generated as:

\begin{equation}
r_{ij}=\sqrt{\rho}\begin{cases}
-1, & \text{with probability $\frac{1}{2\rho}$} \\
0, & \text{with  probability $1-\frac{1}{\rho}$} \\
1, & \text{with probability $\frac{1}{2\rho}$} 
\end{cases}
\label{rij}
\end{equation}
Here $\rho=n/4$, each row of $R$ approximately contains 2-4 nonzero entries to provide a sparse random measurement matrix, therefore, the total number of non-zero entries in $R$ is less than $4k$. Now Eq.~\ref{V} can be formulated as \textcolor{blue}{$v_i$=$\sum_{j=1}^{N_r}r_{ij}x_{j}$}, here \textcolor{blue}{\textit{x}} is the randomly generated block in the target samples or the background regions, and $N_r$ is the total number of blocks. The procedure for sparse representation is shown in Fig.~\ref{sparse0}.
\subsubsection{Classification}
\label{classification}
Naive Bayes classifier is used by Zhang et al.~\cite{ZhangK1} to find the target candidate region which has the highest confidence.  Here sample labels are in  binary form, $y\epsilon (0,1)$ and naive Bayes classifier is used with the assumption of uniform prior, $p(y=1)=p(y=0)$. the classifier confidence H(v) can be expressed below:
\begin{equation*}
H(v)=log\Bigg ( \frac{\prod_{i=1}^{k} p(v_i|y=1)p(y=1)}{\prod_{i=1}^{k} p(v_i|y=0)p(y=0)} \Bigg) \end{equation*}
\begin{equation}
=\sum_{i=1}^{k}log \Bigg( \frac{p(v_i|y=1)}{p(v_i|y=0)} \Bigg)
\label{Hv}
\end{equation}
Here the conditional distributions, $(p(v_i|y=1)$ and $p(v_i|y=0))$, of the classifier H(v) are Gaussian distribution with parameters $(\mu_i^0,\sigma_i^0,\mu_i^1,\sigma_i^1)$. where $p(v_i|y=k)\approx N(\mu_i^k,\sigma_i^k)$ and $\mu$, $\sigma$ are the mean and standard deviation respectively. The Gaussian distribution parameters are updated with the help of learning parameter $\lambda$ using following equations:
\begin{gather}
\label{update}
\mu_i^0 \leftarrow \lambda \mu_i^0+ (1-\lambda)\mu^0  \\
\sigma_i^0 \leftarrow \sqrt{\lambda(\sigma_i^0)^2+(1-\lambda)(\sigma^0)^2+\lambda(1-\lambda)(\mu_i^0-\mu^0)^2}
\label{update1}
\end{gather}
similar equations for $\mu_i^1$ and $\sigma_i^1$ can be defined. 

\subsection{Fast compressive tracker}
The FCT approach~\cite{ZhangK2} is an improvement over the CT method in terms of speed and tracking accuracy. The general process of the FCT is similar to that of CT, except the modification in terms of search (sampling) techniques used. The FCT employed a coarse to fine search strategies, here there are two stages for sampling process (i) first, the aforementioned classification technique (Section~\ref{classification}) is employed with the coarse sampling procedure by sliding the window with a large number of pixels $\Omega_c$ and the search radius $r_c$ to predict an approximate target location inside the rectangular region centered around the preceding target location (ii) in the next step of fine sampling, rectangular region starting from the location predicted after the coarse sampling is employed with a sliding window of narrow radius $r_f$  in a single pixel steps $\Omega_f$. This approach is much better than the CT in terms of fast detection of target location. However, it leads to the problem of drifting due to random selection of appearance features from the target regions.
\subsection{Online weighted multiple instance learning}
\label{WMIL}
To solve the positive samples uncertainty, we take the target location and appearance representation learning with the help of online weighted multiple instance learning~\cite{ZhangK}. Here we will demonstrate the main lines of WMIL for the sake of completeness. In the WMIL framework, we assume that there are $W_h$ positive samples $\{x_{1j}, j=0,...,W_h-1\}$ and $B_l$ negative samples $\{x_{0j},j=W_h,...,W_h+B_l-1\}$. Positive and negative samples are kept into two bags $\{X^+, X^-\}$ and as like in the MIL tracker~\cite{Babenko}, WMIL also consider that the instance label is similar to the bag label and if there is at least one positive instance in the bag then it is labeled as positive, otherwise negative. When a new sample is arrived, the bag label $y_i$(0 or 1) of the sample is used because of sample labels $y_{ij}$ are not available and all the $M$ features $\phi=\{h_1,h_2,...,h_M\}$ are updated in parallel. Then greedily chooses $K$ most discriminative features $h_K$ from the features pool $\phi$  as defined below: 

\begin{equation}
h_K=\argmax_{h\epsilon\phi}L(H_{K-1}+h) 
\label{hk}
\end{equation}
where the bag log-likelihood function ($L$) is 
\begin{equation}
\label{L}
L=\sum_{i=0}^{1} \Big(y_ilog\big(p(y=1|X^+)\big) +(1-y_i)log\big(p(y=0|X^-)\big)\Big)
\end{equation}
and positive bag probability is defined as:
\begin{equation}
p(y=1|X^+)=\sum_{j=0}^{W_h-1}wt_{j0}p(y_1=1|x_{1j})
\label{probability}
\end{equation}
\par Here $H_{K-1}=\sum_{m=1}^{K-1}h_m$ is the strong classifier with the \textit{K-1} selected weak classifiers. Eq.~\ref{probability} weighs the positive instances as per the significance to the bag probability. 
Here weight $wt_{j0}$ can be shown with the help of euclidean distance between the locations of sample $x_{1j}$ and the current frame tracking result  $x_{10}$ as $wt_{j0}=(1/nc)e^{-|F(x_{1j})-F(x_{10})|}$, here nc and $F(\cdot)\epsilon {R^2}$ are the normalization constant and  the location function respectively. \textcolor{blue}{Please refer~\cite{ZhangK} for detail.}  
\begin{figure}[h!]
	\centering
	\includegraphics[width=1\linewidth]{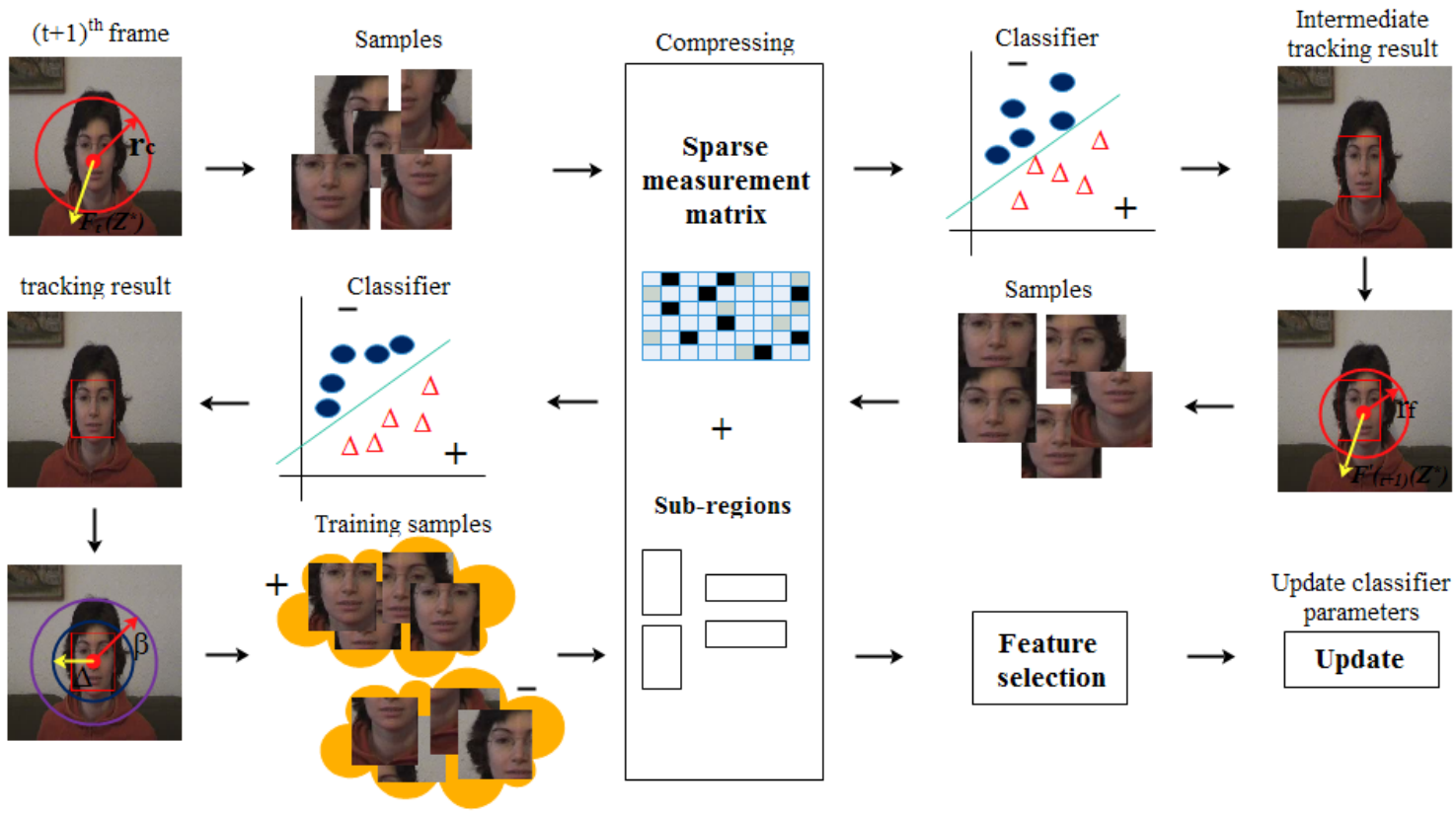}
	\caption{The basic flow of our tracking method.}
	\label{proposed1}
\end{figure}
\section{Proposed work}
\label{proposed}
Our work is an enhancement of the approach proposed by Yan et al.~\cite{YanJ}, where the authors presented an online sparse instance learning (OSIL) based object tracking algorithm, augmenting the compressive tracking algorithm of Zhang et al.~\cite{ZhangK1} with the help of online multiple instance learning framework proposed by Babenko et al.~\cite{BabenkoB1}. In~\cite{YanJ} authors handled the (i) occlusion with the concept of sub-region based feature selection and (ii) the incorrect label sampling problem at the time of appearance model update stage using self learning technique of MIL.   In the proposed approach, we use the same concept of sub-regions based online sparse instance learning. However, different from~\cite{YanJ}, the proposed online weighted multiple instance learning based fast compressive tracking algorithm uses  (i) reliable features selection from the un-occluded randomly generated subregions (Sec.~\ref{sparse1}) (ii) the size of randomly selected rectangular features are constrained by some specific parameters (Sec.~\ref{sparse1}) (iii) coarse-to-fine search strategy based sparse representation technique (Sec.~\ref{learning}) (iv) online weighted multiple instance learning approach to integrate the sample importance into an efficient online sparse instance learning method (Sec.~\ref{learning}).  The basic flow of our tracker is shown in Fig.~\ref{proposed1}.
\subsection{Appearance model based on sparse representation}
\label{sparse1}
It is complicated task to accurately track the object based on data from the previous frame caused by background clutter, rotation, motion blur, varying illumination, occlusion, fast motion and deformation etc. The aforementioned problems lead to tracking drift and error accumulation by delivering incorrect data to the classifier. Therefore, there is a requirement to develop a robust appearance model for dynamic and complex scenes.
\begin{figure}[b!]
	\begin{subfigure}{1\textwidth}
		\centering
		\includegraphics[width=1\linewidth]{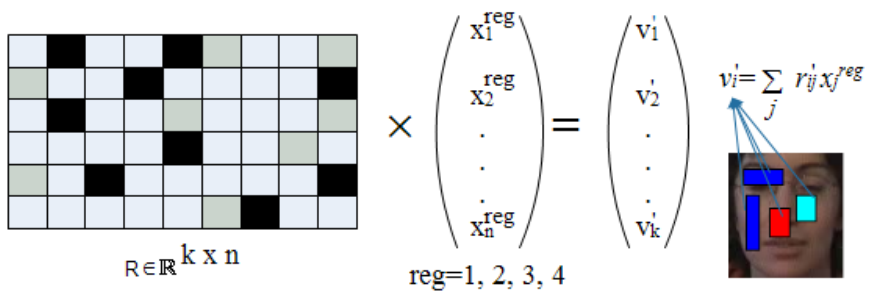}
		\caption{Sparse representation~\cite{YanJ}}
		\label{matrix}
	\end{subfigure} 
	\begin{subfigure}{1\textwidth}
		\vspace{.5cm}
		\centering
		\includegraphics[height=3cm, width=11cm]{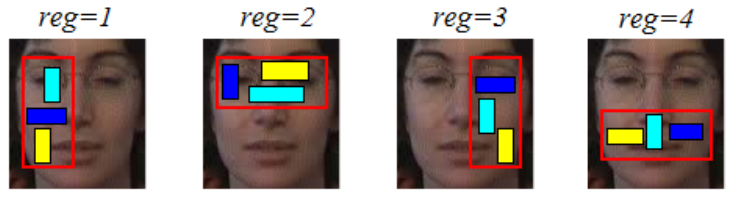}
		\caption{Feature extraction from sub-region}
		\label{region}
	\end{subfigure}
	\caption{Sub-region based compressing features extraction using sparse matrix.}
	\label{sparse}
\end{figure}
\par The texture information (or intensity difference) between the blocks is reflected by the low-dimensional $v_i$ in Fig.~\ref{sparse0} when $\{r_{ij}=-1 \hspace{.1cm}or \hspace{.1cm}0\hspace{.1cm} or\hspace{.1cm} 1\}_j^{N_r}$ and the intensity information of the image sample's appearance is described when $\{r_{ij}=0\hspace{.1cm} or -1\}\}_j^{N_r}$ or $\{r_{ij}=0\hspace{.1cm} or\hspace{.1cm} 1\}\}_j^{N_r}$. In this case $r_{ij}$ is randomly generated using Eq.~\ref{rij} and as given in~\cite{YanJ} the texture information of the appearance model is reflected by almost 70\% of the elements. Moreover, FCT also has same concept of sparse representation as in CT, therefore above calculation is also valid for it.  Furthermore, due to the above computation, FCT extracts and updates the features in the form of rectangular blocks from the entire sample area and weight of each feature is equal. Consequently, FCT will drift or fail at the occurrence of large appearance variations or occlusions. 
\par To solve the above problems, we have used the same concept of sub-regions as given in~\cite{YanJ,ZhuQ} with some additional constraints.  First we will randomly divide the sample regions of  size $W \times H$ into total number of $N_s$ sub-regions of width w and height h using following equation:
\begin{equation}
Pos_i=[rand(1, W-w), rand(1, H-h)]  \hspace{1cm}  i=1 \hspace{.1cm}to \hspace{.1cm} N_s
\label{pos}
\end{equation}
Here $Pos_i$ denotes the upper-left corner position of the sub-region. The appearance variation or occlusion problems will not be accurately solved if we select the large values of \textit{w} or \textit{h}; on the other hand, the sub-region's features will not be stable. Therefore, we experimentally select small \textit{w} and \textit{h} for large target object, otherwise we select large \textit{w} and \textit{h}. In our experimental work, we choose the value of $N_s$ as 4, because smaller value will not be good for tracking results and the complexity will be high with large value of $N_s$.
\par Next, we  use the integrated sparse representation $v'$ \textcolor{blue}{(shown in Fig.~\ref{sparse})}, in place of \textit{v} to preserve the intensity, texture and local spatial features as well as it helps us to attain a better tracking accuracy. The elements $v_i'$ are formulated as follows~\cite{YanJ}:
\begin{equation}
\label{vi1}
v_i'=\sum_{j=1}^{NR}r_{ij}^{'}Recs_{ij}^{reg}
\end{equation}
Here instead of the whole sample area, we extract the Haar-like rectangle features (Recs) from one sub-region. If the size of the rectangular feature is too small then only raw pixel information is captured by it; otherwise there will be a weaker spatial discriminative ability.  Hence, we randomly select the medium size rectangular features and it is constrained as follows:

\begin{gather}
\label{min}
max(w_{min}, \beta _{min}.w)\le width_{rect} \le \beta _{max}.w \\
\label{max}
max(h_{min}, \beta _{min}.h)\le height_{rect} \le \beta _{max}.h
\end{gather}

where $width_{rect}\times height_{rect}$ and $w \times h$ are the size of extracted rectangle feature template and target sample's sub-region respectively. $w_{min}$, $h_{min}$,  $\beta _{min}$, and $\beta _{max}$ are the coefficients set experimentally. The spatial information '$reg$' denotes the $reg^{th}$ sub-region in the image sample. This will help to select the appearance features from the un-occluded sub-regions.
\par The intensity information in the new sparse measurement matrix $R^{'}\epsilon \mathbb{R}^{k\times n}$ is included by considering the probability $p$ and the elements of \{$R'$: $r_{ij}'$\} in Eq.~\ref{vi1} is represented as:
\begin{equation}
\label{rij1}
r_{ij}'=\sqrt{\rho}\begin{cases}
-1, & \text{with probability $\frac{0.22}{\rho}$} \\
0, & \text{with  probability $1-\frac{1}{\rho}$} \\
1, & \text{with probability $\frac{0.78}{\rho}$} 
\end{cases}
\end{equation}

The aim of our work is to extract almost all the important features from the input video sequences at a pre-processing stage for further processing, and here there is no requirement to reconstruct the original video frame from the low dimensional features. In other word we can say that there is no need to satisfy the RIP property  to reconstruct the original signal with minimum error. It is shown in~\cite{YanJ} that both the intensity and texture features are provided with the equal probability $p\approx 0.5$ using Eq.~\ref{rij1} while projecting the sparse representation. So here we can extract the better features in comparison to the features extracted from the Eq.~\ref{rij}.
\subsection{Learning appearance model with online WMIL and FCT}
\label{learning}
To reduce the tracking drift or failure problems in appearance model, due to mis-aligned or noisy sample updated with sparse representation, a robust approach for learning the appearance model with online WMIL and FCT is proposed. Our approach accurately separates the target object from its surrounding background by virtue of its better discriminative performance, closed-form solution and robustness to outliers.
\par At the initial stage of online tracking, we manually find the target object in the first frame of the video sequences. Suppose $F_t$ represents the position of target sample at the $t^{th}$ frame. Then, first we crop some patches $L^{\alpha}=\{Z||F(Z)-F_{t}|<\alpha \}$, within the search radius $\alpha$. Next, these patches are kept into a positive bag $X^+$.  Subsequently, some patches are randomly cropped out from set $L^{\Delta,\beta}= \{Z|\Delta <|F(Z)-F_{t}|<\beta \}$ where  $\alpha < \Delta < \beta$, and put them into a negative bag $X^-$. After that using Eqs.~\ref{vi1}--\ref{rij1}, we compute the sparse representation of each patch of both positive and negative bags to extract the Haar-like features $v_{np}^{'}$ and $v_{nn}^{'}$ respectively. Then, we update the classifier parameters of sparse represented features with the help of Eqs.~\ref{update} and~\ref{update1}. 
\par When the $(t+1)^{th}$ frame comes, some patches are coarsely cropped out $L^{r_c}=\{Z||F_{t+1}(Z)-F_t(Z^*)|<r_c \}$, where  $F_t(Z^*)$ represents the tracking position at frame \textit{t}. Subsequently using Eqs.~\ref{vi1}--\ref{rij1}, we compute the features $v_{r_c}^{'}$ for each sample patch. After that, log ratio of weak classifier $h_k(x)$ (given in Eq.~\ref{H1}) is used to measure the confidence that sample \textit{Z} in each bag would be classified as positive or negative. 
{\SetAlgoNoLine%
	\begin{algorithm}[h!]
		\SetKwInOut{Input}{Input}
		\SetKwInOut{Output}{Output}
		\caption{Tracking via our proposed method}
		\Input{The ${(t+1)}^{th}$ image frame}
		\vspace{-.3cm}
		\begin{enumerate}
			\item Coarser operation
			\vspace{-.3cm}
			\begin{enumerate}
				\item Coarsely cropped out a set of image samples $L^{r_c}=\{Z||F_{t+1}(Z)-F_t(Z^*)|<r_c\}$, where $F_t(Z^*)$ is the tracking position at frame t.
				\vspace{-.3cm}
				
				\item Extract the features $v_{r_c}^{'}$ for each sample using Eqs.~\ref{vi1}--\ref{rij1}
				
				\vspace{-.3cm}
				\item \For {i=1 to K } 
				{ 
					\subitem \hspace{.5cm}	Estimate classifier $H(v_i^{'})$ depending on $v_{r_c}^{'}$ using Eq.~\ref{HVrc1} 
				}
				\vspace{-.3cm}
				\item 
				$F_{t+1}^{'}(Z^*)=\argmax_{Z\epsilon L^{V_{r_c}^{'}}}(H(Z))$

			\end{enumerate}
			\vspace{-.3cm}
			\item Finer operation
			\vspace{-.3cm}
			\begin{enumerate}
				\item Finely cropped out a set of image samples $L^{r_f}=\{Z||F_{t+1}(Z)-F_{t+1}^{'}(Z^*)|<r_f\}$
				\vspace{-.3cm}
				\item Extract the features $v_{r_f}^{'}$ for each sample using Eqs.~\ref{vi1}--\ref{rij1}
				\vspace{-.3cm}
				\item \For {i=1 to K } 
				{ 
					\subitem \hspace{.5cm}	Estimate classifier $H(v_i^{'})$ depending on $v_{r_f}^{'}$ using Eq.~\ref{HVrc2} 
				}
				\vspace{-.3cm}
				\item 
				$F_{t+1}(Z^*)=\argmax_{Z\epsilon L^{v_{r_f}^{'}}}(H(Z))$ 
				
			\end{enumerate}
			\vspace{-.3cm}
			\item Crop positive samples $x^+$ using $L^\alpha=\{Z||F(Z)-F_{t+1}(Z^*)|<\alpha$ and negative samples $x^-$ using $L^{\Delta\beta}=\{Z|\Delta < |F(Z)-F_{t+1}(Z^*)|<\beta\}$, where $\alpha < \Delta < \beta$
			\vspace{-.3cm}
			\item Extract the features $v_{pp}^{'}$ and $v_{nn}^{'}$ corresponding to the $x^{+}$ and $x^-$ respectively using Eqs.~\ref{vi1}--\ref{rij1}.
			\vspace{-.3cm}
			\item Choose \textit{K} selectors from \textit{M} weak classifiers using Eq.~\ref{hk} \linebreak
			\For {$r_f$=1 to K } 
			{ 
				\subitem \hspace{.5cm}	Update the classifier parameters using Eqs.~\ref{update}, \ref{update1}.
			}
		\end{enumerate}
		\vspace{-.3cm}
		\Output {Tracking location $F_{t+1}(Z^*)$, \textit{K} selectors and classifier parameters}
		\label{Algo2}
	\end{algorithm} 
}
\begin{equation}
h_k(x)=log \Bigg ( \frac{P(v_{r_c}^{'} (Z)|y=1)}{P(v_{r_c}^{'} (Z)|y=0)}\Bigg )
\label{H1}
\end{equation}
Where the values of  $P(v_{r_c}^{'} (z)|y=1)$ and $P(v_{r_c}^{'} (z)|y=0)$ can be computed as in Eq.~\ref{Hv} and the bag probability  \textit{P} is estimated by Eq.~\ref{probability}. These bag probability are used to select \textit{K} elements $\{v_{r_c}^{'}\}_{i=1}^K$ from the extracted features with the help of Eq.~\ref{hk}. Here by considering $K<M$ means, the reliable classifiers are only utilized to find the new position of targets. Now the strong classifier (in Eq.~\ref{HVrc1}) depended on $v_{r_c}^{'}$ is applied to the patches cropped from the $(t+1)^{th}$ frame and select the new location of target $F_{t+1}^{'}(Z^*)$ corresponding to the maximum classifier response given in Eq.~\ref{argmax}. 
\begin{equation}
\label{HVrc1}
H(v_{r_c}^{'})=\sum_{i=1}^{K} log \Bigg ( \frac{P(v_{r_c}^{'} (z)|y=1)}{P(v_{r_c}^{'} (z)|y=0)}\Bigg )
\end{equation}

\begin{equation}
\label{argmax}
F_{t+1}^{'}(Z^*)=\argmax_{Z\epsilon L^{r_c}}(H(Z))
\end{equation}
In the next stage some patches are finely cropped out $L^{r_f}=\{Z||F_{t+1}(Z)-F_{t+1}^{'}(Z^*)|<r_f \}$. Now use same operations as above with $L^{r_f}$ in place of $L^{r_c}$, and compute the value of $H(v_{r_f}^{'})$ using following equation: 
\begin{equation}
\label{HVrc2}
H(v_{r_f}^{'})=\sum_{i=1}^{K} log \Bigg ( \frac{P(v_{r_f}^{'} (z)|y=1)}{P(v_{r_f}^{'} (z)|y=0)}\Bigg )
\end{equation}
Now, the final target location $F_{t+1}(Z^*)$ corresponding to the maximum classifier response can be computed using Eq.~\ref{argmax1}.
\begin{equation}
\label{argmax1}
F_{t+1}(Z^*)=\argmax_{Z\epsilon L^{r_f}}(H(Z))
\end{equation}
The above procedures are repeated by our model for succeeding frames. The main steps of our approach are summarized in Algorithm~\ref{Algo2}.
\section{Difference with related works:}
\label{difference}
It should be noted that robustness and stability are the key characteristics of our presented approach and this method is different from some latest works based on sparse representation and appearance model learning like CT~\cite{ZhangK1}, FCT~\cite{ZhangK2}, MIL~\cite{BabenkoB1}, WMIL~\cite{ZhangK}, DWCM~\cite{ChenT}, OSIL~\cite{YanJ} and other state-of-the-art techniques in the following way: the first significant difference in the form of appearance model. In~\cite{ZhangK1,ZhangK2,BabenkoB1,ZhangK,ChenT}, the Haar-like features for appearance model representation are extracted randomly from whole sample region and these features will not be robust enough to track the object in the case of occlusion and appearance variations. Our method handles the aforesaid problems well, by selecting the features from the un-occluded sub-regions. However this concept is used by OSIL~\cite{YanJ}, but different from it, we also provided some additional constraints to effectively extract the features, like, the size of the rectangular features are decided  by some parameters (Eq.~\ref{min} and Eq.~\ref{max}) for not selecting the features with high dissimilarity. The second difference lies in the terms of computation complexity. In OSIL~\cite{YanJ}, compressive tracking (CT) approach is employed to find the samples near the target objects in the current frame. Furthermore we have used the coarse to fine resolution based search strategy in which object location is significantly accurate and total number of search windows is less, thus considerably reducing the computational cost. Finally, the most significant difference is that our scheme employs a weighting concept to favor the best performing samples.  The method proposed by Yan et al.~\cite{YanJ} uses the \textcolor{blue}{multiple} instance learning~\cite{BabenkoB1} techniques to learn, update, and to find the best performing classifier after giving equal significance to all the samples. Therefore it leads to inaccurate tracking results by not considering the sample importance. Furthermore for better performance, the samples near the target object region should provide the more weight in comparison to others. By considering this problem, we have employed the weighting mechanism with the help of online WMIL method~\cite{ZhangK} and get the superior results than OSIL.  
\section{Experimental results and analysis}
\label{experiment}

\textcolor{blue}{To make a fair comparison, we have carried out the experimental evaluation on two benchmark datasets, Object Tracking Benchmark (OTB)~\cite{wu2015object} and Visual Object Tracking (VOT)~\cite{kristan2015visual}.}  Our method has been implemented in MATLAB R2013a environment and executed on Intel (R) core (TM) i7-4770 cpu@3.40GHz processor with 4GB RAM. \textcolor{blue}{In the following subsections, first, we will discuss about the parameter setting for our method in Sec.~\ref{parameter}. Subsequently, an overview of the used benchmark datasets are provided in Sec.~\ref{dataset}. The experimental evaluation based on OTB100 and VOT2015 datasets are given in Sec.~\ref{OTB} and~\ref{VOT} respectively. Finally, qualitative analysis based on different challenging attributes are presented in Sec.~\ref{qualitative}.}
\subsection{Parameters setting}
\label{parameter} 
We have used the fixed parameters for all the used datasets for fair evaluation. Two important parameters, namely positive and negative search radius are based on speed of appearance changes. A large value of search radius $\alpha$ is required to acquire more positive samples, if the object moves fast otherwise, small $\alpha$ value will be good to reduce computing time.  Furthermore, cropped negative samples should contain sufficient discriminative information as well as less overlapping with the positive ones. It was noticed that more constructive results could be achieved with $\alpha=4$ and total number of 50 negative samples generated within the search ranges of $\Delta$=8 and $\beta$=22. We have divided the samples into four sub-regions. A large value of updated parameter $\lambda$ builds the high weight on the old parameters.  Therefore, if the appearance changes slowly, then a large value of $\lambda$ should be selected to maintain the parameters as stable as possible. In our work, the value of $\lambda$ is set to 0.9. The large number of features are extracted if the appearance of the objects changes quickly and we have extracted the \lq\textit{M}=100\rq~Haar-like features. Total number of \textit{K}=20 high confidence features are selected for learning step. For comparison purpose, we have used the same parameters as suggested by the authors of corresponding algorithms. 
\subsection{Datasets}
\label{dataset}
\begin{figure*}
	\hspace{-.25cm}
	\resizebox{1\textwidth}{!}{
		\begin{tabular}{cccccc}
			\vspace{.2cm}
			\begin{subfigure}{0.18\textwidth}\centering\includegraphics[height=2.3cm, width=2.51cm]{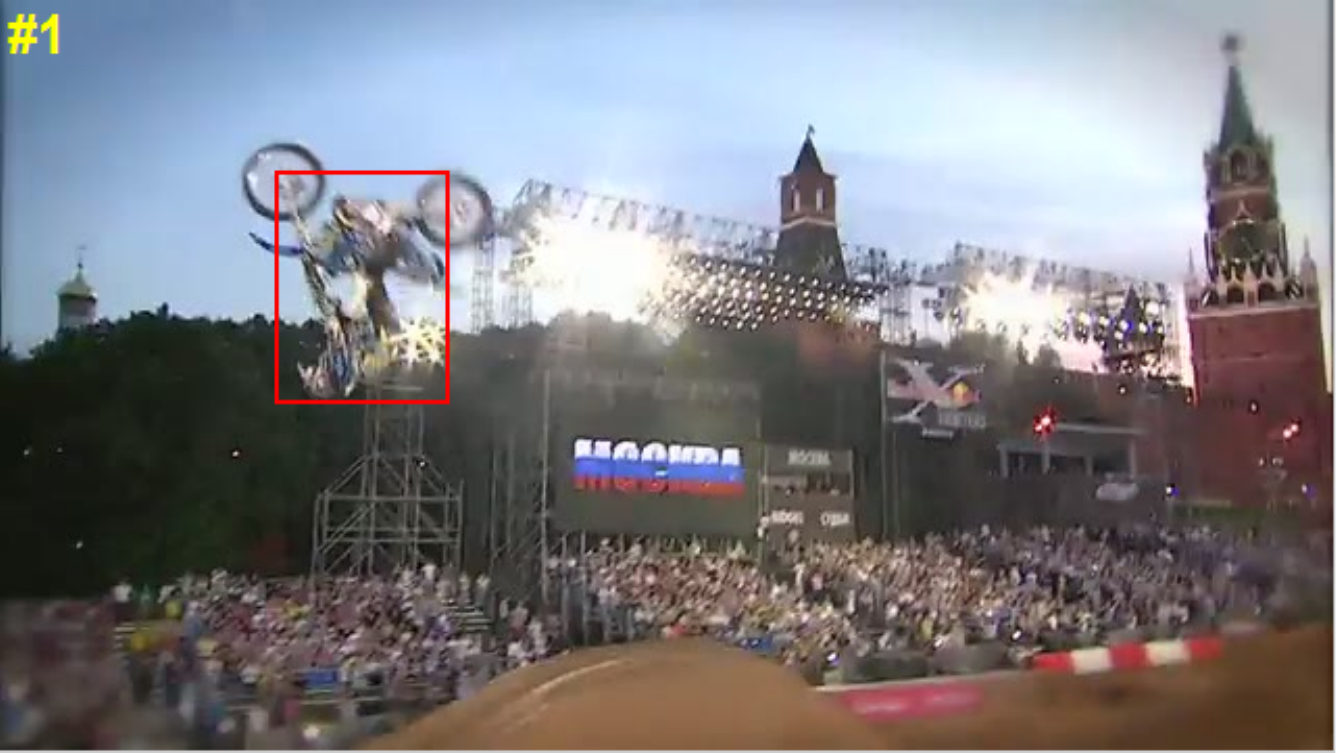}\end{subfigure}&
			\begin{subfigure}{0.18\textwidth}\centering\includegraphics[height=2.3cm, width=2.51cm]{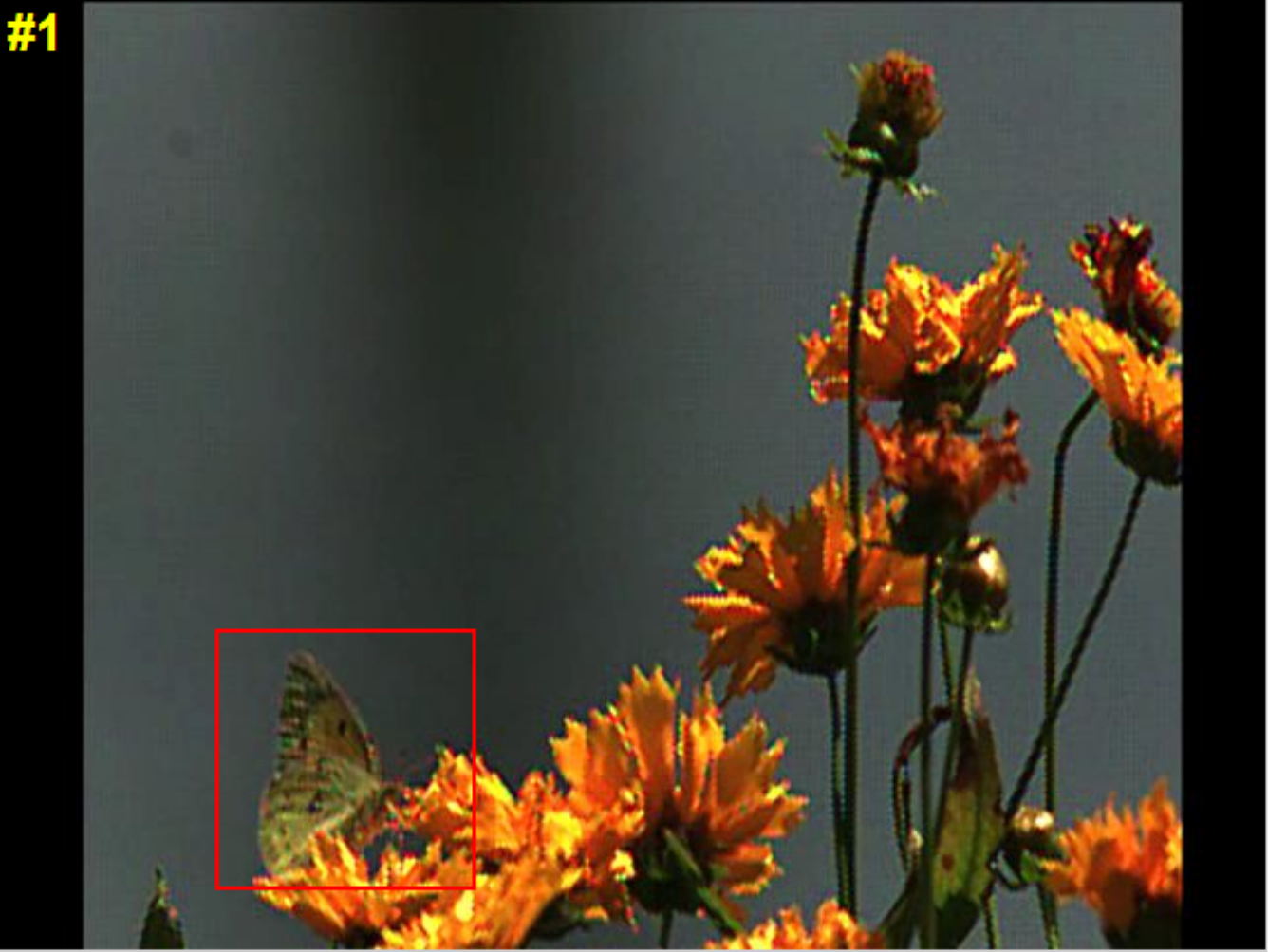}\end{subfigure}&
			\begin{subfigure}{0.18\textwidth}\centering\includegraphics[height=2.3cm, width=2.51cm]{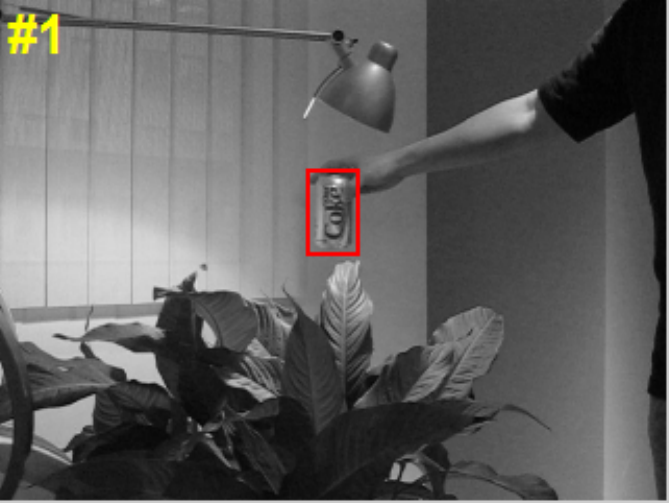}\end{subfigure} &
			\begin{subfigure}{0.18\textwidth}\centering\includegraphics[height=2.3cm, width=2.51cm]{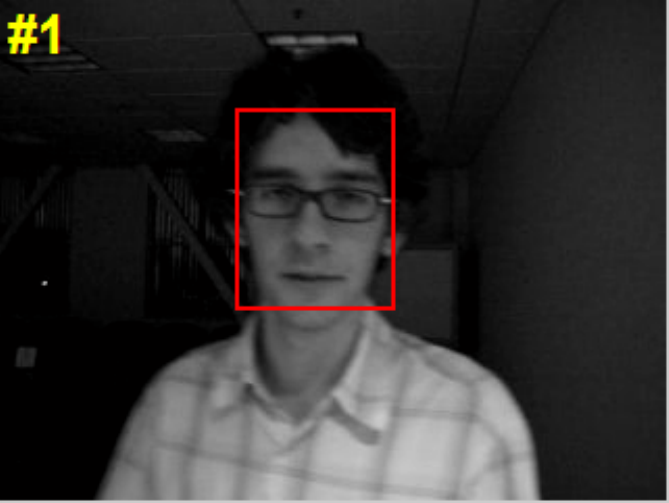}\end{subfigure} &
			\begin{subfigure}{0.18\textwidth}\centering\includegraphics[height=2.3cm, width=2.51cm]{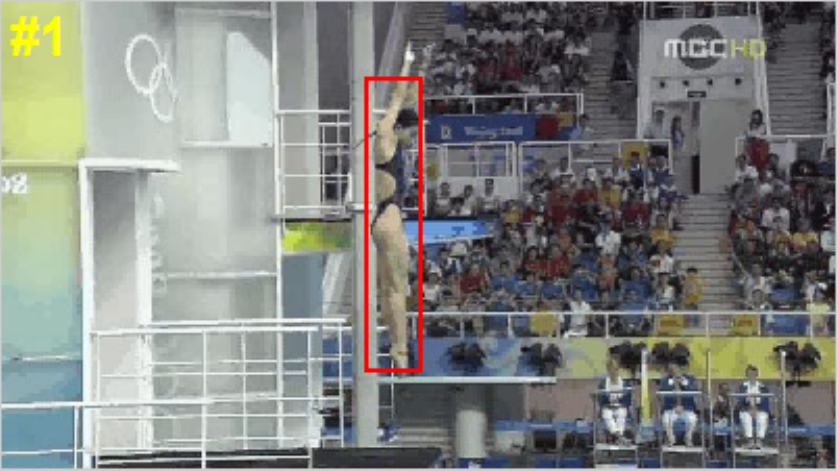}\end{subfigure} &
			\begin{subfigure}{0.18\textwidth}\centering\includegraphics[height=2.3cm, width=2.51cm]{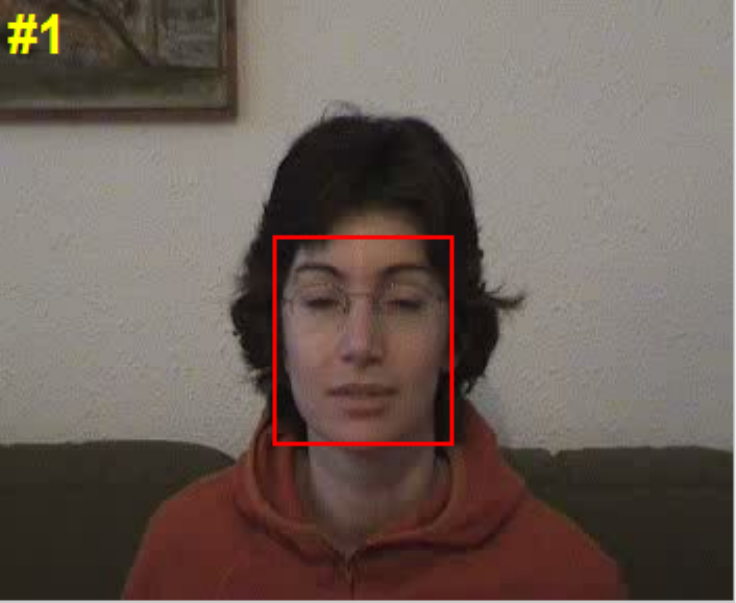}\end{subfigure} 
			\\ \vspace{.2cm}
			\begin{subfigure}{0.18\textwidth}\centering\includegraphics[height=2.3cm, width=2.51cm]{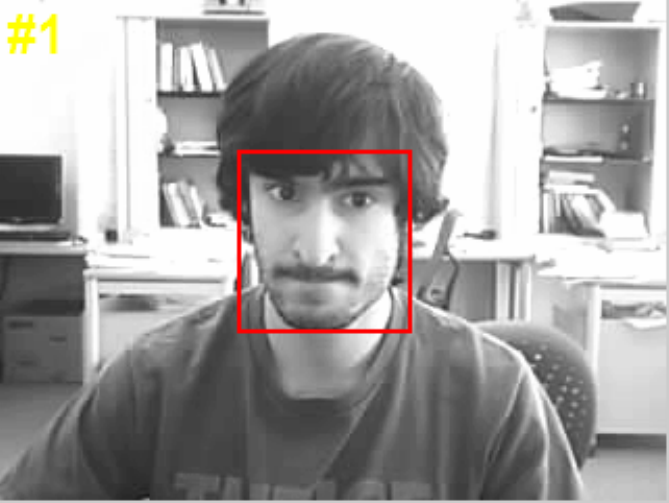}\end{subfigure} &
			\begin{subfigure}{0.18\textwidth}\centering\includegraphics[height=2.3cm, width=2.51cm]{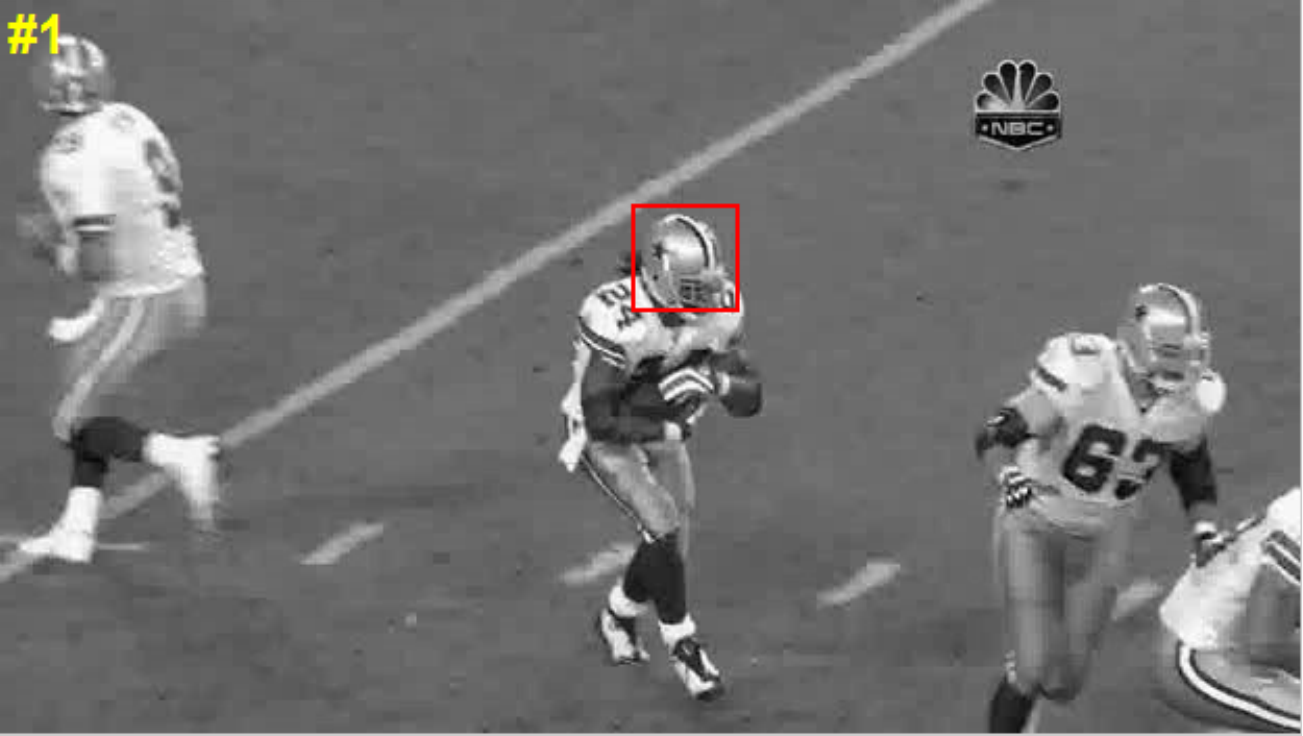}\end{subfigure} &
			\begin{subfigure}{0.18\textwidth}\centering\includegraphics[height=2.3cm, width=2.51cm]{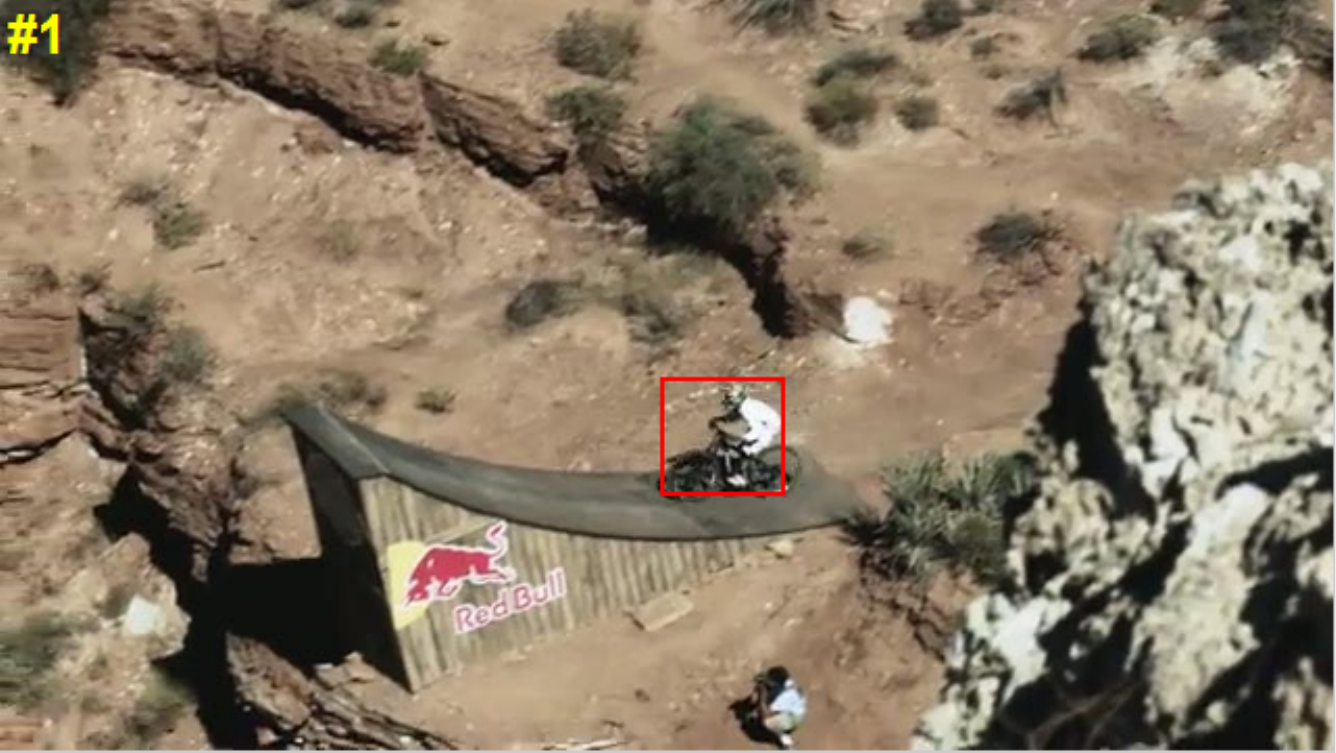}\end{subfigure} &
			\begin{subfigure}{0.18\textwidth}\centering\includegraphics[height=2.3cm, width=2.51cm]{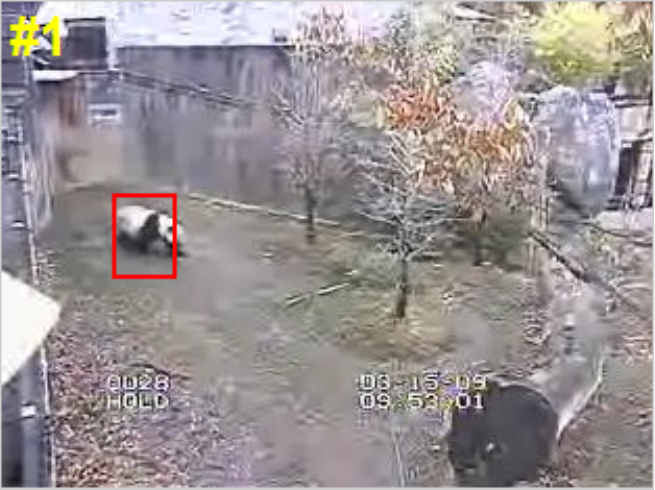}\end{subfigure}&
			\begin{subfigure}{0.18\textwidth}\centering\includegraphics[height=2.3cm, width=2.51cm]{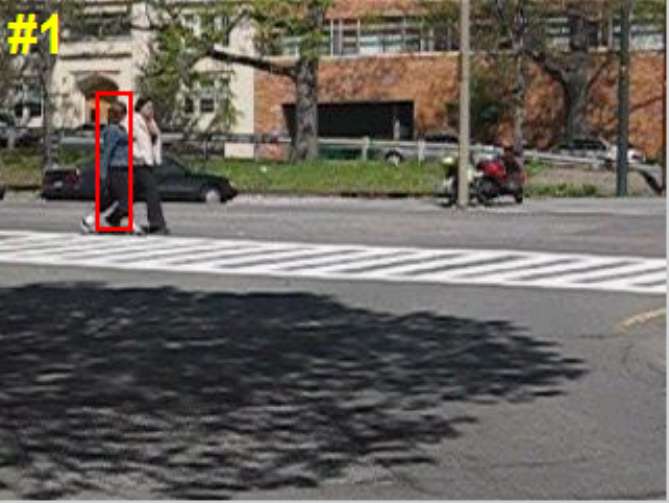}\end{subfigure}&
			\begin{subfigure}{0.18\textwidth}\centering\includegraphics[height=2.3cm, width=2.51cm]{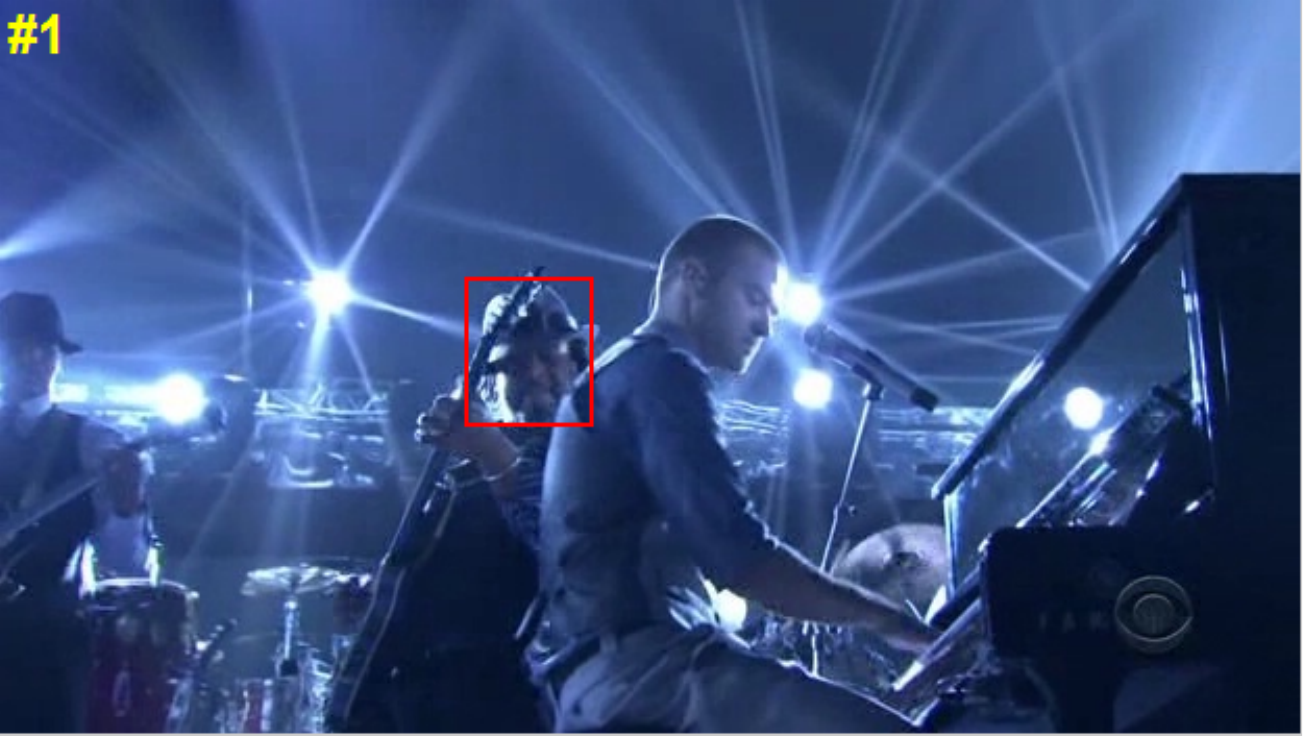}\end{subfigure} 
			\\ \vspace{.2cm}
			\begin{subfigure}{0.18\textwidth}\centering\includegraphics[height=2.3cm, width=2.51cm]{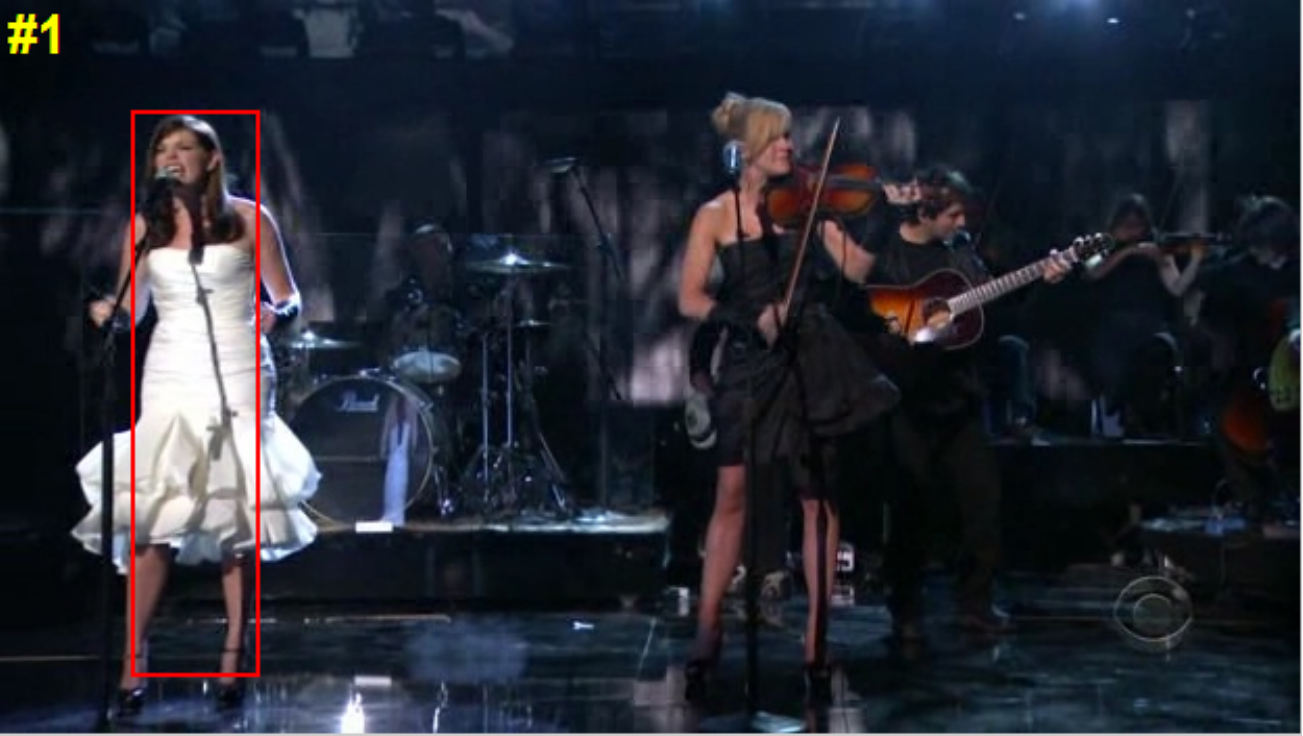}\end{subfigure} &
			\begin{subfigure}{0.18\textwidth}\centering\includegraphics[height=2.3cm, width=2.51cm]{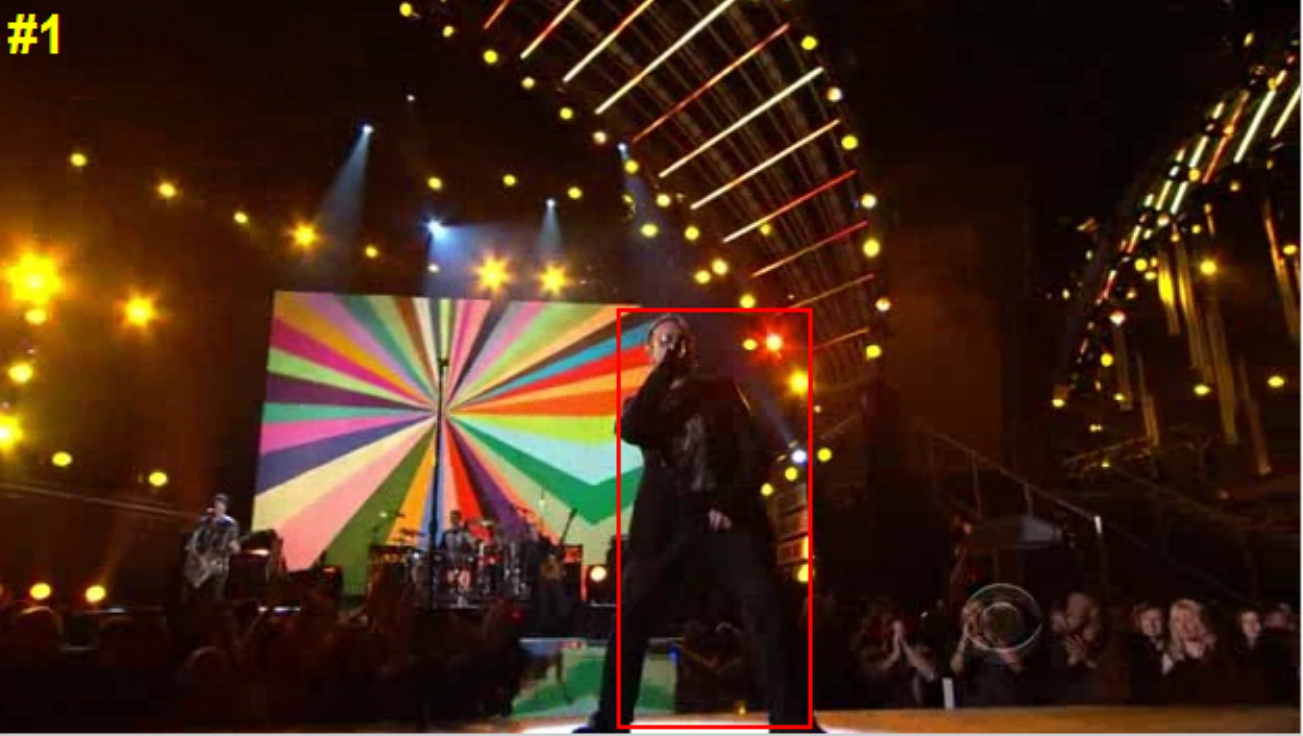}\end{subfigure} &
			\begin{subfigure}{0.18\textwidth}\centering\includegraphics[height=2.3cm, width=2.51cm]{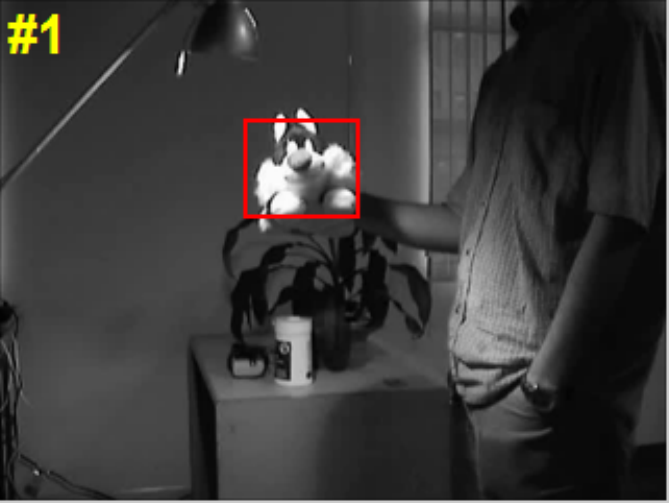}\end{subfigure}&
			\begin{subfigure}{0.18\textwidth}\centering\includegraphics[height=2.3cm, width=2.51cm]{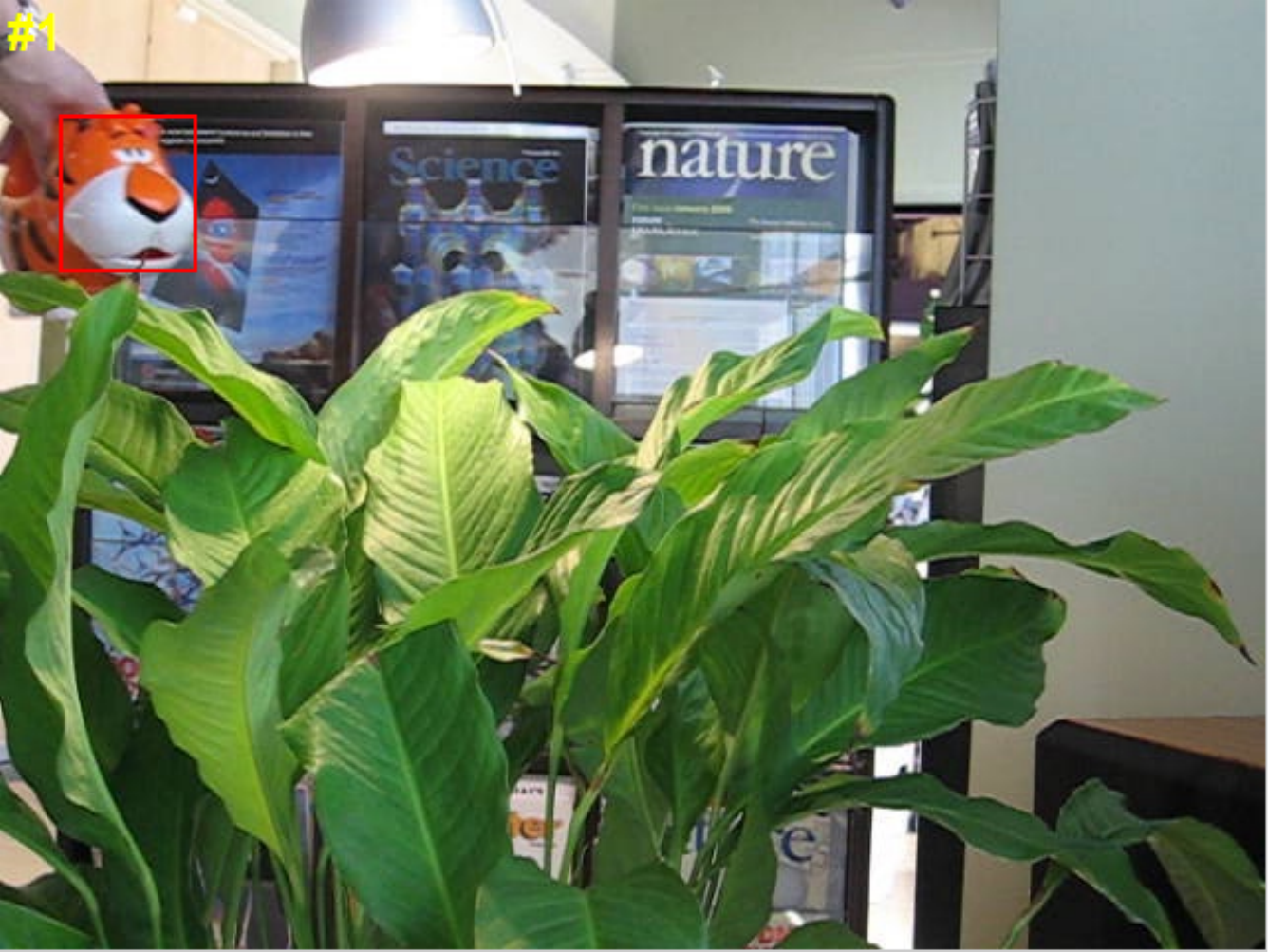}\end{subfigure}&
			\begin{subfigure}{0.18\textwidth}\centering\includegraphics[height=2.3cm, width=2.51cm]{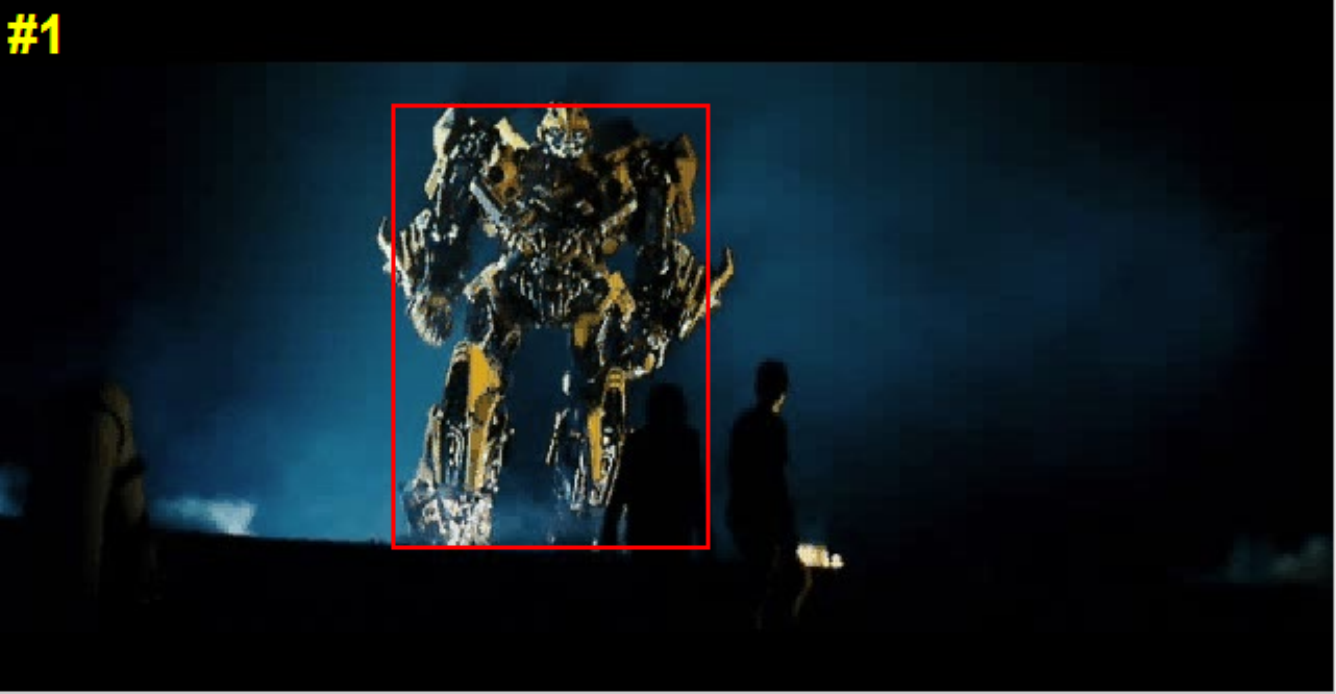}\end{subfigure} &
			\begin{subfigure}{0.18\textwidth}\centering\includegraphics[height=2.3cm, width=2.51cm]{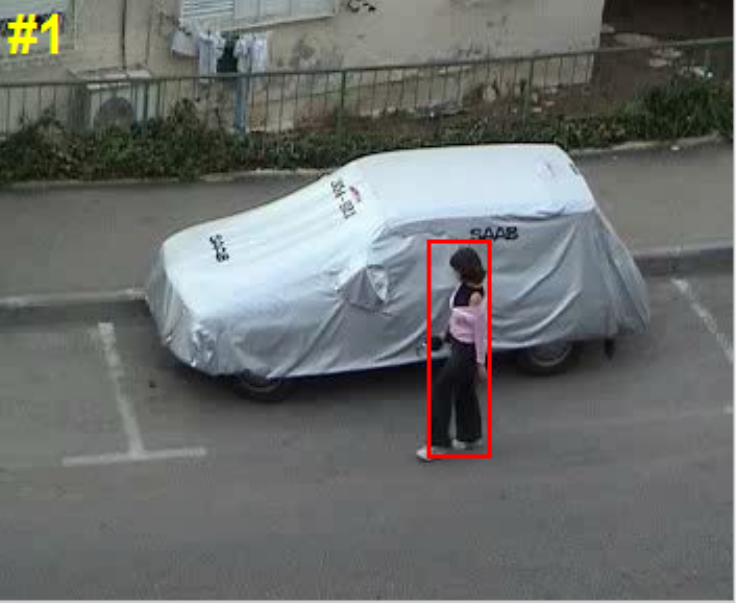}\end{subfigure} 
		\end{tabular}
	}
	\caption{
		\label{ground}\textcolor{blue}{The first frames of \textcolor{blue}{the randomly selected videos of used benchmark datasets} with the ground truth bounding box. Top row, from left to right: Motocross1, Butterfly, Coke1, David, Diving and Occluded face1. Middle row, from left to right: Occluded face2, Football, Mountain-bike, Panda, Pedestrian3 and Shaking. Bottom row, from left to right: Singer1, Singer2, Sylvester, Tiger2, Trans and Woman.}}
	\end{figure*}

\textcolor{blue}{OTB100 and VOT2015 are popular benchmarks, which contain 100 and 60 fully annotated video sequences respectively with complex and challenging environments for tracking. These   datasets are categorized with the following attributes:  background clutters, deformation, fast motion, illumination variation, in-plane rotation, out-of-plane rotation, low resolution, motion blur, and occlusion} to measure the strength and weaknesses of the trackers in a better way.  For fair evaluation, the ground truth~\footnote{\url{http://cvlab.hanyang.ac.kr/tracker_benchmark/datasets.html}},\footnote{\url{http://www.votchallenge.net/vot2015/dataset.html}} in the form of bounding box of these datasets are used to measure the performance of ours and the other existing methods. \textcolor{blue}{Due to the large number of video sequences in both the used benchmarks, the target object in the first frames using ground truth bounding box of some of the randomly chosen videos} are displayed in Fig.~\ref{ground}. 
\textcolor{blue}{
\subsection{Evaluation on OTB}
\label{OTB}
The details about evaluation protocol followed by the comparison with other approaches under   different categories are provided in the following subsections.}
\subsubsection{Evaluation protocol}
\label{protocol}
\textcolor{blue}{As suggested in~\cite{nam2016learning},} we use the success rate score (SR) based on bounding box overlap criteria and the center location error (CLE)~\cite{GaoY} to perform the quantitative evaluation analysis. SR is computed as: 
\begin{equation}
SR=\frac{area(B_T \cap B_G)}{area(B_T \cup B_G)}
\end{equation}
Where $B_G$ and $B_T$ are the ground truth and the tracking bounding box respectively. The notations $\cup$, $\cap$, and area represent the union, intersection, and the number of pixels in the bounding boxes respectively. Here we consider that, if SR is greater than 0.5, then the target is tracked successfully.
The CLE is computed as the Euclidean distance between the central position of the tracker output box ($C_T$) and the ground truth box ($C_G$) respectively. It can be shown as:
\begin{equation}
CLE=||C_T-C_G||
\end{equation}
\begin{figure*}
	\hspace{-.4cm}
	\resizebox{1\textwidth}{!}{
		\begin{tabular}{ccc}
			\vspace{.2cm}
			\begin{subfigure}{0.34\textwidth}\centering\includegraphics[height=4.3cm, width=4.3cm]{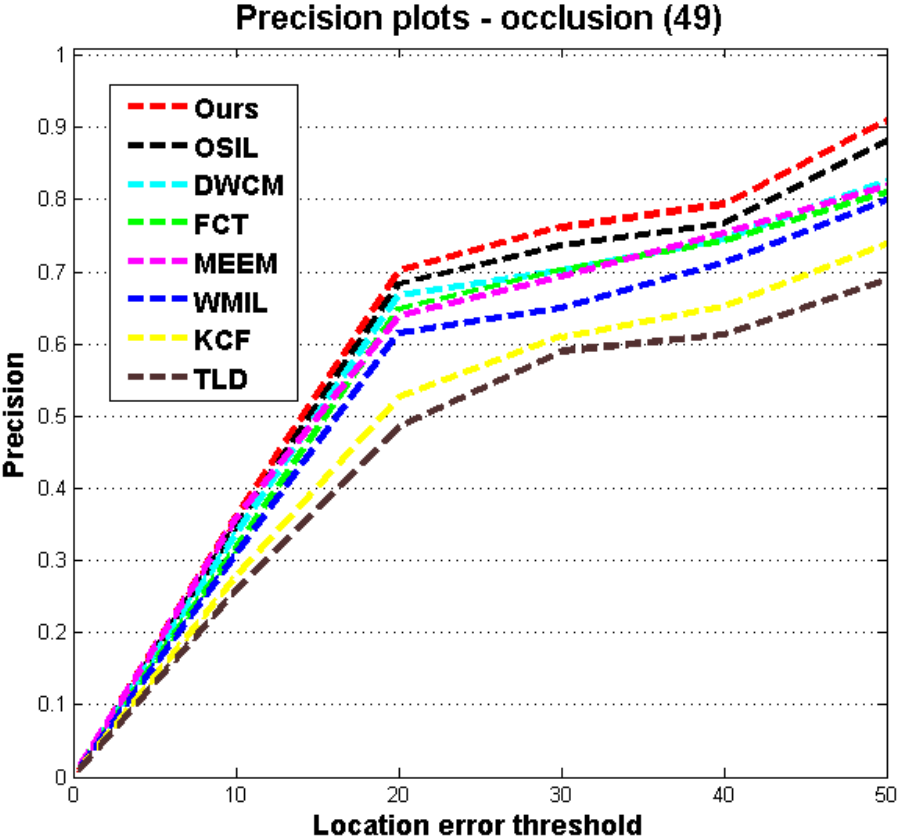}\end{subfigure}&
			\begin{subfigure}{0.34\textwidth}\centering\includegraphics[height=4.3cm, width=4.3cm]{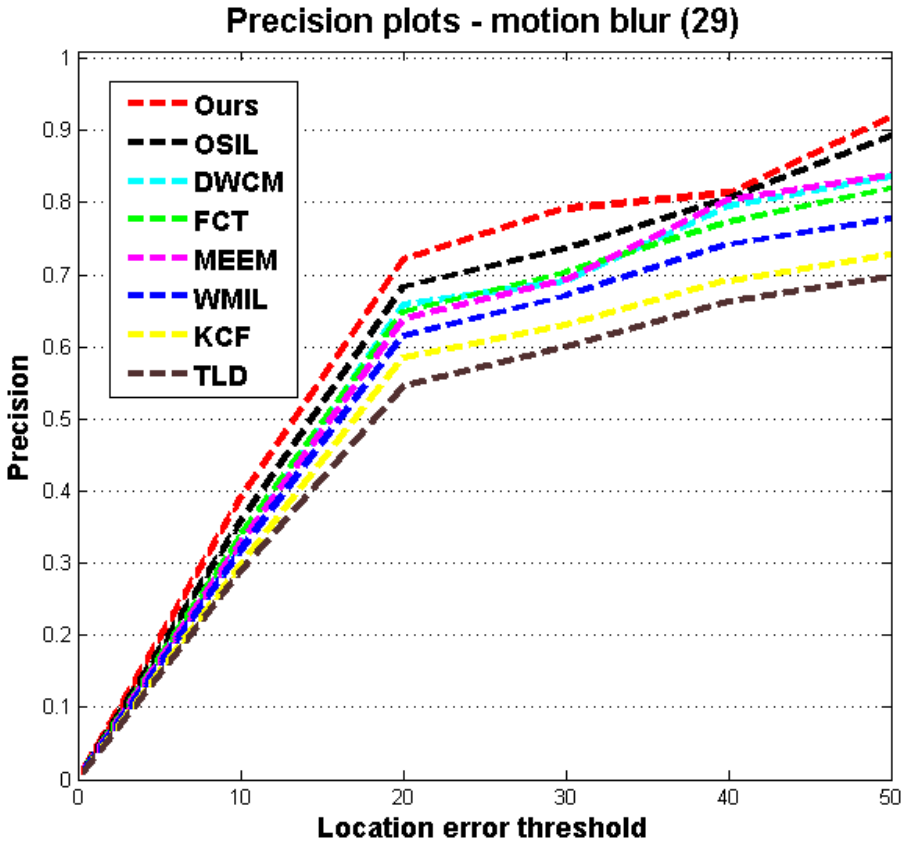}\end{subfigure}&
			\begin{subfigure}{0.34\textwidth}\centering\includegraphics[height=4.3cm, width=4.3cm]{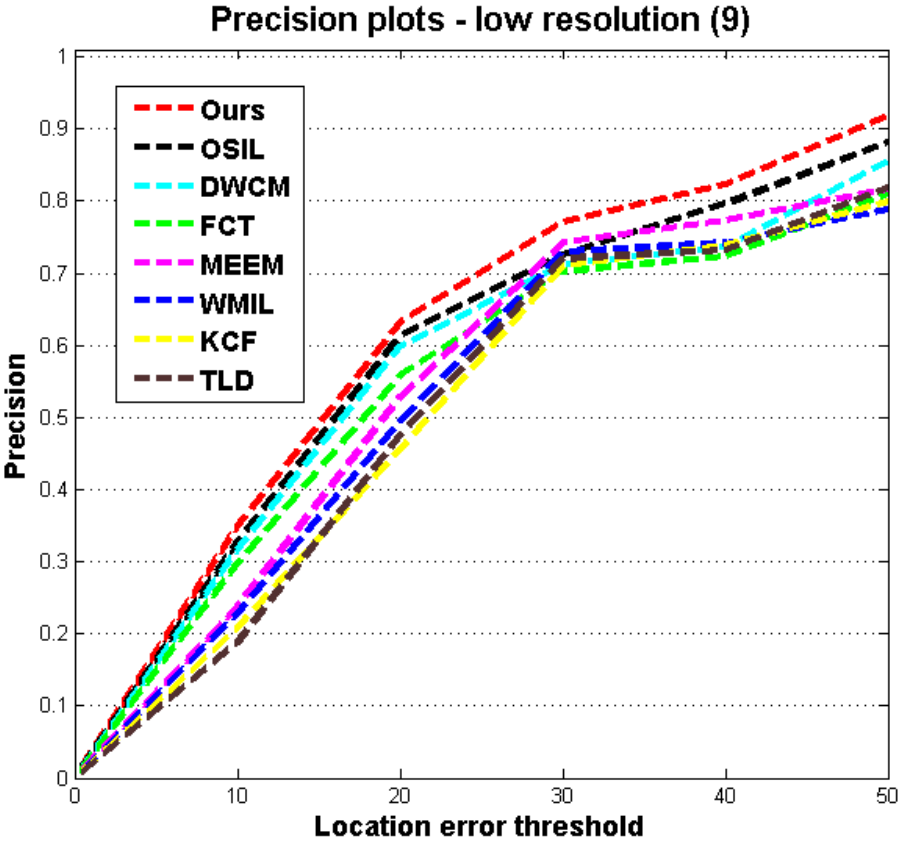}\end{subfigure} \\
			\vspace{.2cm}
			\begin{subfigure}{0.34\textwidth}\centering\includegraphics[height=4.3cm, width=4.3cm]{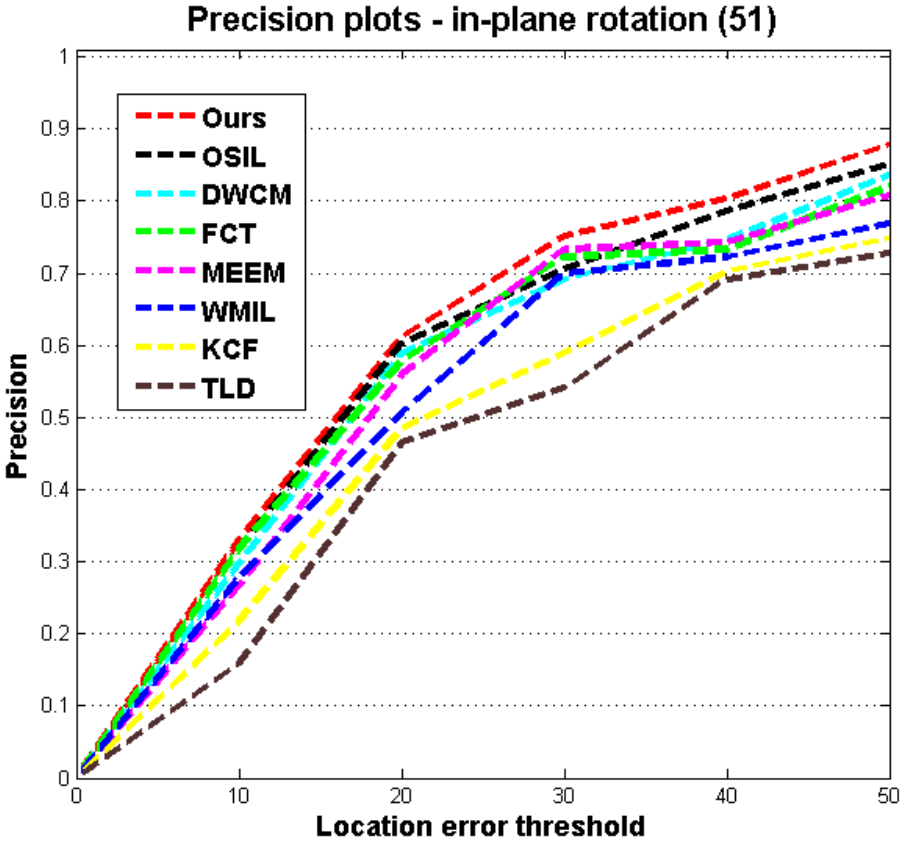}\end{subfigure}&
			\begin{subfigure}{0.34\textwidth}\centering\includegraphics[height=4.3cm, width=4.3cm]{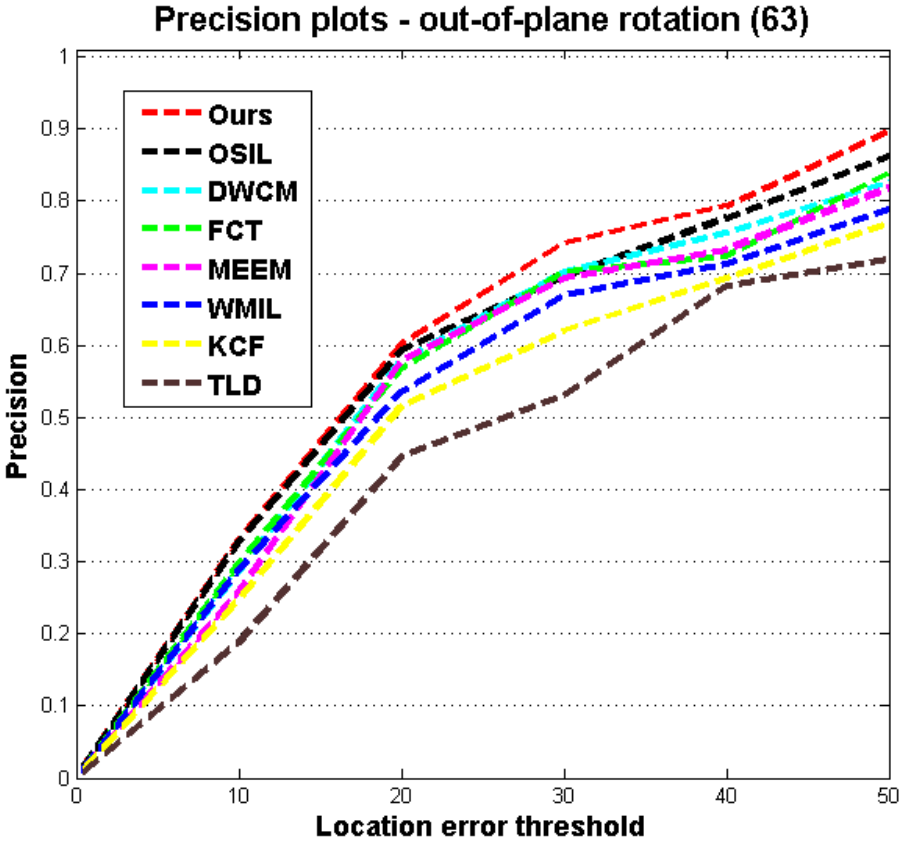}\end{subfigure}&
			\begin{subfigure}{0.34\textwidth}\centering\includegraphics[height=4.3cm, width=4.3cm]{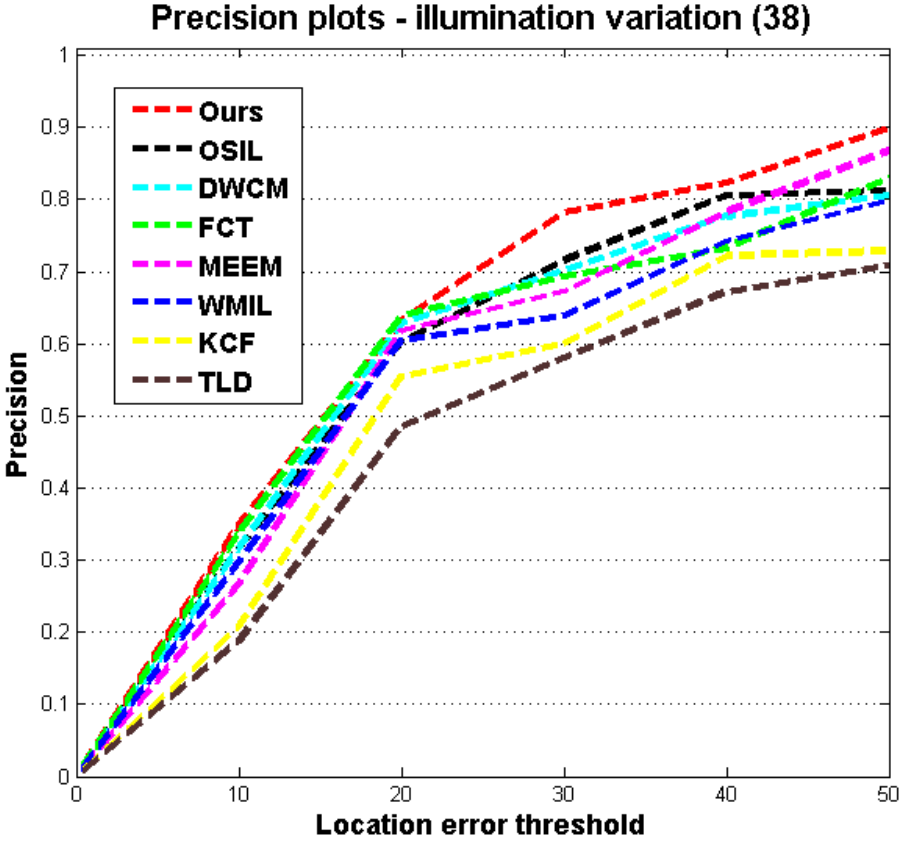}\end{subfigure} 	\\
			\vspace{.2cm}
			\begin{subfigure}{0.34\textwidth}\centering\includegraphics[height=4.3cm, width=4.3cm]{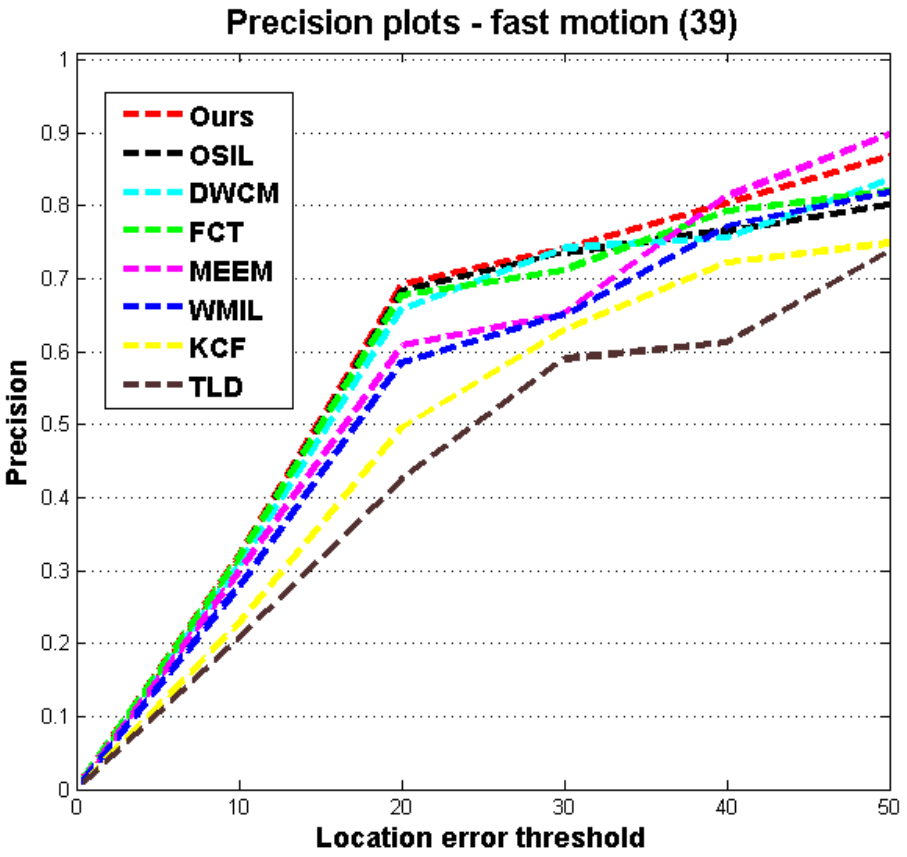}\end{subfigure} &
			\begin{subfigure}{0.34\textwidth}\centering\includegraphics[height=4.3cm, width=4.3cm]{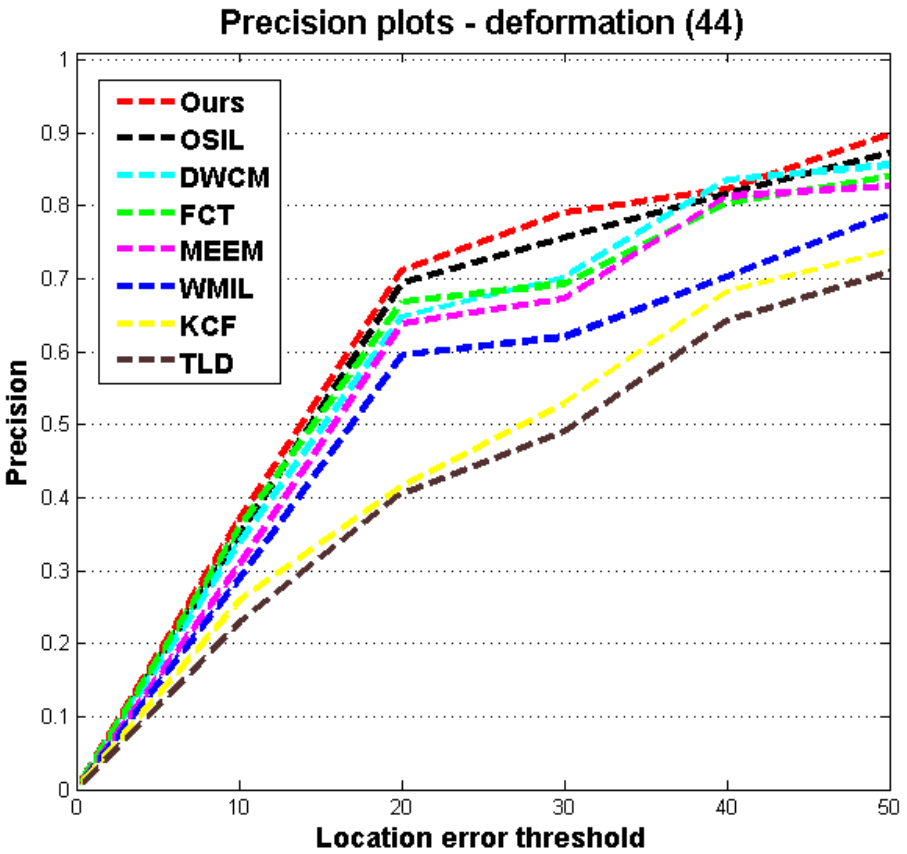}\end{subfigure} &
			\begin{subfigure}{0.34\textwidth}\centering\includegraphics[height=4.3cm, width=4.3cm]{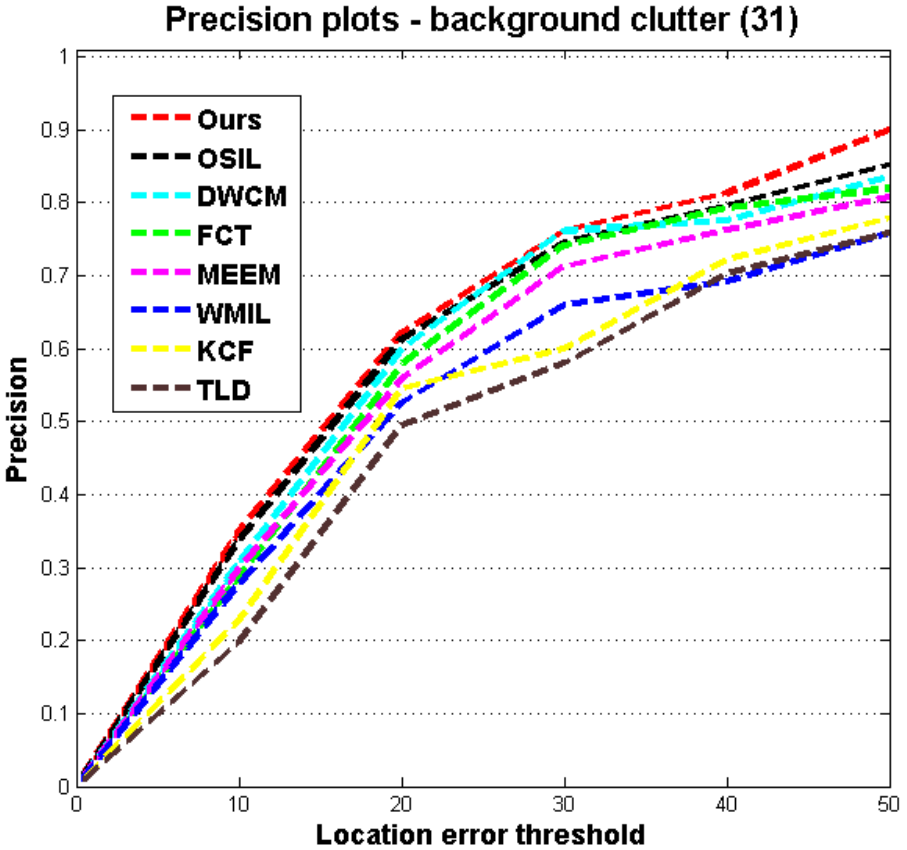}\end{subfigure} 
		\end{tabular}
	}
	\caption{
		\label{precision}The precision plots of the Ours, \textcolor{blue}{KCF, MEEM, TLD,} CT and MIL based trackers (FCT, WMIL, DWCM and OSIL) on the various attributes. Numeric value in the bracket indicates the total number of sequences of OTB100 dataset used for particular attribute.} 
\end{figure*}
\begin{figure*}
	\hspace{-.4cm}
	\resizebox{1\textwidth}{!}{
		\begin{tabular}{ccc}
			\vspace{.2cm}
			\begin{subfigure}{0.34\textwidth}\centering\includegraphics[height=4.3cm, width=4.3cm]{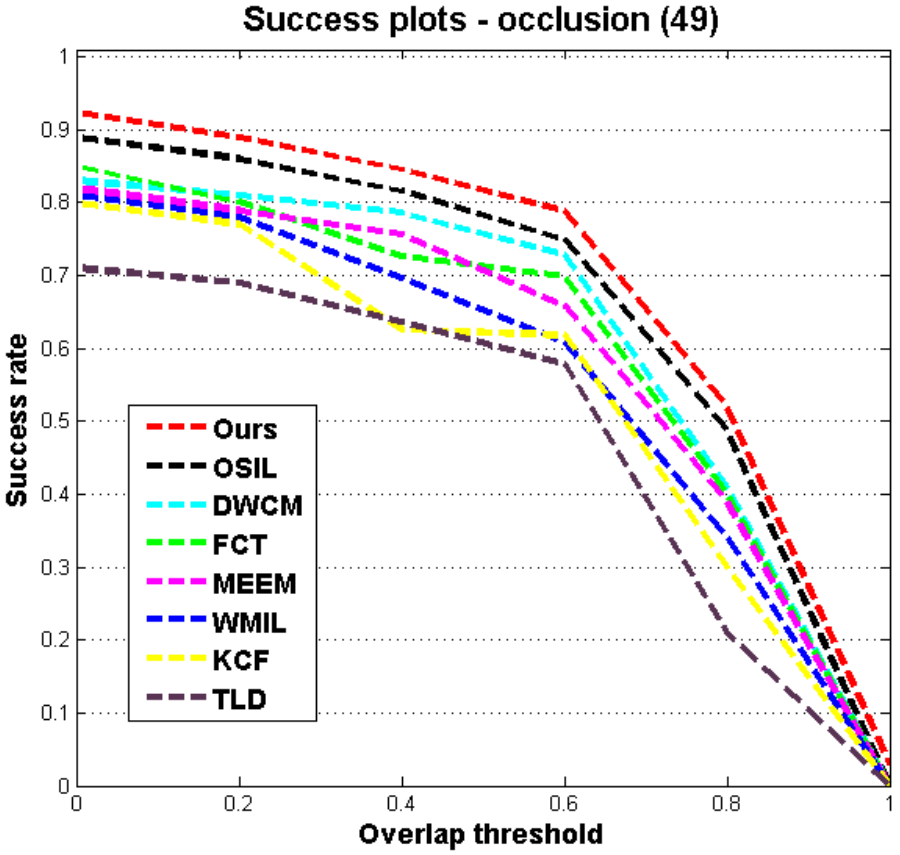}\end{subfigure}&
			\begin{subfigure}{0.34\textwidth}\centering\includegraphics[height=4.3cm, width=4.3cm]{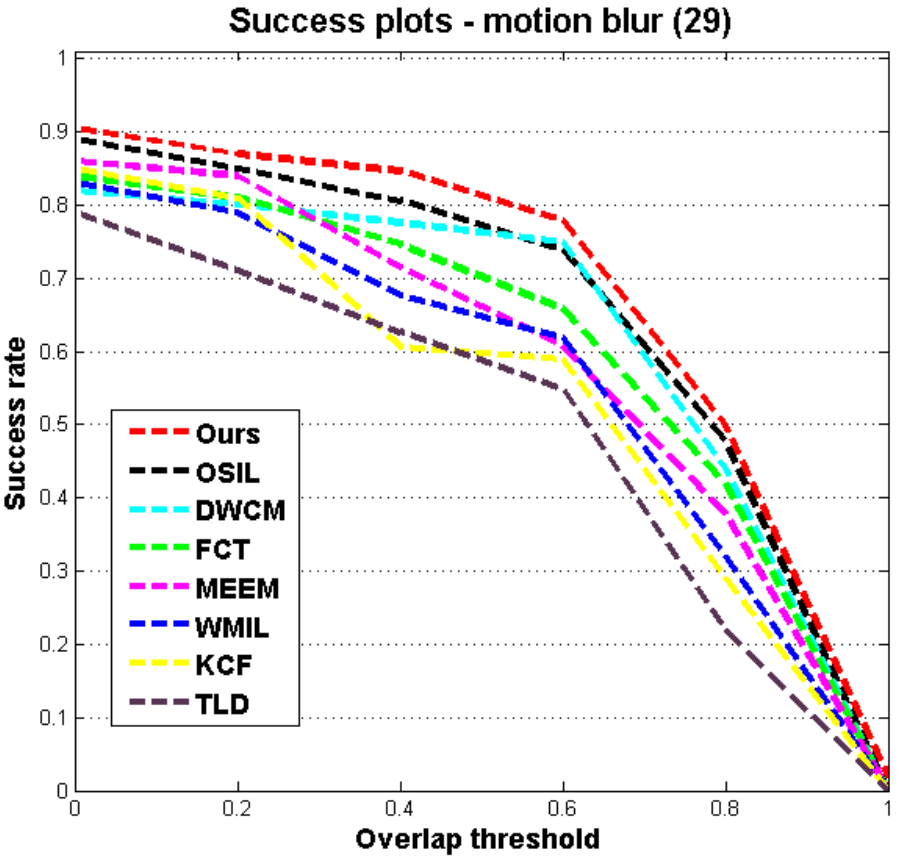}\end{subfigure}&
			\begin{subfigure}{0.34\textwidth}\centering\includegraphics[height=4.3cm, width=4.3cm]{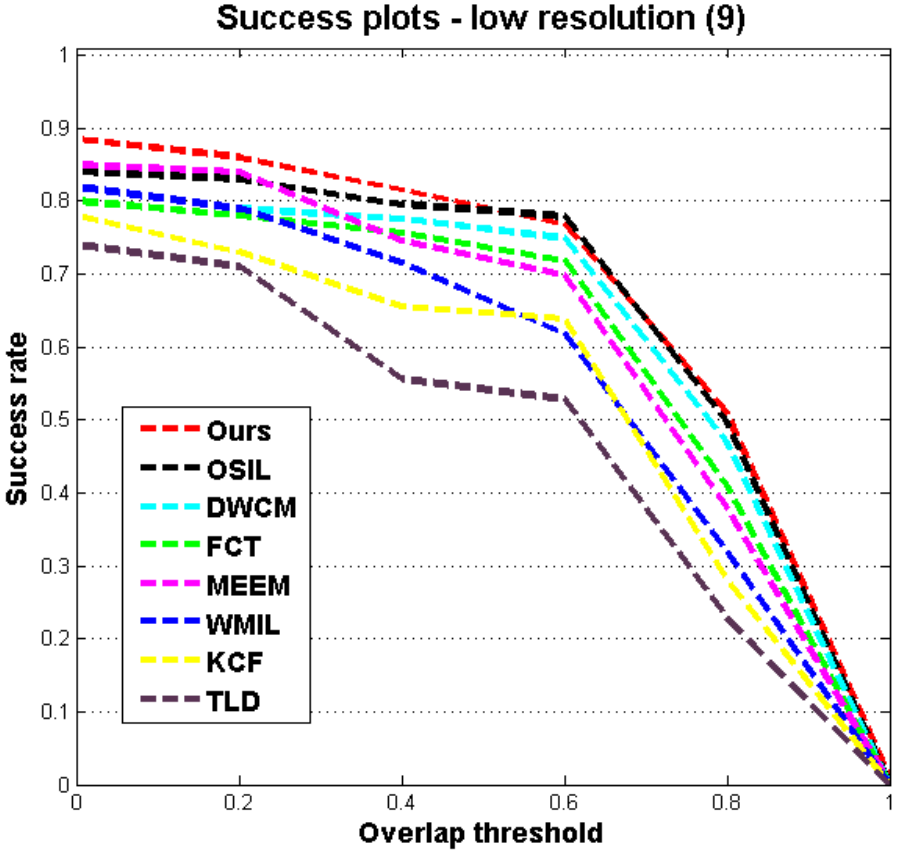}\end{subfigure} \\
			\vspace{.2cm}
			\begin{subfigure}{0.34\textwidth}\centering\includegraphics[height=4.3cm, width=4.3cm]{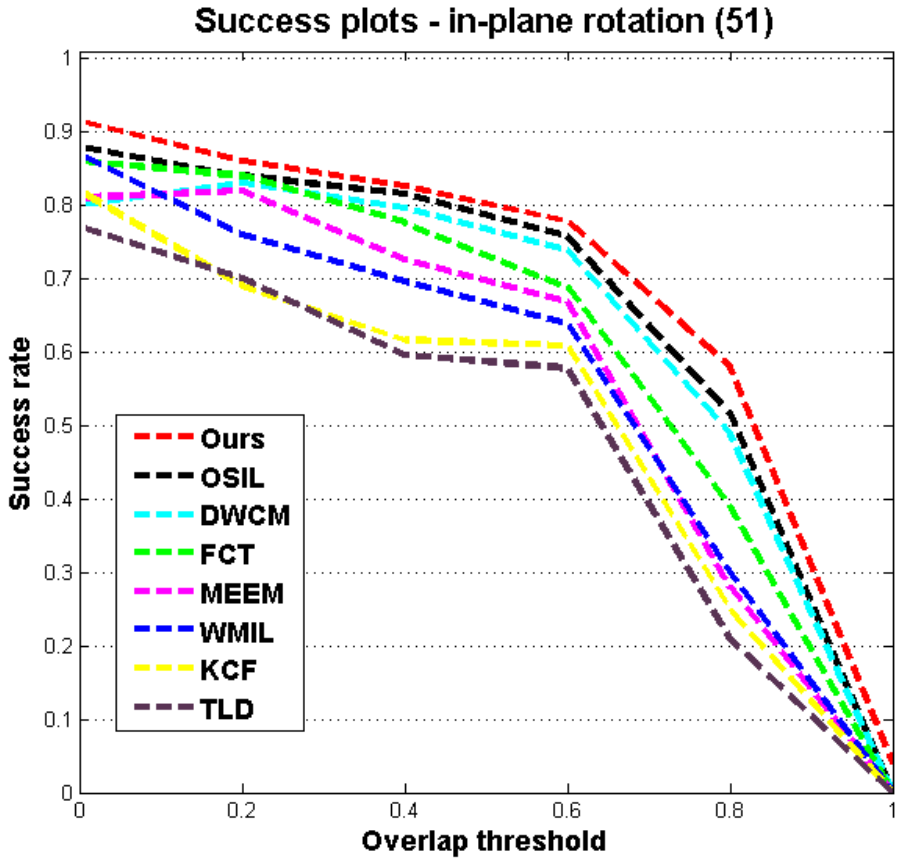}\end{subfigure}&
			\begin{subfigure}{0.34\textwidth}\centering\includegraphics[height=4.3cm, width=4.3cm]{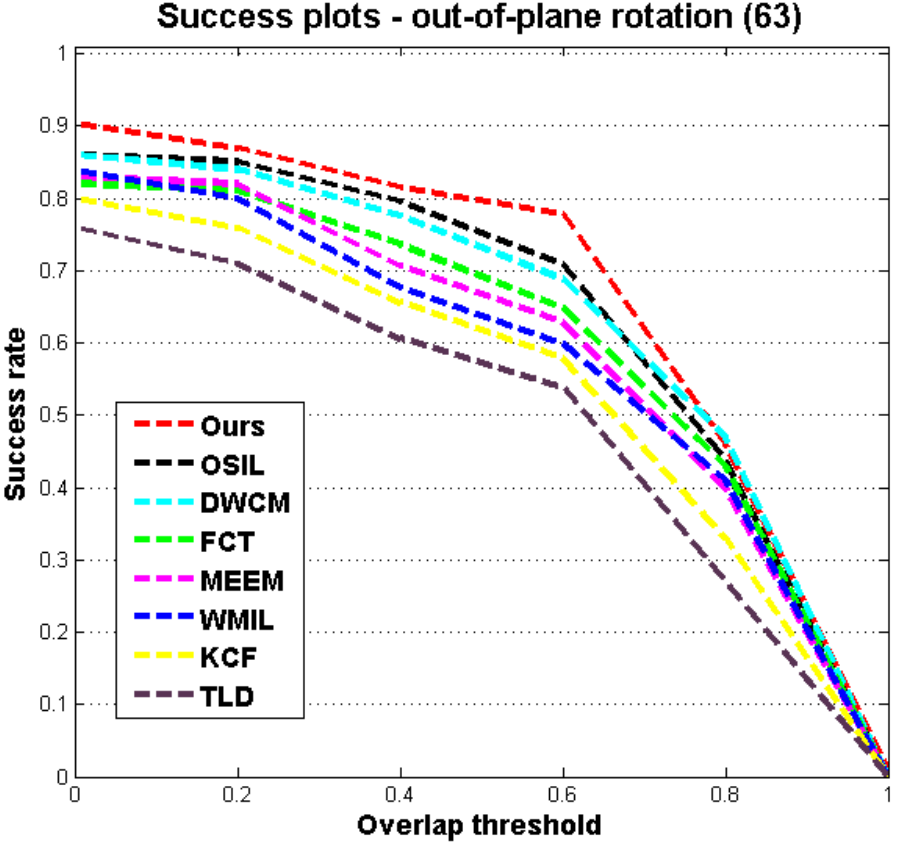}\end{subfigure}&
			\begin{subfigure}{0.34\textwidth}\centering\includegraphics[height=4.3cm, width=4.3cm]{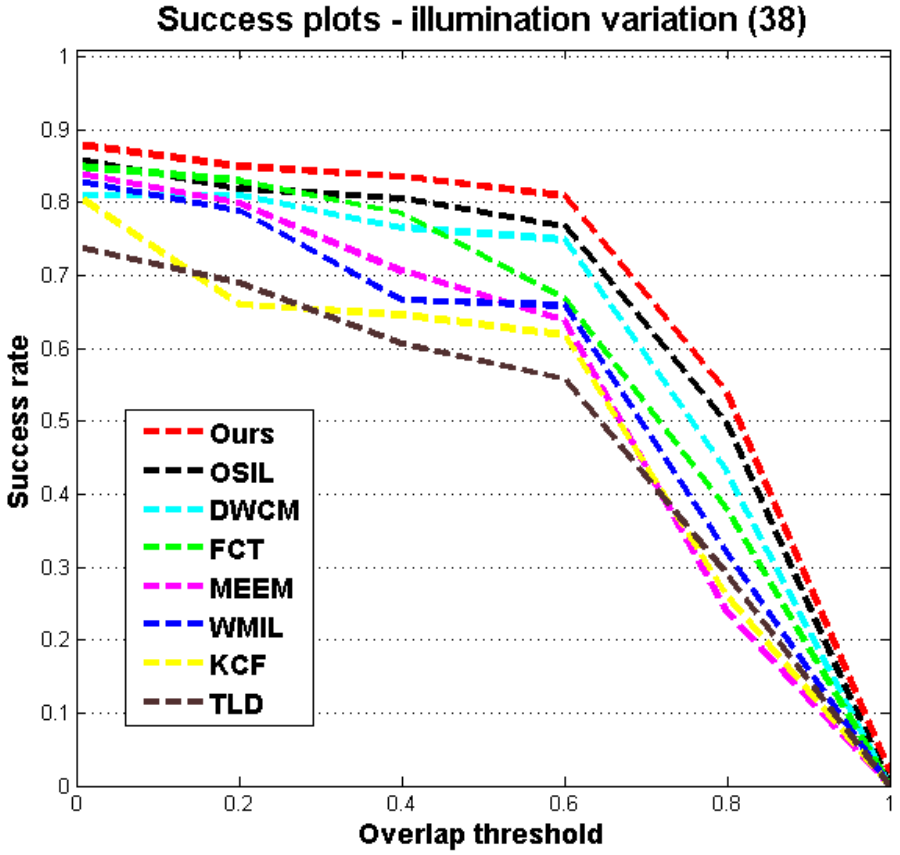}\end{subfigure} 	\\
			\vspace{.2cm}
			\begin{subfigure}{0.34\textwidth}\centering\includegraphics[height=4.3cm, width=4.3cm]{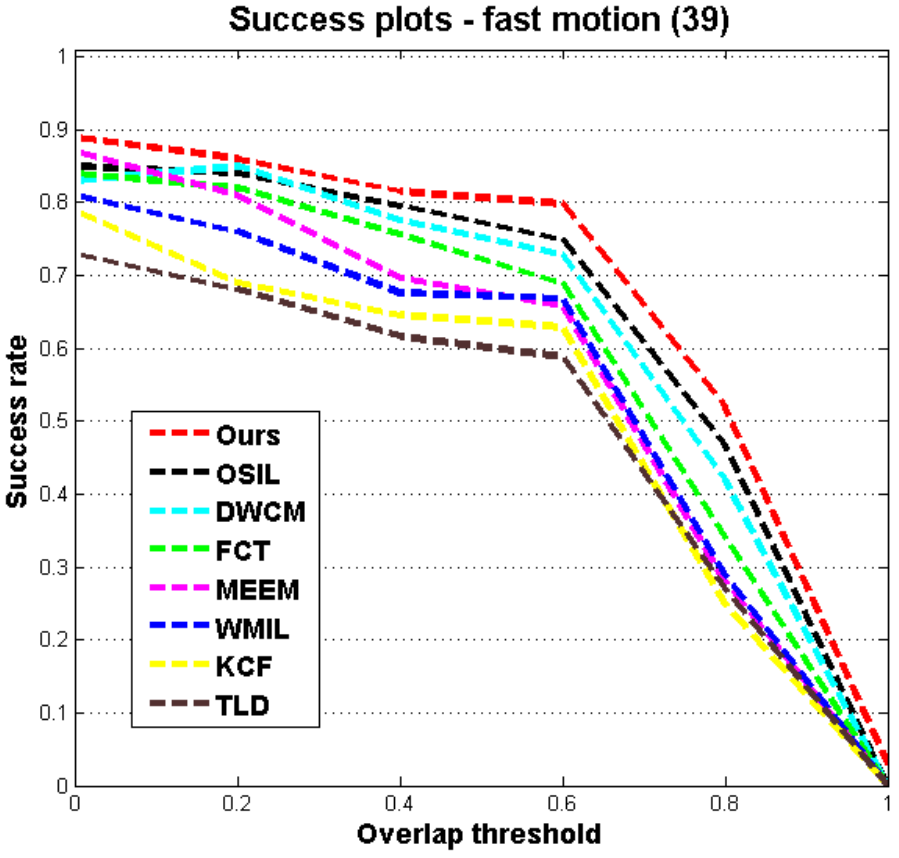}\end{subfigure} &
			\begin{subfigure}{0.34\textwidth}\centering\includegraphics[height=4.3cm, width=4.3cm]{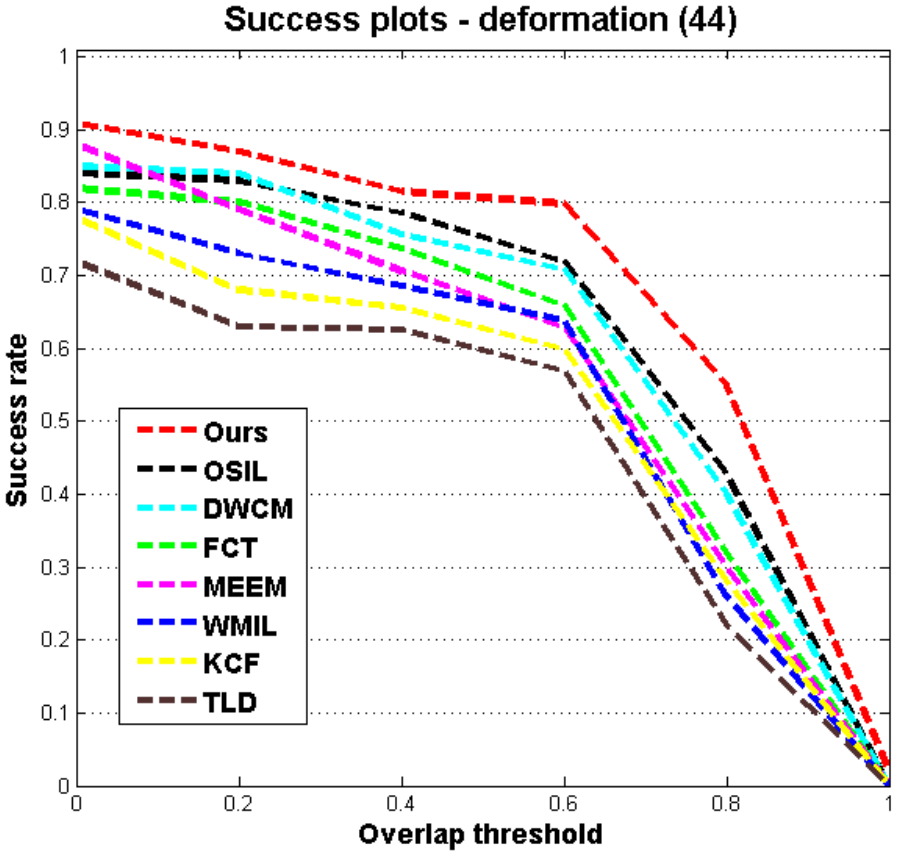}\end{subfigure} &
			\begin{subfigure}{0.34\textwidth}\centering\includegraphics[height=4.3cm, width=4.3cm]{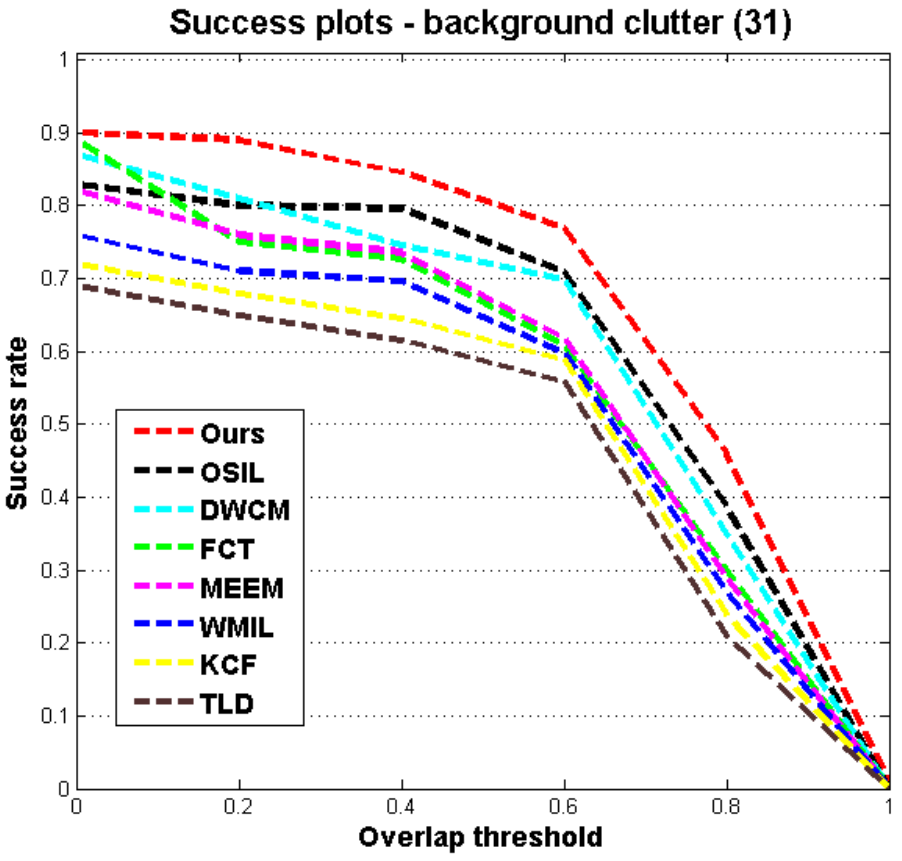}\end{subfigure} \end{tabular}
	}
	\caption{
		\label{success}The success plots of the Ours, \textcolor{blue}{KCF, MEEM, TLD,} CT and MIL based trackers (FCT, WMIL, DWCM and OSIL) on the various attributes. Numeric value in the bracket indicates the total number of sequences of OTB100 dataset used for particular attribute.} 
\end{figure*}
\subsubsection{Comparison with CT and MIL-based trackers}
\label{compare1}
We compared proposed method with the existing four representative CT and MIL-based trackers: FCT~\cite{ZhangK2}, WMIL~\cite{ZhangK}, DWCM~\cite{ChenT}, and OSIL~\cite{YanJ}. The FCT approach models the object appearance based on the non-adaptive random projections of rectangular Haar-like features directly. Furthermore, the computational complexity in the detection process is reduced by a coarse-to-fine search strategy. The DWCM technique is also based on the same concept of non-adaptive random projections, along with several random measurement matrices with different dimensions instead of single matrix as used in FCT to extract the features. A WMIL tracker is the extended method of MIL and it integrates the sample importance in the online learning process to effectively discriminate the positive samples.  The MIL and CT based integrated OSIL method reduces the effects of appearance change and occlusion.  In contrast to these approaches, our hybrid algorithm encompasses the advantages of both the WMIL and FCT methods to effectively and efficiently track the target object.\\ 
\textbf{Attribute based performance:} To compare the performance of the proposed technique with other CT and MIL based trackers for different challenging scenarios, Fig.~\ref{precision} and~\ref{success} show the precision and success plots of our method with the existing FCT, WMIL, DWCM and OSIL approaches for various attributes. Here we have shown the precision plots for 0 to 50 location error threshold and our method gives relatively high value of precision for all the attributes except fast motion and \textcolor{blue}{deformation} at some threshold. However the average precision of our method is relatively high with these two attributes also. The success rate of our method and other existing techniques are also shown with the help of different threshold parameter $\theta$ in the interval [0, 1]. It is also considerably high with all the attributes except on the some threshold values of \textcolor{blue}{low resolution,} out-of-plane rotation attributes. However the area under curve (AUC) values of our method for these attributes are high. 
\par The MIL, WMIL, CT, FCT and DWCM schemes do not perform well with most of the sequences because due to the appearance variations or occlusion, the spatial information of the target is lost by its Haar-like features, consequently the selected high confidence features are less distinguishable. We can see in the aforementioned figures that sub-region based integrated OSIL scheme has better results than other methods, which consider the concept of spatial information by acquiring the features from non-occluded sub-regions and also combines both the CT and MIL methods. However this scheme does not consider the sample importance and coarse-to-fine search strategies in the account. Furthermore, as evidenced from the above mentioned figures, our method  further enhances the tracking performance and provides more stable results by (i) combining the WMIL method with FCT scheme, which reduces the shortcoming of the OSIL method and (ii) considering the spatial information into account.  
\textcolor{blue}{ 
\subsubsection{Comparison with KCF, MEEM and TLD  trackers} 
\label{compare_other}
We have also compared our technique with three methods,  KCF~\cite{henriques2015high}, MEEM \cite{zhang2014meem}, and TLD~\cite{Kalal} of other categories. KCF tracker exploits an online support vector machine learning process in Fourier domain.  This method also uses the circulant matrix computations to acquire high processing speed. The mixer of experts which depends on entropy minimization are employed in the MEEM tracker; here online SVM with twin prototypes are exploited as the base tracker. TLD train an object detector based on patches found on the trajectory of the optic-flow-based approach. In this method, if the discovered patches are analogous to the initial patch then only the updates are carried out. Furthermore, to prove the effectiveness of the proposed method, similar to Sec.~\ref{compare1} we compare the techniques based on different challenging attributes in the OTB100 dataset and  it is evident from the Figs.~\ref{precision} and \ref{success} that, our method outperforms the KCF, MEEM and TLD trackers for all the challenging environments.}  
\subsubsection{Comparisons with the state-of-the-art trackers}
\label{compare2}
The proposed approach has been compared with \textcolor{blue}{16} state-of-the-art tracking algorithms, namely, (i) SP (Sparse prototypes based Tracker)~\cite{WangD}, (ii) MFT (Median-Flow tracker) \cite{KalalZ}, (iii) L1T (L1 tracker)~\cite{MeiX},  (iv) IVT (Incremental visual tracker)~\cite{RossD}, (v) CT (Compressive tracking)~\cite{ZhangK1}, (vi) DCT (Dynamic compressive tracking)~\cite{ChenT1},  (vii) VTD (Visual tracking decomposition)~\cite{KwonJ}, (viii) TLD (Tracking learning detection)~\cite{Kalal}, (ix) L1APG (L1  tracker using accelerated proximal gradient approach)~\cite{BaoC}, (x) STRUCK (Struck method)~\cite{HareS}, (xi) OSIL (Online sparse instance learning)~\cite{YanJ}, (xii) DWCM (Dynamic weighted compressive model)~\cite{ChenT}, (xiii) WMIL (Weighted multiple instance boosting based tracker)~\cite{ZhangK}, (xiv) FCT (Fast compressive tracking)~\cite{ZhangK2}, \textcolor{blue}{(xv) KCF (Kernel correlation filter)~\cite{henriques2015high}, and (xvi) MEEM (Multiple experts using entropy minimization)~\cite{zhang2014meem}}. The summary of all these approaches are provided in Table.~\ref{testedMethod}. For the experimental purpose, we used the publicly available source code for most of the methods. However the source code for DWCM and OSIL approaches are not provided by the authors, therefore we have implemented these methods as per the original publication. For the fair comparison, the same parameters setting as employed by the authors in their original work has been used.
\begin{sidewaystable}\small
	\caption{Summary of all the tested tracking algorithms.}
	\label{testedMethod}
	\resizebox{.9\textwidth}{!}{
		\begin{tabular*}{1\textwidth}{ l cccc}
			\cline{1-5}
			Trackers & Object representation & Appearance model & Approach & Classifier  \\
			\cline{1-5}
			L1APG~\cite{BaoC}, L1T~\cite{MeiX} & holistic image intensity & sparse representation & generative & - \\
			SP~\cite{WangD} & holistic image intensity & sparse principal component analysis & generative & - \\
			IVT~\cite{RossD} &holistic image intensity & incremental principal component analysis & generative & - \\
			MFT~\cite{KalalZ} & image intensity & - & generative & -\\
			CT~\cite{ZhangK1}, FCT~\cite{ZhangK2}, DCT~\cite{ChenT1} & Haar-like features & - & discriminant & naive Bayes\\
			TLD~\cite{Kalal} & Haar-like features & - & discriminant & cascaded\\
			\textcolor{blue}{KCF}~\cite{henriques2015high} & \textcolor{blue}{histogram of
			oriented gradients features} & - & \textcolor{blue}{discriminant} & \textcolor{blue}{linear kernel}\\
			WMIL~\cite{ZhangK} & Haar-like features & online multiple instance learning & discriminant & boosting\\
			\textcolor{blue}{MEEM}~\cite{zhang2014meem} & \textcolor{blue}{image intensity} & - & \textcolor{blue}{discriminant} & \textcolor{blue}{linear SVM}\\
			Struck~\cite{HareS}& Haar-like features & - & discriminant & structured SVM\\
			VTD~\cite{KwonJ}& hue, saturation, intensity and edge template & sparse principal component analysis & generative & -\\
			DWCM~\cite{ChenT} & Haar-like features & sparse representation & discriminant & naive Bayes\\
			Ours, OSIL~\cite{YanJ}& Haar-like Features  & sparse representation & discriminant & boosting\\
			
			\cline{1-5}
		\end{tabular*}	
	}
\end{sidewaystable}

\begin{table}[b!]\small  
	\caption{Quantitative analysis done by average center location errors (in pixels). The best and the second best performing techniques are displayed in \textbf{Bold} and \underline{Underline}, respectively.}
	\label{CLE}
	\centering
	\hspace{-5.9cm}
	\resizebox{0.54\textwidth}{!}{
		\begin{tabular*}{1\textwidth}{lccccccccccccccccc}
			\cline{1-18}
			Video & MFT & L1APG & SP & L1T & IVT & CT & TLD & KCF& WMIL & MEEM & DCT & STRUCK & VTD & DWCM & FCT & OSIL & Ours\\
			\cline{1-18}
			Motocross1 &65&58&44&36&38&42&37&31&20&25&19&14&16& 15 & 23&\underline{10} & \textbf{8} \\
			Coke1 & 47&57&54&40&43&26&\underline{12}&17&30&14&19&\textbf{10}&25&18&22&16&13\\
			David & 71&48& 14&42&\underline{10}&36&58&29&31&26&23&28&13&17&21&14&\textbf{8}\\
			Diving &46&61&53&44&39&26&33&21&17&18&13&12&22&\underline{6}&9&\underline{6}&\textbf{5}\\
			Football &59&48&54&44&41&38&34&39&17&31&29&24&\textbf{9}&19&32&13&\underline{11}  \\
			Mountain bike &83&21&99&65&59&53&37&12&27&10&46&14&39&33&34&\underline{10}&\textbf{7}\\
			Occluded face1 & 89 &21& 10&40&53& 37&14&23&28&22&32&12&38&14&20&\underline{9}&\textbf{6}\\
			Occluded face2 & 55&89&39&48&41&22&33&37&44&35&26&11&14&21&17&\underline{10}&\textbf{9}\\
			Panda &55&69&47&38&63&28&44&64&23&33&\underline{12}&88&103&13&19&33&\textbf{11}\\
			Shaking &89&67&110&53&59&49&150&37&41&12&\underline{11}&36&18&12&31&27&\textbf{8}\\
			Singer1 &69&79&71&98&64&35&33&63&49&32&20&16&\underline{12}&15&19&14&\textbf{11}\\
			Singer2 &65&83&99&53&59&39&37&59&46&25&13&19&34&\textbf{7}&12&\underline{10}&\underline{10}\\
			Sylvester &89&37&59&68&33&27&10&39&47&41&21&13&29&17&11&\underline{9}&\textbf{7}\\
			Tiger 2 &68&39&36&28&52&35&48&39&11&\underline{9}&17&13&43&21&16&\textbf{8}&\textbf{8}\\
			Trans & 79&57&43&36&28&22&15&45&26&24&10&13&17&\underline{5}&8&8&\textbf{3}\\
			Woman &149&29&\textbf{8}&96&127&87&35&40&79&43&59&18&72&49&64&13&\underline{10}\\
			\textcolor{blue}{OTB100} &76.1&61.0&58.5&54.7& 62.3 & 38.4 & 41.9& 39.3&35&28.8& 25.6 & 24.8 &37.5 &18.1&23.4& \underline{15.2} & \textbf{12.1} \\
			\textcolor{blue}{Average FPS} & 20.5 & 5.1 & 7.9 & 4.7 & 9.8 & \underline{23.8} & 10.3 &21& 11.9&10.5 & 7.0 & 8.1 & 4.7 & 9.6 & \textbf{35.6} & 9.8 & 10.8 \\
			\cline{1-18}
		\end{tabular*}
	}
\end{table}
\begin{table}[t]\small
	\caption{Quantitative analysis done by success rates (SR) (\%). The best and the second best performing techniques are displayed in \textbf{Bold} and \underline{Underline}, respectively.}
	\label{SR}
	\centering
	\hspace{-5.9cm}
	\resizebox{0.54\textwidth}{!}{
		\begin{tabular*}{1\textwidth}{lccccccccccccccccc}
			\cline{1-18}
			Video & MFT & L1APG & SP & L1T & IVT & CT & TLD & KCF & WMIL & MEEM & DCT & STRUCK & VTD & DWCM & FCT & OSIL & Ours \\
			\cline{1-18}
			Motocross1 & 34 & 49 & 59 & 62 & 74 & 70 & 78 & 76& 82& 86 & 85 & 89 & 87 & 90 & \underline{93} & 92 & \textbf{95} \\
			Coke1        & 28 & 22 & 27 & 31 & 30 & 61 & \underline{83} & 54&55&69 & 72 & \textbf{85} & 66 & 72 & 69 & 80 & 82\\
			David        & 21 & 65 & \underline{97} & 70 & \underline{97} & 77 & 44 & 62&79&73 & 81 & 77 & 96 & 88 & 85 & 90 & \textbf{98} \\
			Diving       & 67 & 21 & 49 & 75 & 79 & 84 & 87 & 71&86&80 & 89 & 91 & 85 & \underline{94} & 93 & \underline{94} & \textbf{96} \\
			Football     & 21 & 51 & 37 & 55 & 59 & 66 & 70 & 70&78&63 & 71 & 72 & \textbf{84} & 76 & 68 & \underline{82} & 81 \\
			Mountain bike & 41 & 83 & 19 & 43 & 59 & 62 & 75 & 78&79&81 & 67 & 89 & 72 & 77 & 80 & \underline{91} & \textbf{98}\\
			Occluded face1 &  17&58&91&37&22&22&69&40&37&61&54&87&25&75&66&\underline{95}&\textbf{97}\\
			Occluded face2 & 56&42&66&59&68&84&79&55&60&86&83&94&92&85&90&\underline{96}&\textbf{97}\\
			Panda &58&43&61&67&51&76&64&50&77&70&91&24&14&\underline{93}&85&71&\textbf{94}\\
			Shaking &53&62&27&73&66&65&11&65&71&80&\underline{91}&77&89&90&79&83&\textbf{94}\\
			Singer1  &39&38&24&61&65&69&71&82&81&75&83&87&\underline{92}&89&91&90&\textbf{94}\\
			Singer2  &43&41&19&62&59&72&75&58&67&55&85&89&80&\textbf{94}&91&89&\underline{92}\\
			Sylvester &24&61&47&39&69&76&\underline{88}&49&52&65&81&85&78&83&82&86&\textbf{89}\\
			Tiger 2 &11&32&39&38&15&32&17&65&75&71&57&77&25&45&55&\underline{86}&\textbf{89}\\
			Trans &29&42&62&71&79&81&83&69&79&71&93&89&86&\underline{95}&94&92&\textbf{100}\\
			Woman &21&86&\textbf{91}&36&29&41&73&60&48&86&62&85&51&68&57&\underline{90}&89\\
			\textcolor{blue}{OTB100} & 33.3 & 41.4 & 44.1 & 45.5 & 53.7 & 54.1 & 57.0 & 57.2& 59.3&65.6 & 71.5 & 77.2 & 68.5 & 79.8 & 73.3 & \underline{81.7} & \textbf{90.2} \\
			\cline{1-18}
		\end{tabular*}
	}
\end{table}
\begin{figure*}
	\hspace{-.5cm}
	\resizebox{1\textwidth}{!}{
		\begin{tabular}{ccc}
			\vspace{.2cm}
			\begin{subfigure}{0.5\textwidth}\centering\includegraphics[height=6cm, width=6.3cm]{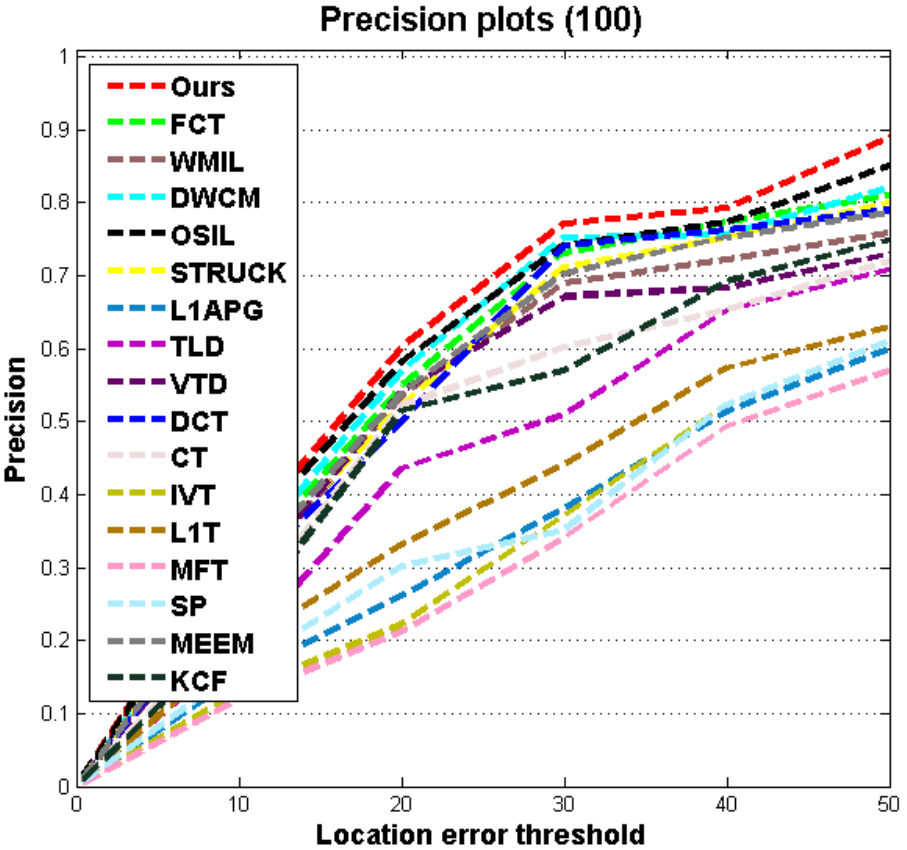}\end{subfigure}&
			\hspace{.3cm}
			\begin{subfigure}{0.5\textwidth}\centering\includegraphics[height=6cm, width=6.3cm]{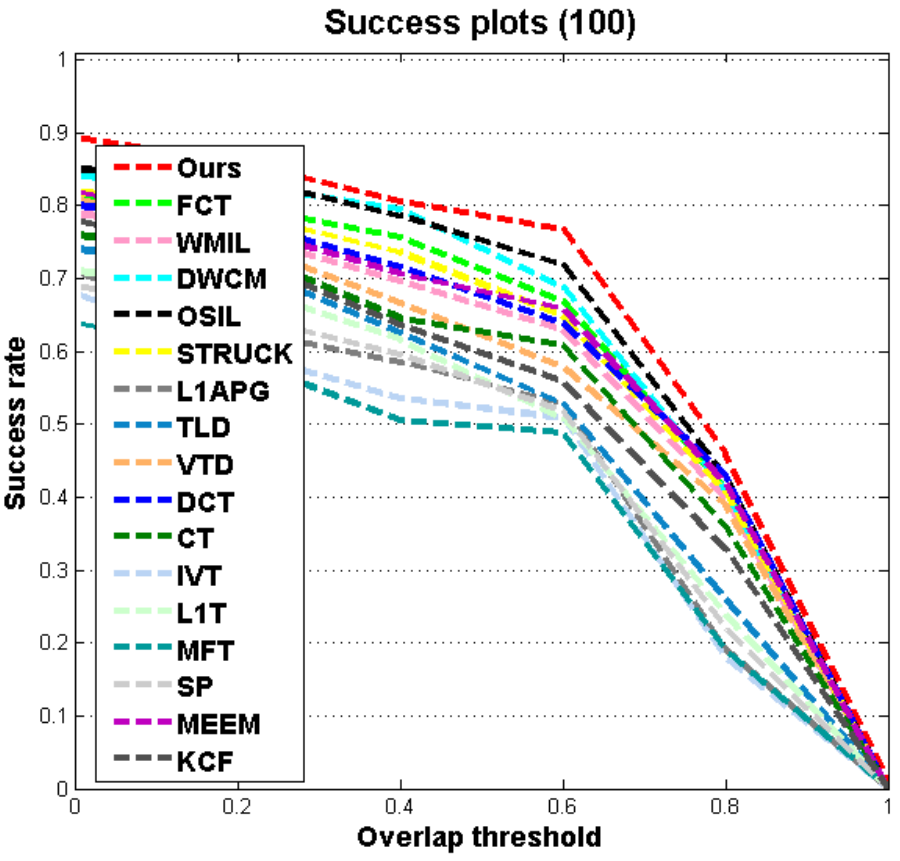}\end{subfigure}
		\end{tabular}
	}
	\caption{
		\label{overall}
		Plots of precision and success rate for the \textcolor{blue}{OTB100 dataset}: a comparison of our technique with the state-of-the-art approaches. Numeric value \textcolor{blue}{within} the bracket indicates the total number of \textcolor{blue}{video} sequences used.} 
\end{figure*}
\textbf{Quantitative analysis:} Table~\ref{CLE} shows the mean of the average center location errors on \textcolor{blue}{sixteen video sequences of OTB100 dataset chosen randomly. It also shows the average errors of all the 100 video sequences of OTB100 benchmark.} The best results are represented in \textbf{bold} \textcolor{blue}{while} the second best are \underline{underlined}. Here the best possible centre location \textcolor{blue}{errors} would be zero. \textcolor{blue}{The proposed method gives the best or second best results of average center location error for most of the videos, especially}, in the Diving and Trans sequences, our tracker has superior performance. Furthermore, our method does not have the best or the second best performance for Coke1 sequence. \textcolor{blue}{However in the second last row (for the dataset OTB100), the proposed method achieves the lowest average mean 12.1 among all the compared trackers.} Table~\ref{SR} also summarizes the success rate with the average overlapping of at least 50\% between the bounding box of the ground truth and tracker. Here the best possible success rate would be 100. In the Trans sequences, our technique achieves the success rate of 100\%. In the \textcolor{blue}{Motocross1,} David, Diving, mountain-bike,   Occluded face1 and Occluded face2 sequences, the proposed technique achieved the success rate above 95\%. In the Coke1, Football and Woman sequences, the success rate of our method is neither the best nor the second best. \textcolor{blue}{However in the last row (for the dataset OTB100), the mean of average success rate of all the 100 sequences is 90.2\% for our tracker, which is the highest among all. As evident from} both the center location error and \textcolor{blue}{the} percentage area overlap (Table~\ref{CLE} and~\ref{SR} ) evaluation criteria, our approach performs the best and \textcolor{blue}{achieves better} results than other  algorithms on most of the \textcolor{blue}{video sequences of the OTB100 dataset.} 
\par  Furthermore, In Fig.~\ref{overall} we have also shown the comparison of our tracker with the state-of-the-art approaches with the help of precision and success plots for all the sequences of \textcolor{blue}{OTB100 dataset}. Our algorithm achieves considerably higher precision and success rate than other approaches. Moreover, the \textcolor{blue}{precision} rate of the proposed technique is slightly lower than that of OSIL on the smaller threshold value, but the area under curve (AUC) value for our method is considerably higher than these techniques. Overall in terms of precision \textcolor{blue}{as well as} success rate, our approach has noticeably better tracking performance than the state-of-the-art algorithms on all the tested sequences.
\par \textbf{Runtime performance:} As shown in the last row \lq{Average FPS}\rq~(i.e. average number of frames per second) of Table~\ref{CLE}, our tracking method has better average tracking speed (\textcolor{blue}{10.8} FPS) than most of the tested approaches except the \textcolor{blue}{MFT, CT, KCF, WMIL, and FCT}. Here, the tracking speed of our method is considerably lower than some methods due to (i) selection of robust features after dividing the sample regions into sub-regions and subsequently performing the operations on un-occluded sub-regions (ii) assigning weights to the positive samples. However, tracking accuracy of our tracker is significantly higher than the aforesaid methods.
\textcolor{blue}{
\subsection{Evaluation on VOT}
\label{VOT}
To prove the robustness and stability, the proposed method is also evaluated on VOT2015 benchmark dataset, which includes 60 video sequences with different challenging environments. The VOT challenges offer the community of visual tracking with an accurately defined and repeatable way of comparing trackers i.e. the target is initialized in the first frame. It is re-initialized again whenever the tracker fails (target lost).  The evaluation protocol in terms of accuracy score and robustness score is used to measure the performance of the tracker. These scores are estimated based on the bounding box overlapping and failure rate measures respectively.  Furthermore, a ranking analysis based on both the statistical and realistic significance of the accuracy and robustness performance gap between approaches are provided by the VOT evaluation. Finally, these ranks are averaged before they are finalized. Please refer~\cite{kristan2015visual} for detailed description.}
\par
\begin{figure*}
	\hspace{-.5cm}
	\resizebox{1\textwidth}{!}{
		\begin{tabular}{ccc}
			\begin{subfigure}{0.5\textwidth}\centering\includegraphics[height=6cm, width=6.3cm]{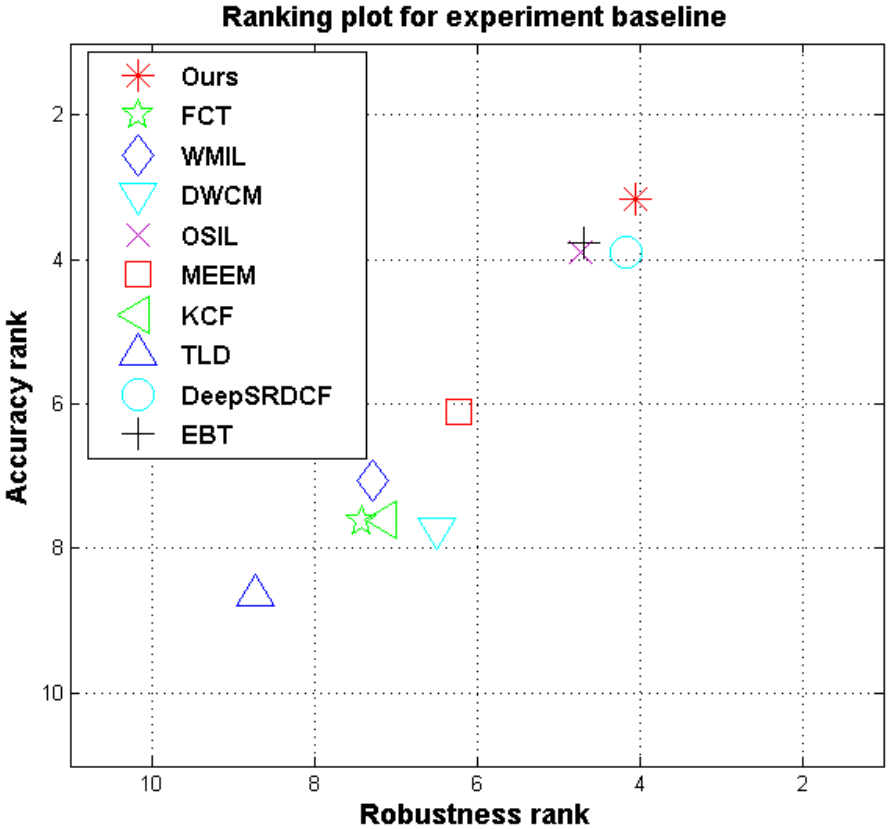}\caption{}\end{subfigure}&
			\hspace{.3cm}	\begin{subfigure}{0.5\textwidth}\centering\includegraphics[height=6cm, width=6.3cm]{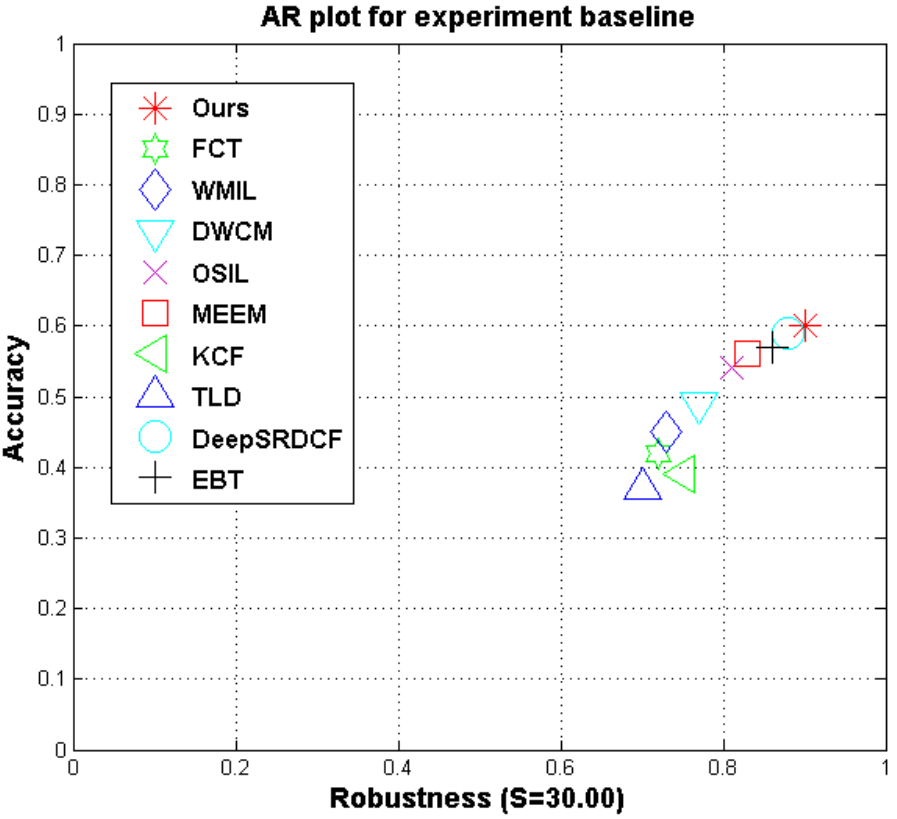}\caption{}\end{subfigure}
		\end{tabular}
	}
	\caption{
		\label{VOTgraph}
		The accuracy and robustness ranking and score plots of tested algorithms with VOT2015 benchmark. (a) ranking plot; (b) score plot.} 
\end{figure*}

 \textcolor{blue}{ Our method has been compared with the top 4 trackers of the VOT2015 benchmark (FCT~\cite{ZhangK2}, MEEM~\cite{zhang2014meem}, DeepSRDCF~\cite{danelljan2015convolutional}, and EBT~\cite{zhu2015tracking}). In addition to these we have also compared our method with one method of the VOT2014 challenge (KCF~\cite{henriques2015high}), two tracker of the OTB100 (KCF~\cite{henriques2015high}, TLD~\cite{Kalal}), and other three state-of-the-art trackers (WMIL~\cite{ZhangK}, DWCM~\cite{ChenT}, and OSIL~\cite{YanJ}).  All the 60 video sequences of VOT2015 dataset are used to generate the results and the experimental results produced by the VOT2015 toolkit~\cite{VOT2015} are shown in Table~\ref{rank}. As given in the aforementioned table, our tracker achieves the least failure rate as well as highest overlap value, which proves the robustness of the proposed method. Furthermore, the trackers are ordered as per the final rank and the proposed tracker acquires the best final rank (displayed in bold). The overall results on all the video sequences of the VOT2015 datasets are also displayed in Fig.~\ref{VOTgraph}. Here Fig.~\ref{VOTgraph}(a) and Fig.~\ref{VOTgraph}(b) display the accuracy-robustness rank and score of each tracker respectively. Each tracker in these plots is denoted as a point. A tracker close to the upper-right corner in these plots indicates a better result. As shown in Fig.~\ref{VOTgraph}, the proposed tracker is the closest to the upper-right corner and better than the DeepSRDCF and EBT tracker, which have top rank in the VOT2015 benchmark. Finally, experimental evaluations on the VOT2015 benchmark prove that our tracker is more robust and stable than other tested trackers.}
\begin{table}
	\caption{\textcolor{blue}{The experimental results produced by the VOT2015 benchmark toolkit.}}
	\label{rank}
	\hspace{.35cm}
	\begin{tabular*}{1\textwidth}{l|P{1.2cm}P{1.3cm}P{1.3cm}P{1.4cm}P{1.3cm}}
		\cline{1-6}
		Tracker & Overlap & Failure \newline rate & Accuracy rank & Robustness rank & Final rank \\
		\cline{1-6}
		Ours& \textbf{0.57}&\textbf{1.01}&\textbf{3.16}&\textbf{4.05}&\textbf{3.61}\\
		DEEPSRDCF&0.53&1.05&3.89&4.17&4.03\\
		EBT&0.45&1.06&3.77&4.69&4.23\\
		OSIL&0.52&1.17&3.90&4.74&4.32\\
		MEEM&0.46&2.05&6.11&6.23&6.17\\
		DWCM&0.41&2.08&7.73&6.51&7.12\\
		WMIL&0.44&2.98&7.05&7.28&7.17\\
		KCF&0.43&2.51&7.60&7.13&7.37\\
		FCT&0.43&3.34&7.62&7.43&7.53\\
		TLD&0.39&4.13&8.63&8.54&8.59\\
		\cline{1-6}
	\end{tabular*}
\end{table}

\subsection{Qualitative analysis}
\label{qualitative}
\begin{figure*}
	\resizebox{.987\textwidth}{!}{
		\begin{tabular}{ccccc}
			\begin{subfigure}{0.2\textwidth}\centering\includegraphics[height=2.8cm, width=2.8cm]{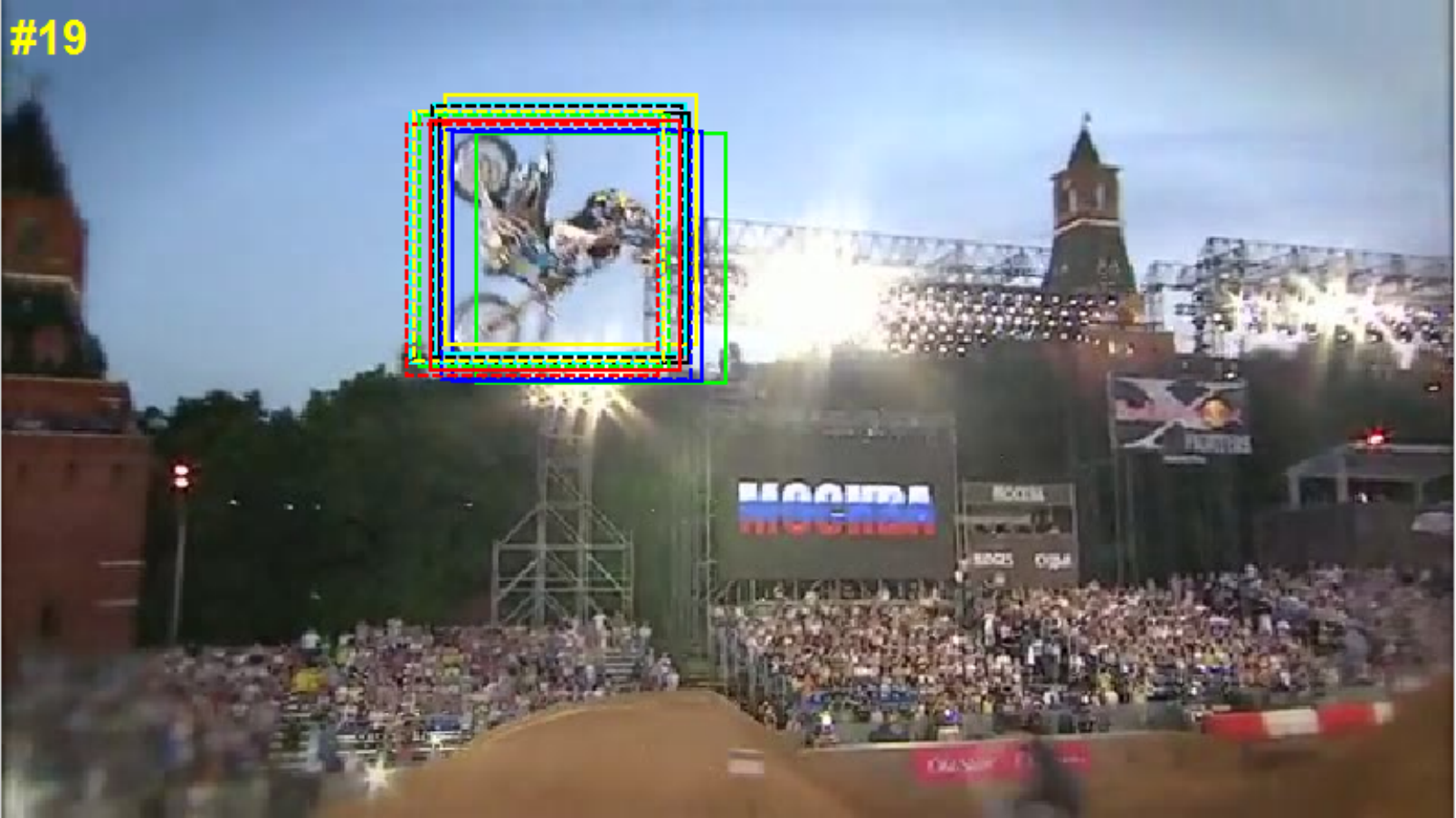}\end{subfigure}&
			\begin{subfigure}{0.2\textwidth}\centering\includegraphics[height=2.8cm, width=2.8cm]{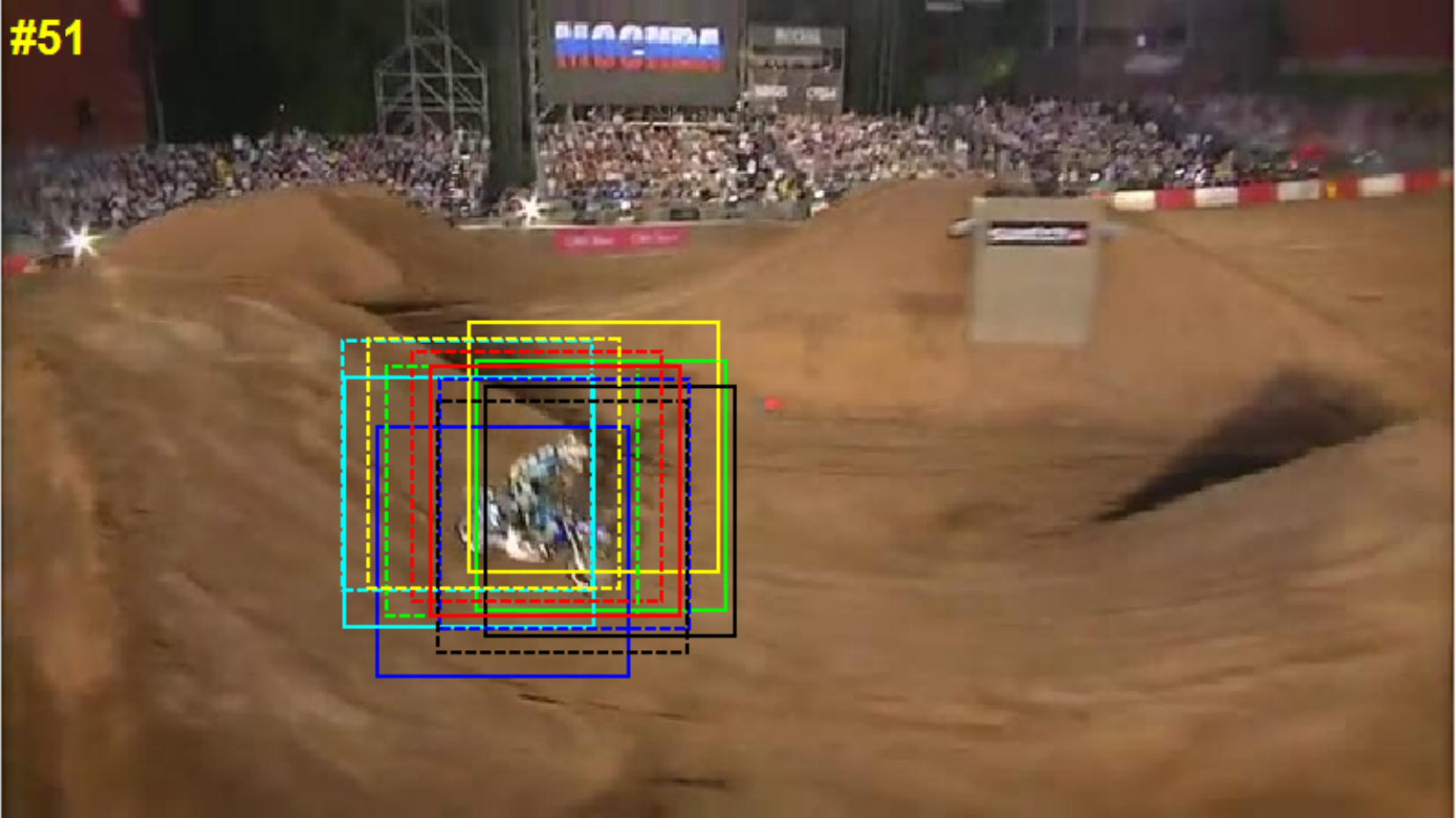}\end{subfigure}&
			\begin{subfigure}{0.2\textwidth}\centering\includegraphics[height=2.8cm, width=2.8cm]{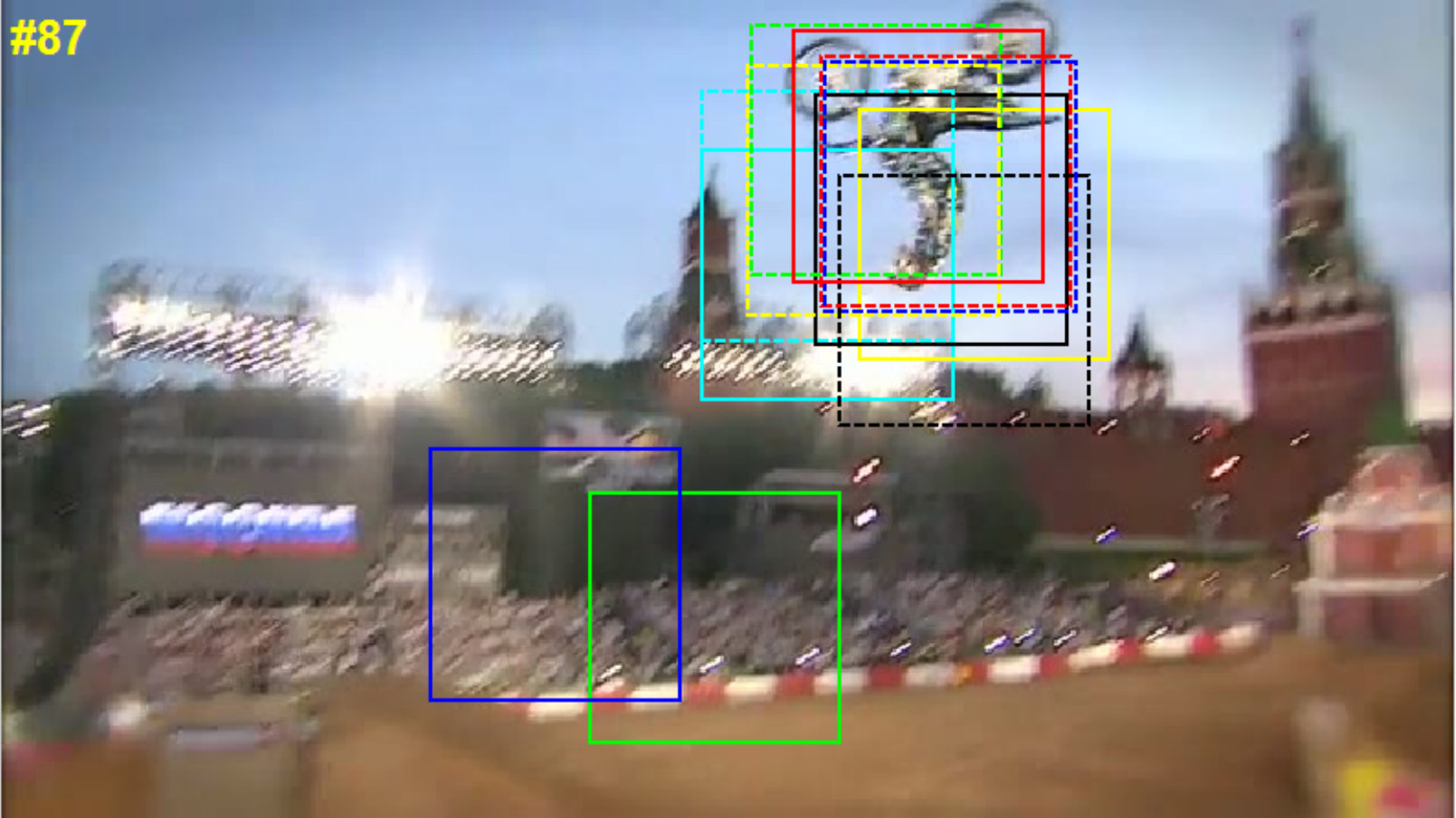}\end{subfigure} &
			\begin{subfigure}{0.2\textwidth}\centering\includegraphics[height=2.8cm, width=2.8cm]{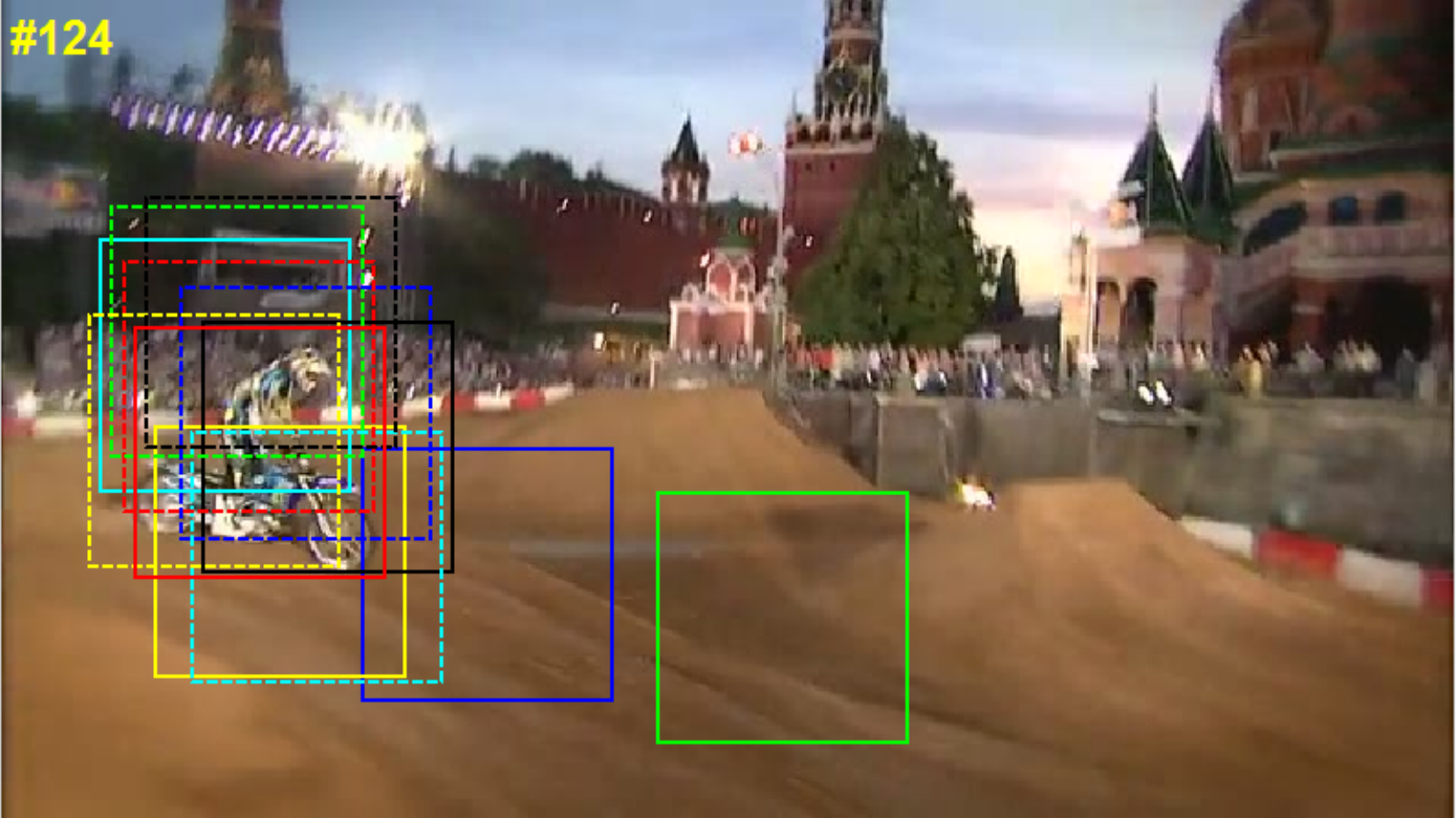}\end{subfigure} &
			\begin{subfigure}{0.2\textwidth}\centering\includegraphics[height=2.8cm, width=2.8cm]{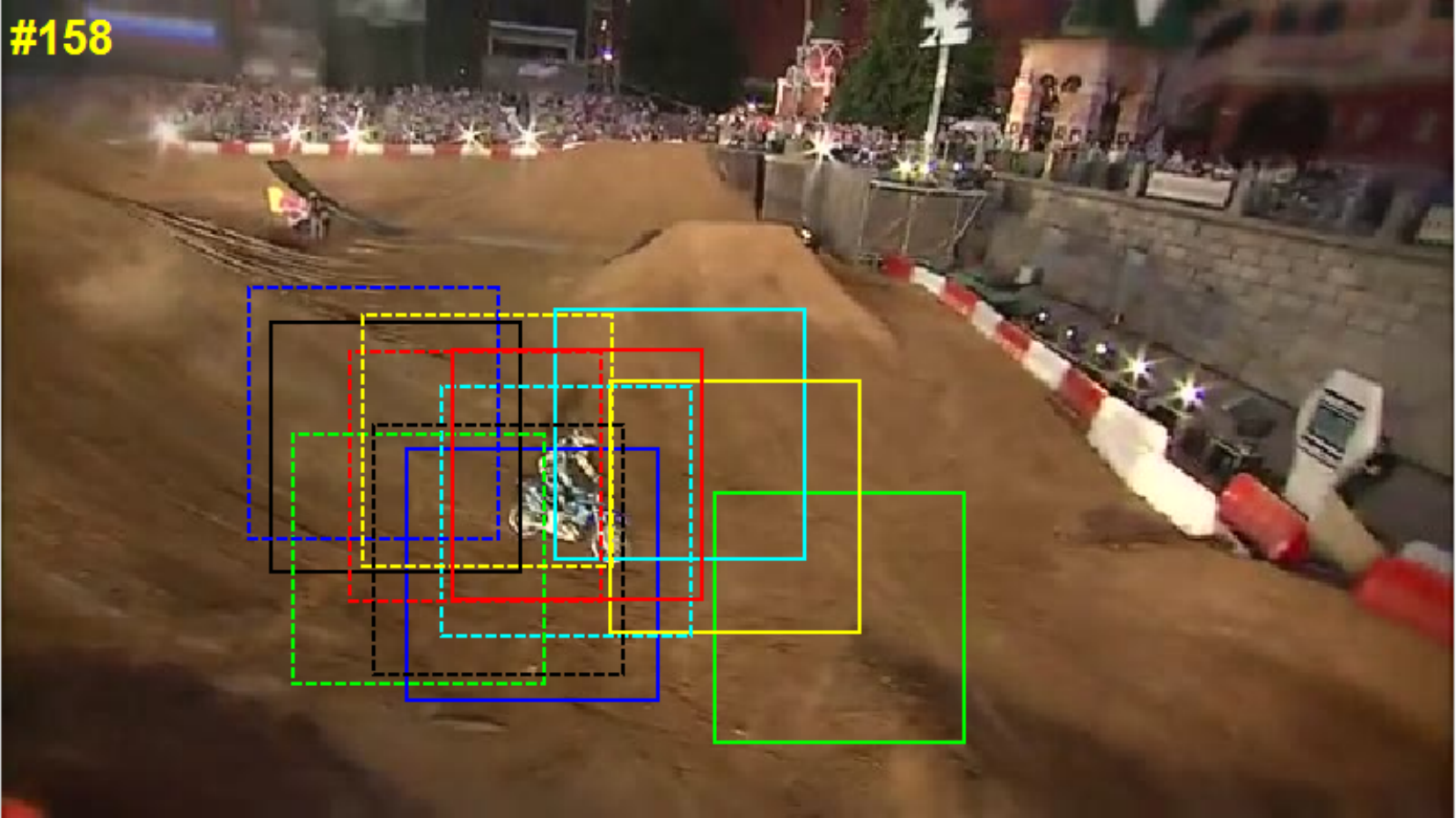}\end{subfigure} \\
			\multicolumn{5}{c}{(a) \textcolor{blue}{Motocross1}} \\
			\begin{subfigure}{0.2\textwidth}\centering\includegraphics[height=2.8cm, width=2.8cm]{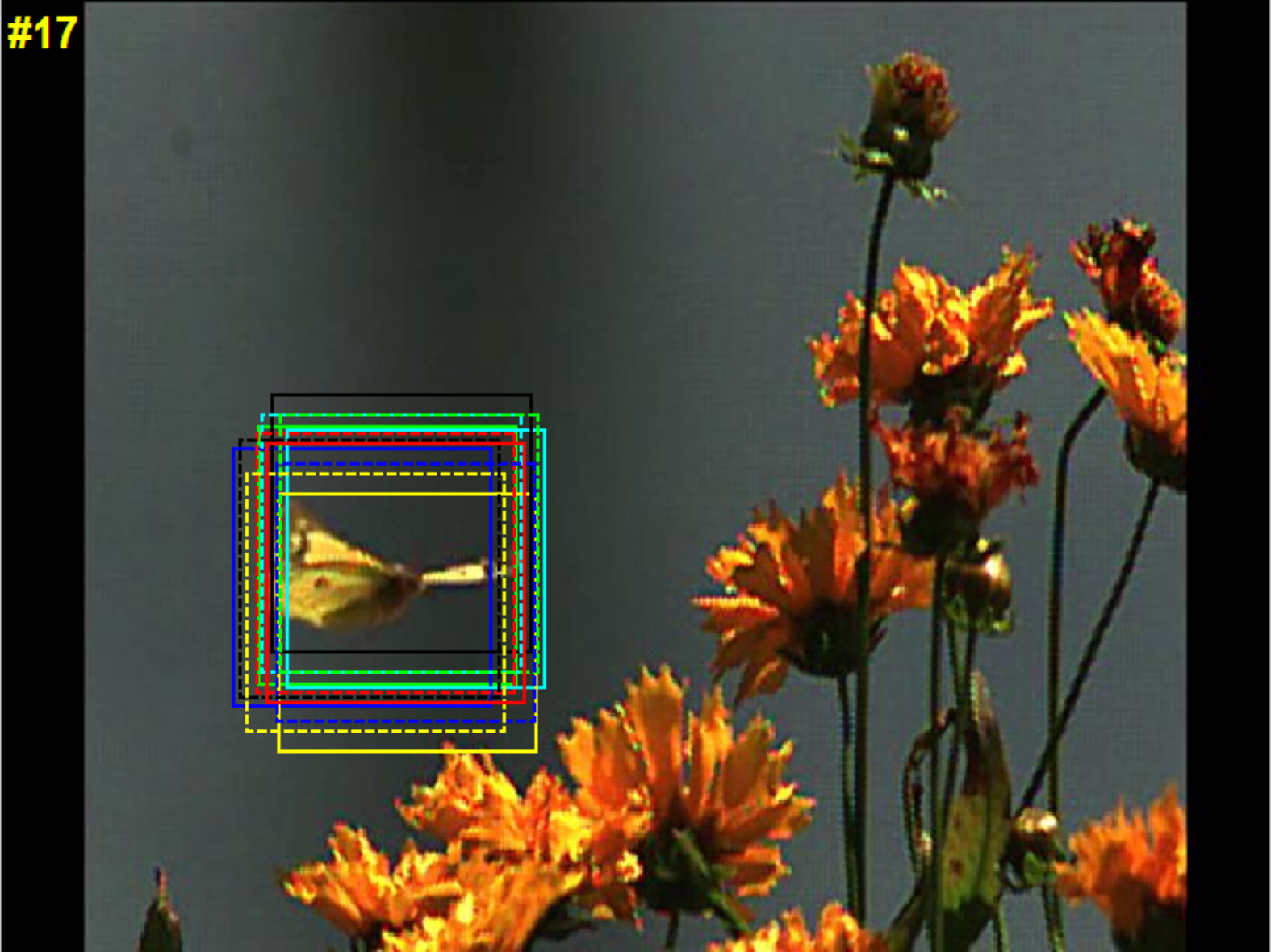}\end{subfigure}&
			\begin{subfigure}{0.2\textwidth}\centering\includegraphics[height=2.8cm, width=2.8cm]{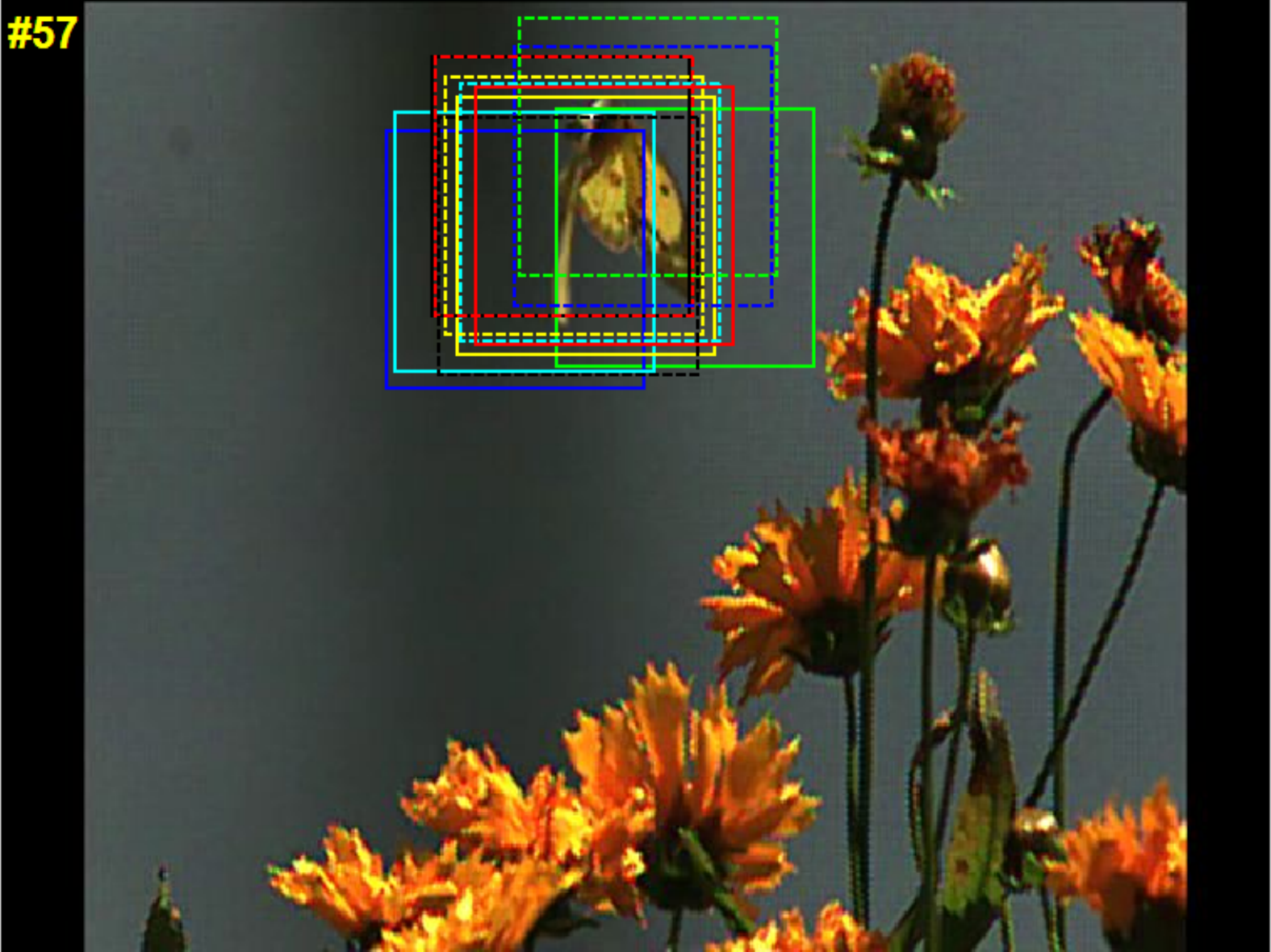}\end{subfigure}&
			\begin{subfigure}{0.2\textwidth}\centering\includegraphics[height=2.8cm, width=2.8cm]{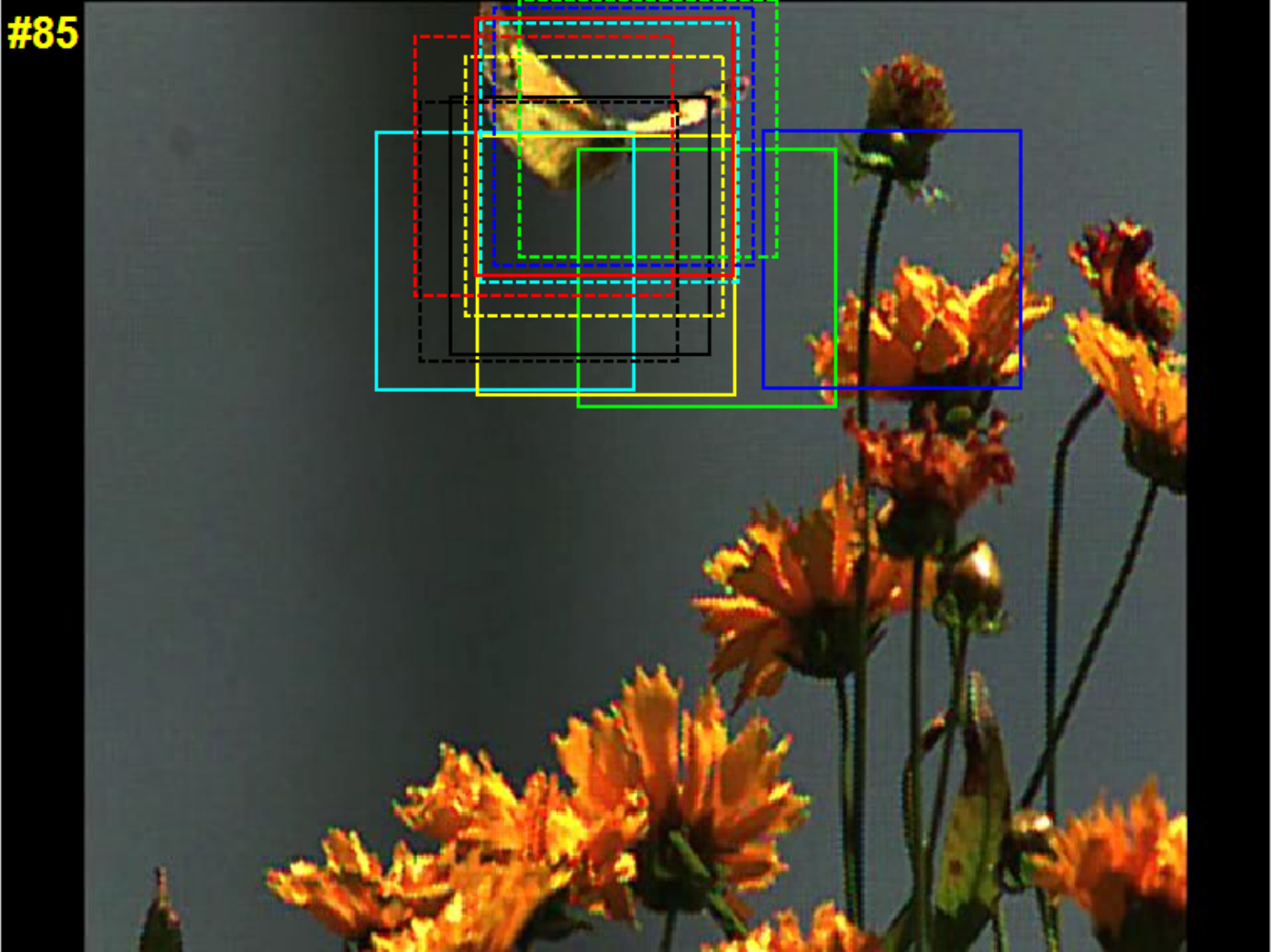}\end{subfigure} &
			\begin{subfigure}{0.2\textwidth}\centering\includegraphics[height=2.8cm, width=2.8cm]{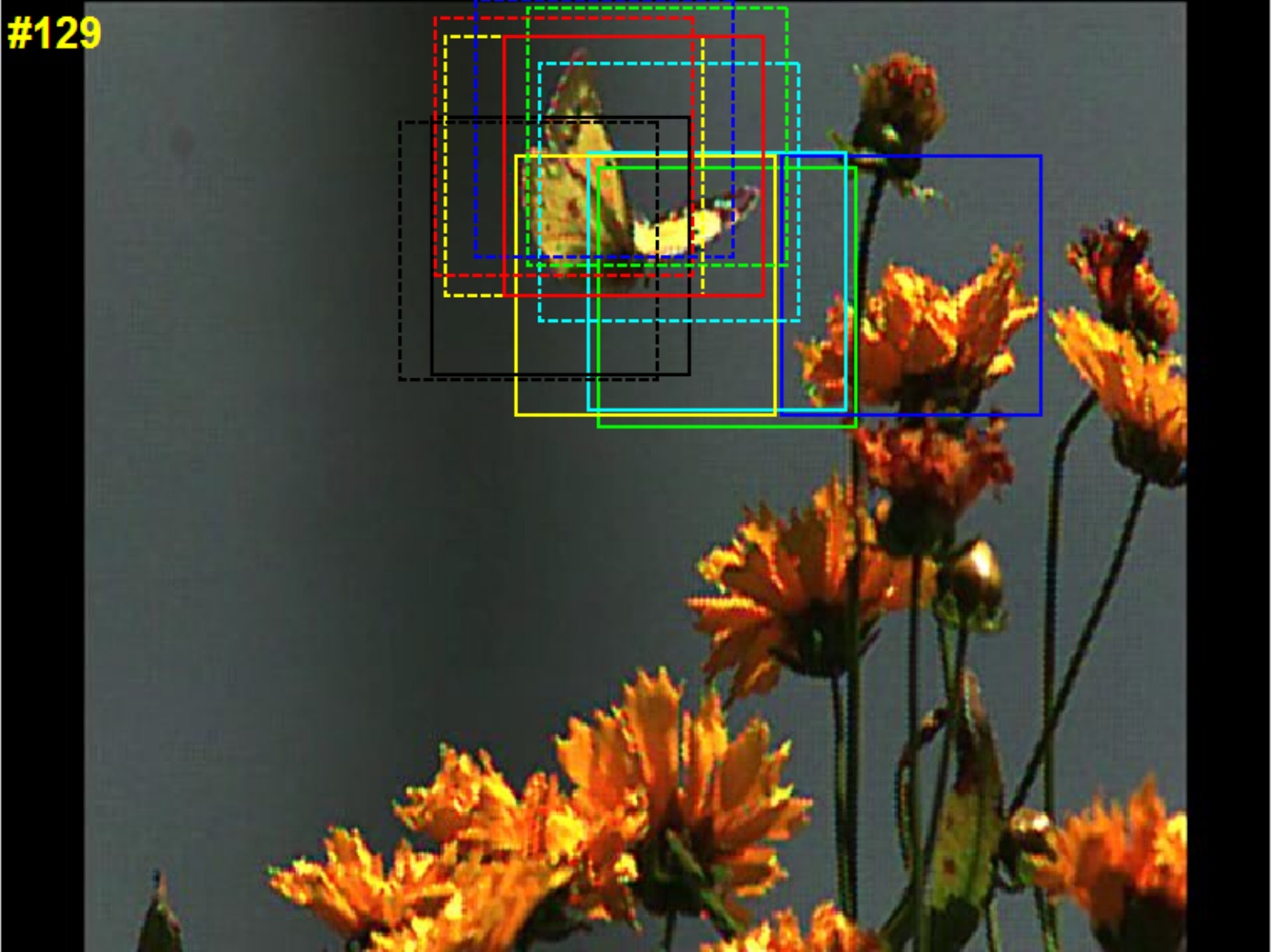}\end{subfigure} &
			\begin{subfigure}{0.2\textwidth}\centering\includegraphics[height=2.8cm, width=2.8cm]{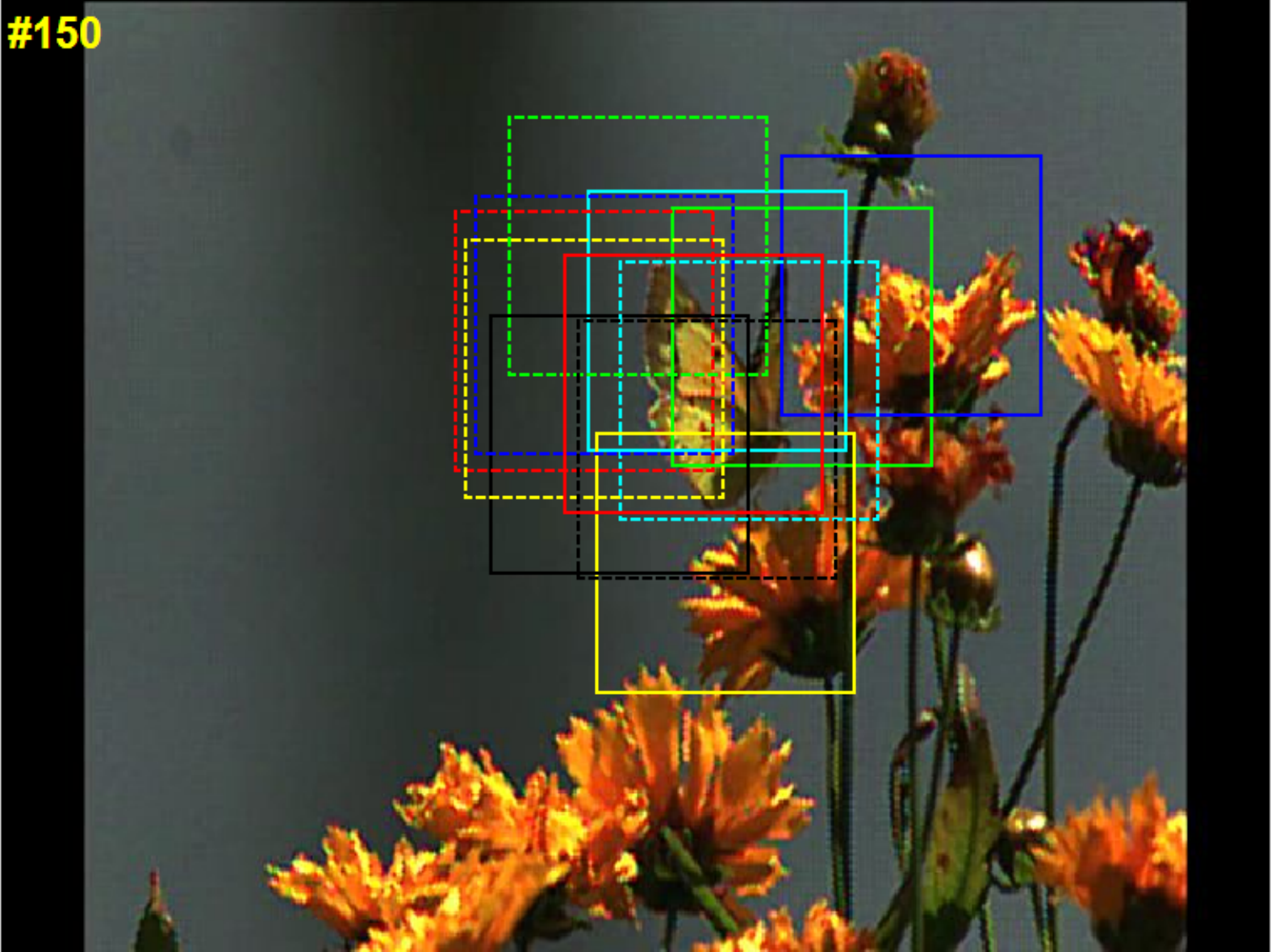}\end{subfigure} \\
			\multicolumn{5}{c}{(b) \textcolor{blue}{Butterfly}} \\
			\begin{subfigure}{0.2\textwidth}\centering\includegraphics[height=2.8cm, width=2.8cm]{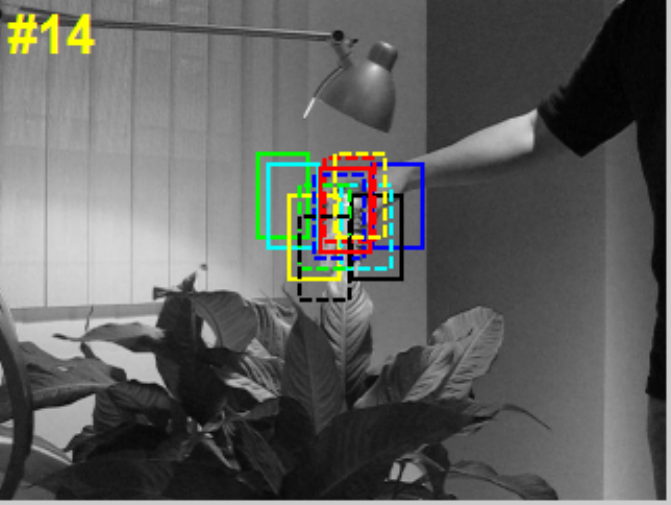}\end{subfigure}&
			\begin{subfigure}{0.2\textwidth}\centering\includegraphics[height=2.8cm, width=2.8cm]{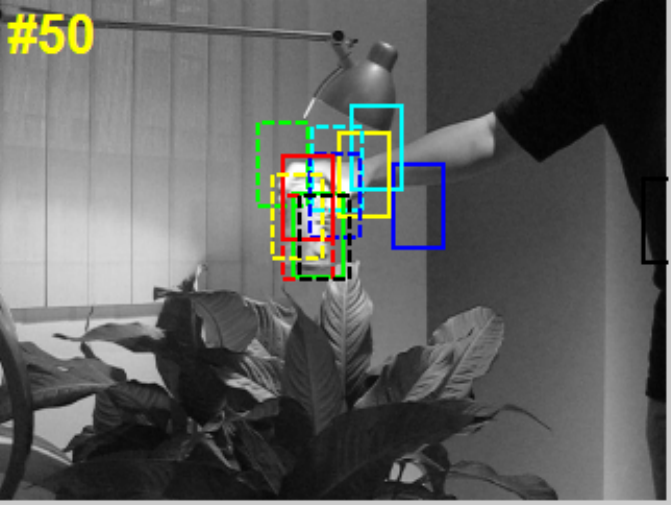}\end{subfigure}&
			\begin{subfigure}{0.2\textwidth}\centering\includegraphics[height=2.8cm, width=2.8cm]{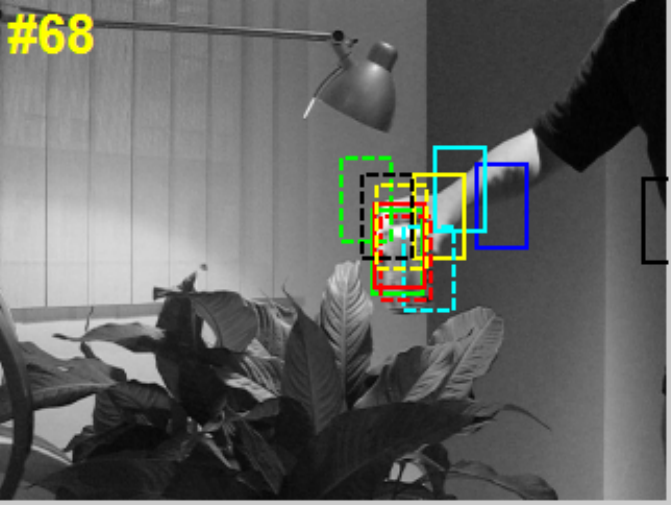}\end{subfigure} &
			\begin{subfigure}{0.2\textwidth}\centering\includegraphics[height=2.8cm, width=2.8cm]{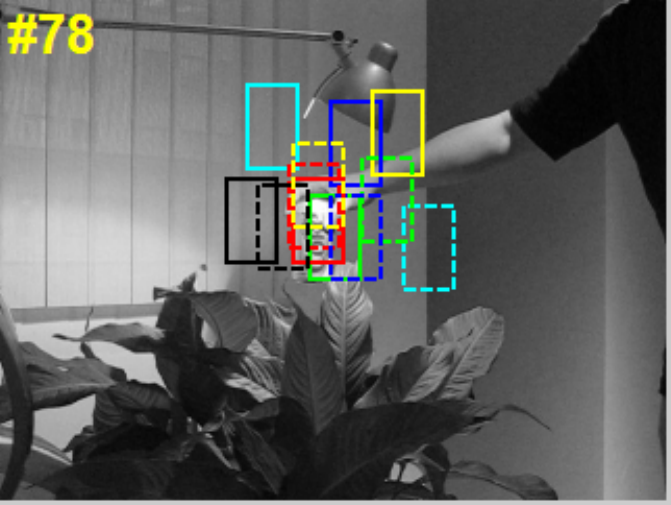}\end{subfigure} &
			\begin{subfigure}{0.2\textwidth}\centering\includegraphics[height=2.8cm, width=2.8cm]{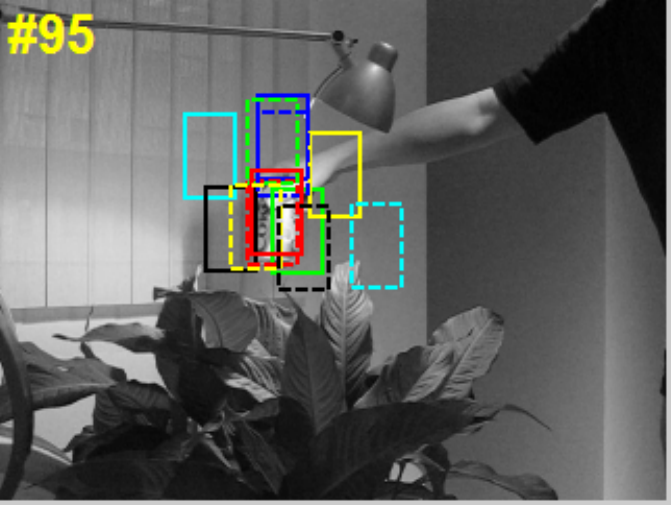}\end{subfigure} \\
			\multicolumn{5}{c}{(c) Coke1} \\
			\begin{subfigure}{0.2\textwidth}\centering\includegraphics[height=2.8cm, width=2.8cm]{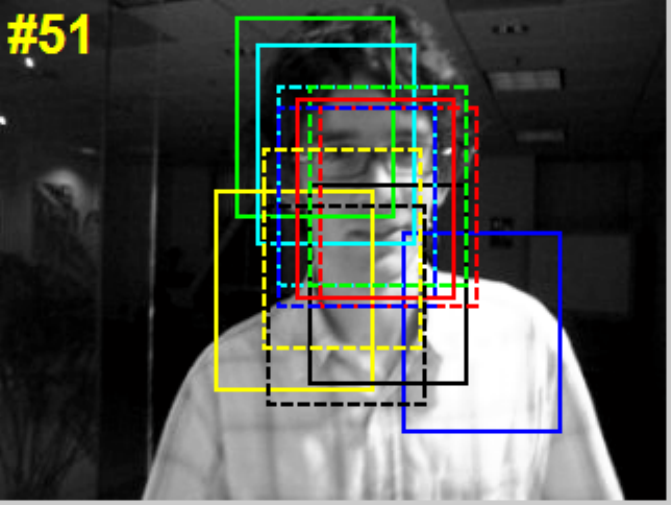}\end{subfigure}&
			\begin{subfigure}{0.2\textwidth}\centering\includegraphics[height=2.8cm, width=2.8cm]{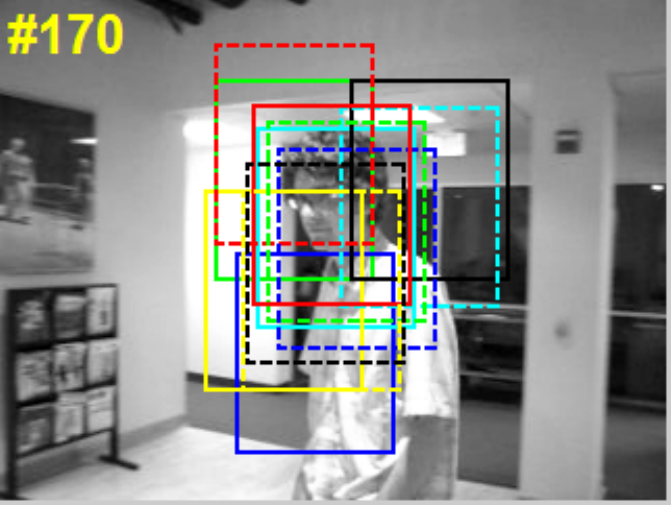}\end{subfigure}&
			\begin{subfigure}{0.2\textwidth}\centering\includegraphics[height=2.8cm, width=2.8cm]{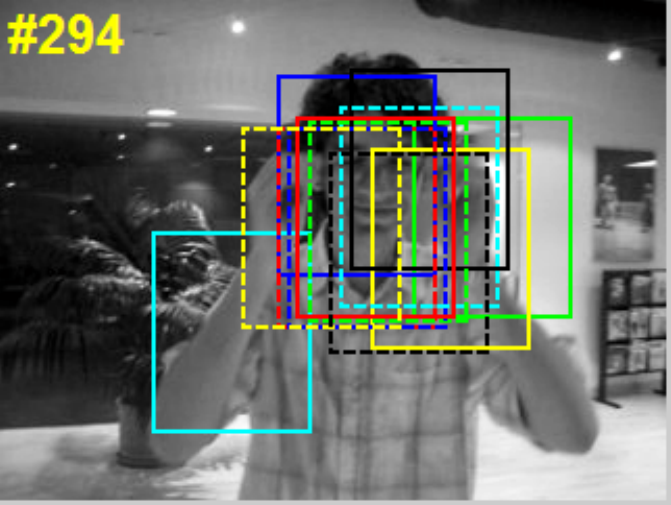}\end{subfigure} &
			\begin{subfigure}{0.2\textwidth}\centering\includegraphics[height=2.8cm, width=2.8cm]{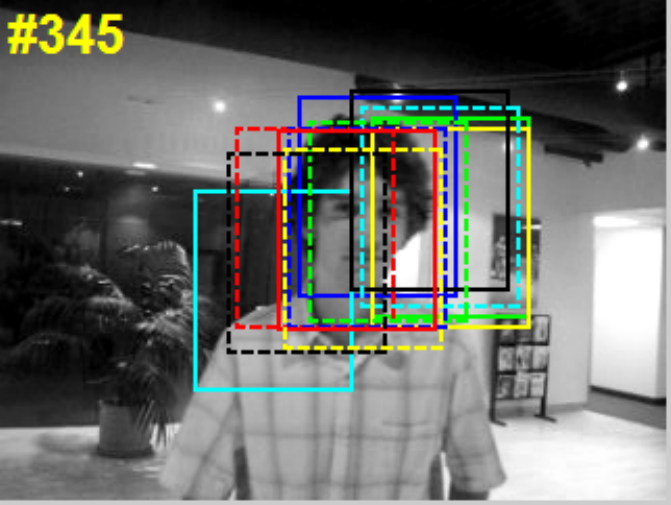}\end{subfigure} &
			\begin{subfigure}{0.2\textwidth}\centering\includegraphics[height=2.8cm, width=2.8cm]{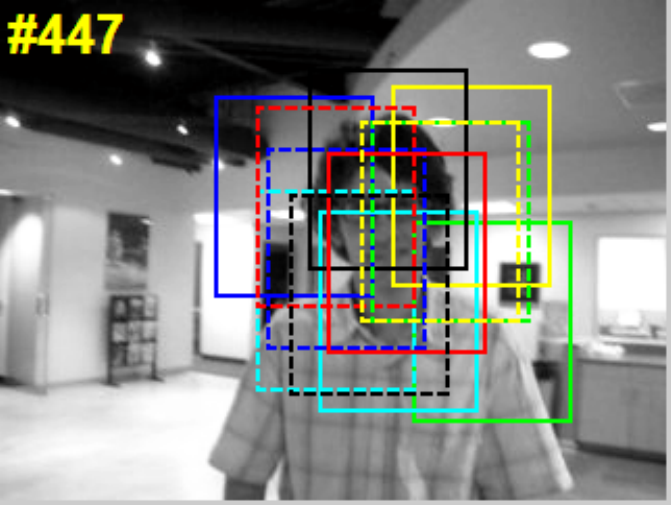}\end{subfigure} \\
			\multicolumn{5}{c}{(d) David} \\
			\begin{subfigure}{0.2\textwidth}\centering\includegraphics[height=2.8cm, width=2.8cm]{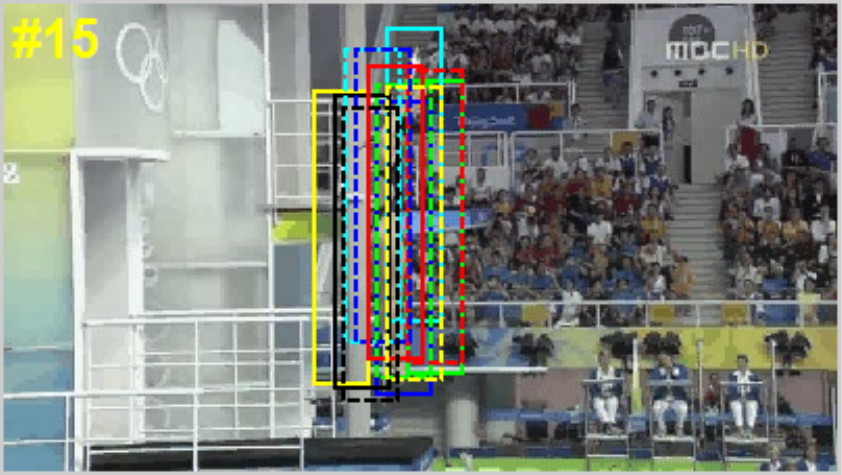}\end{subfigure}&
			\begin{subfigure}{0.2\textwidth}\centering\includegraphics[height=2.8cm, width=2.8cm]{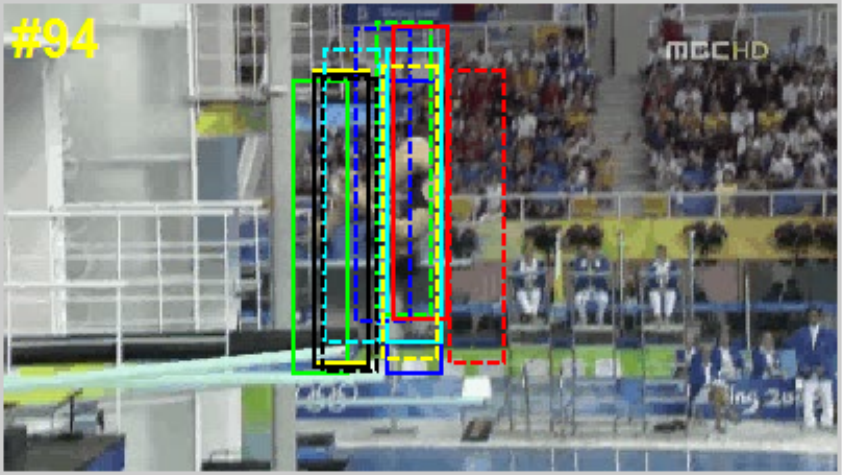}\end{subfigure}&
			\begin{subfigure}{0.2\textwidth}\centering\includegraphics[height=2.8cm, width=2.8cm]{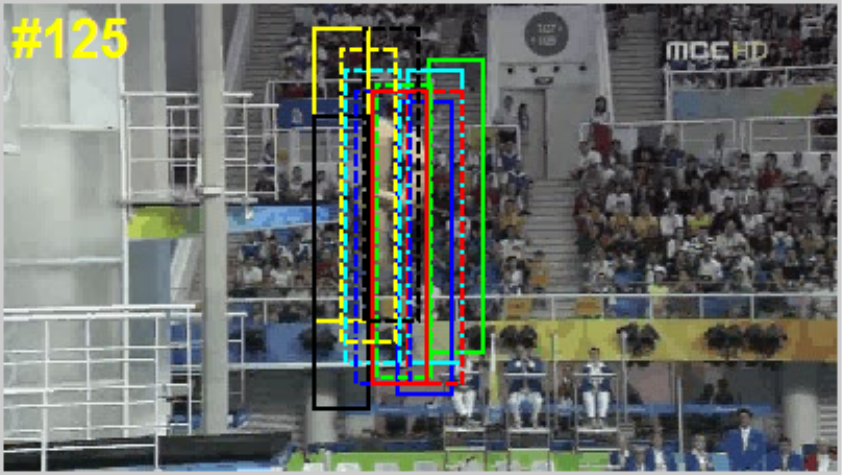}\end{subfigure} &
			\begin{subfigure}{0.2\textwidth}\centering\includegraphics[height=2.8cm, width=2.8cm]{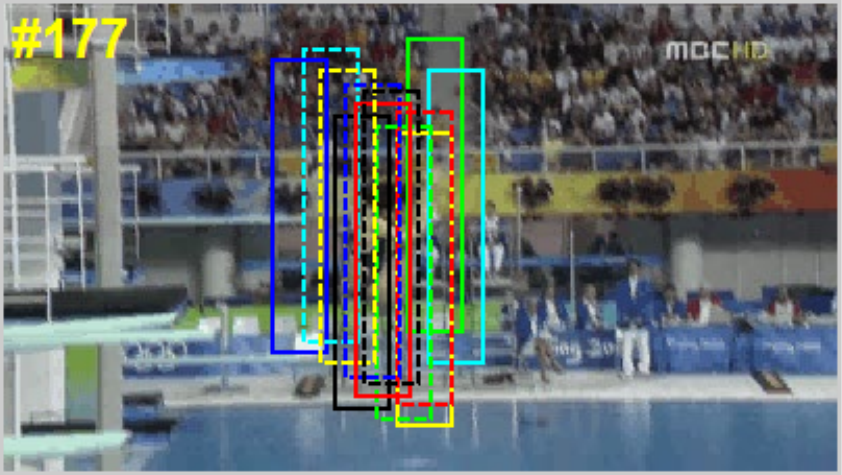}\end{subfigure} &
			\begin{subfigure}{0.2\textwidth}\centering\includegraphics[height=2.8cm, width=2.8cm]{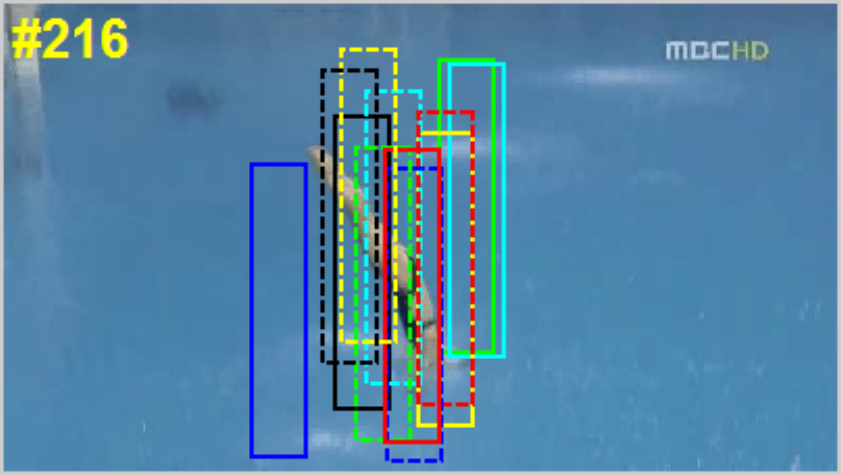}\end{subfigure} \\
			\multicolumn{5}{c}{(e) Diving} \\
			\begin{subfigure}{0.2\textwidth}\centering\includegraphics[height=2.8cm, width=2.8cm]{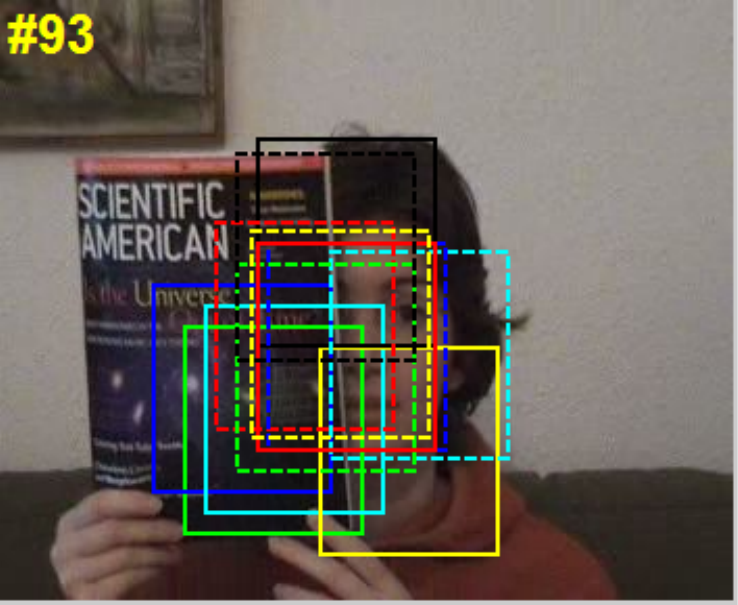}\end{subfigure}&
			\begin{subfigure}{0.2\textwidth}\centering\includegraphics[height=2.8cm, width=2.8cm]{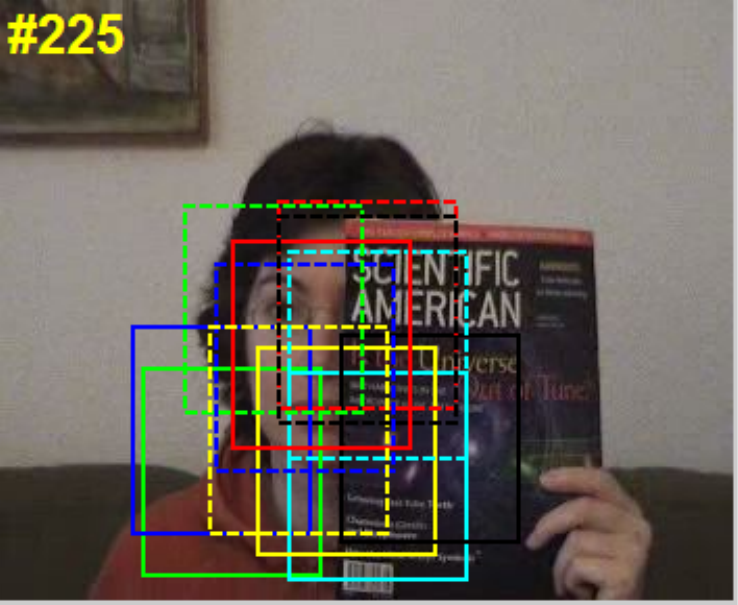}\end{subfigure}&
			\begin{subfigure}{0.2\textwidth}\centering\includegraphics[height=2.8cm, width=2.8cm]{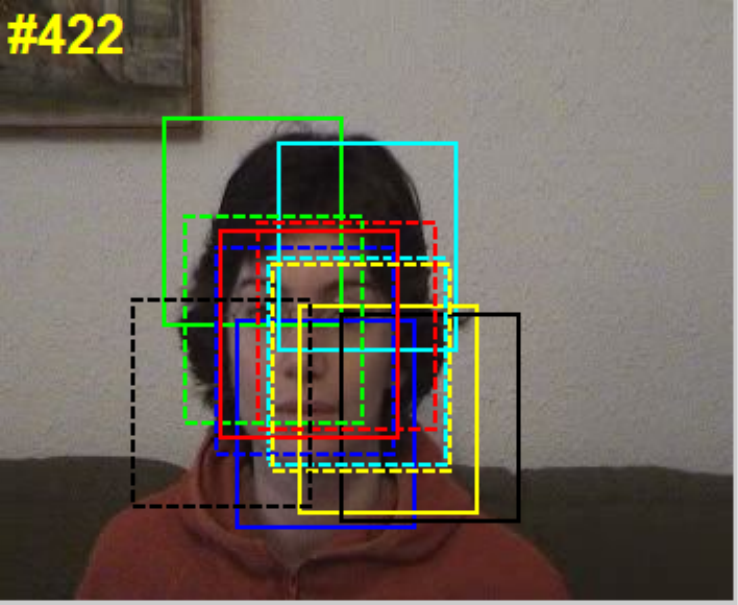}\end{subfigure} &
			\begin{subfigure}{0.2\textwidth}\centering\includegraphics[height=2.8cm, width=2.8cm]{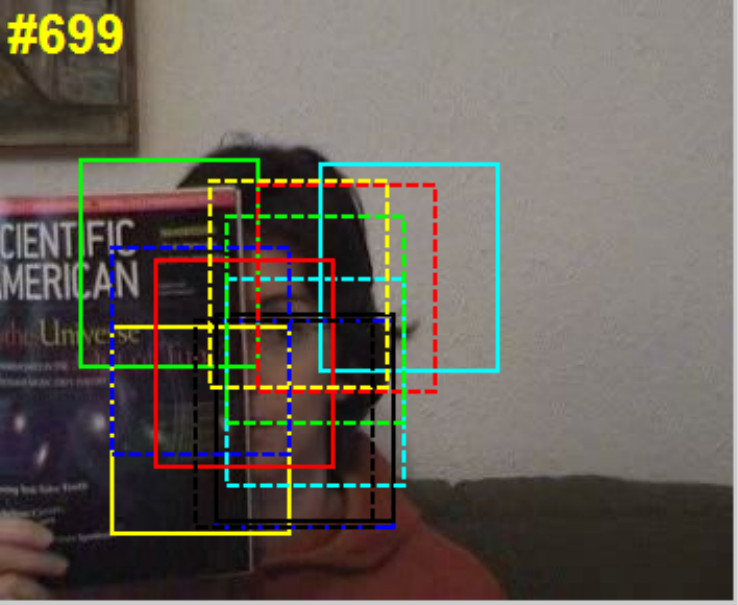}\end{subfigure} &
			\begin{subfigure}{0.2\textwidth}\centering\includegraphics[height=2.8cm, width=2.8cm]{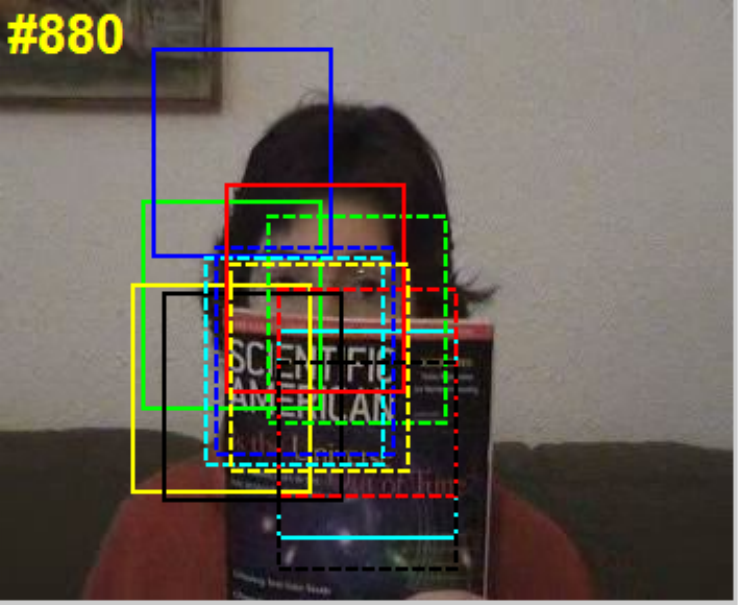}\end{subfigure} \\
			\multicolumn{5}{c}{(f) Occluded face1} 
		\end{tabular}
	}
	\begin{subfigure}{1.0\textwidth}\hspace{.15cm}\centering\includegraphics[height=1cm, width=11.5cm]{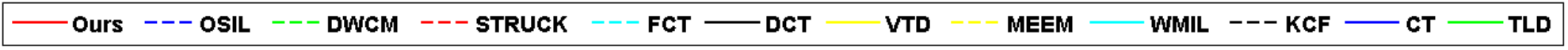}\end{subfigure}
	\caption{
		\label{output1}
		Screenshots of some sample tracking results, from left to right and top to bottom. For clarity, we only draw the tracking results of \textcolor{blue}{12} high performing trackers.} 
\end{figure*}

\begin{figure*}
	\resizebox{.987\textwidth}{!}{
		\begin{tabular}{ccccc}
			\begin{subfigure}{0.2\textwidth}\centering\includegraphics[height=2.8cm, width=2.8cm]{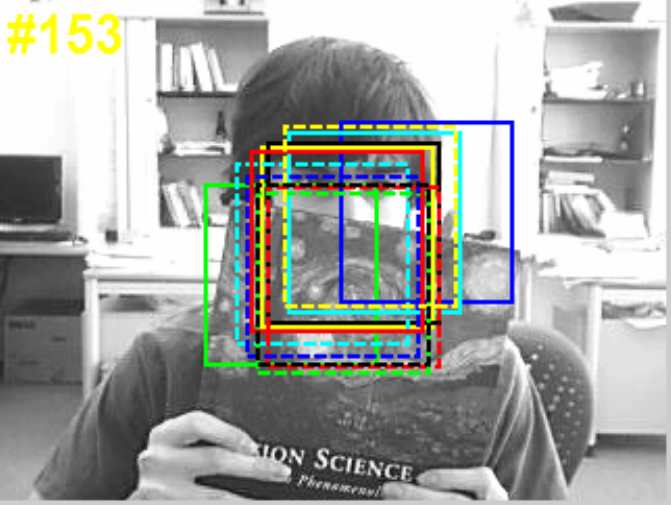}\end{subfigure}&
			\begin{subfigure}{0.2\textwidth}\centering\includegraphics[height=2.8cm, width=2.8cm]{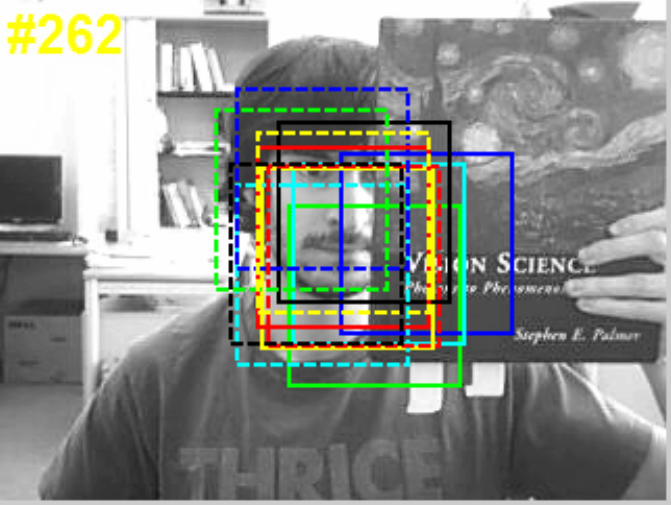}\end{subfigure}&
			\begin{subfigure}{0.2\textwidth}\centering\includegraphics[height=2.8cm, width=2.8cm]{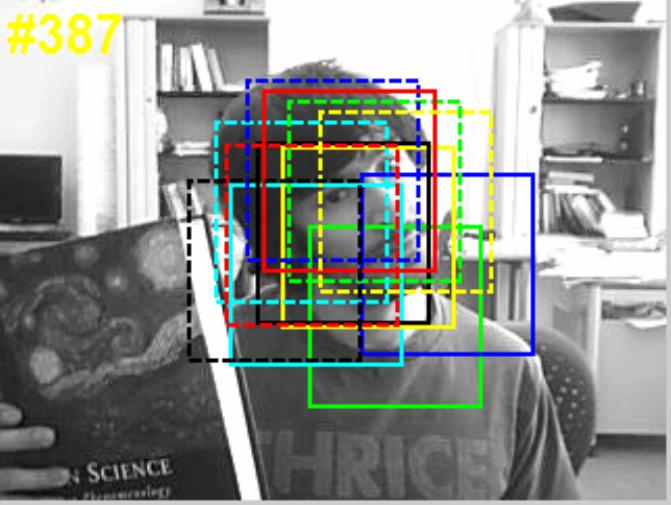}\end{subfigure} &
			\begin{subfigure}{0.2\textwidth}\centering\includegraphics[height=2.8cm, width=2.8cm]{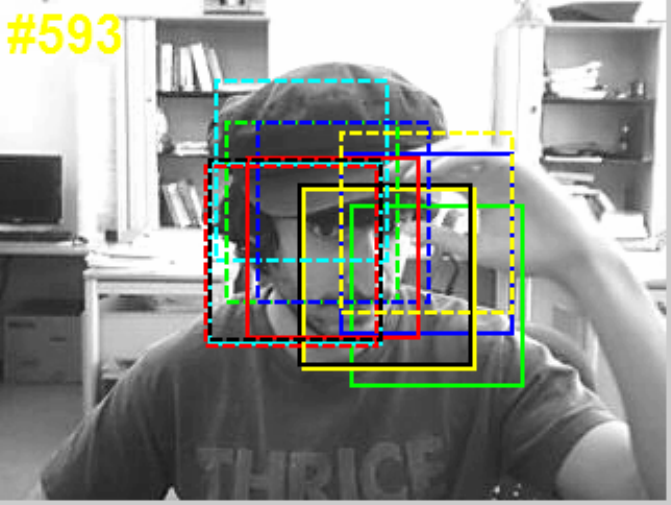}\end{subfigure} &
			\begin{subfigure}{0.2\textwidth}\centering\includegraphics[height=2.8cm, width=2.8cm]{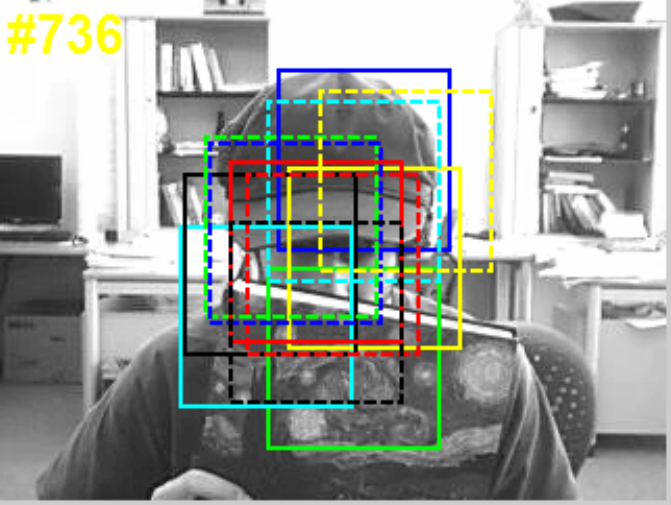}\end{subfigure} \\
			\multicolumn{5}{c}{(a) Occluded face2} \\
			\begin{subfigure}{0.2\textwidth}\centering\includegraphics[height=2.8cm, width=2.8cm]{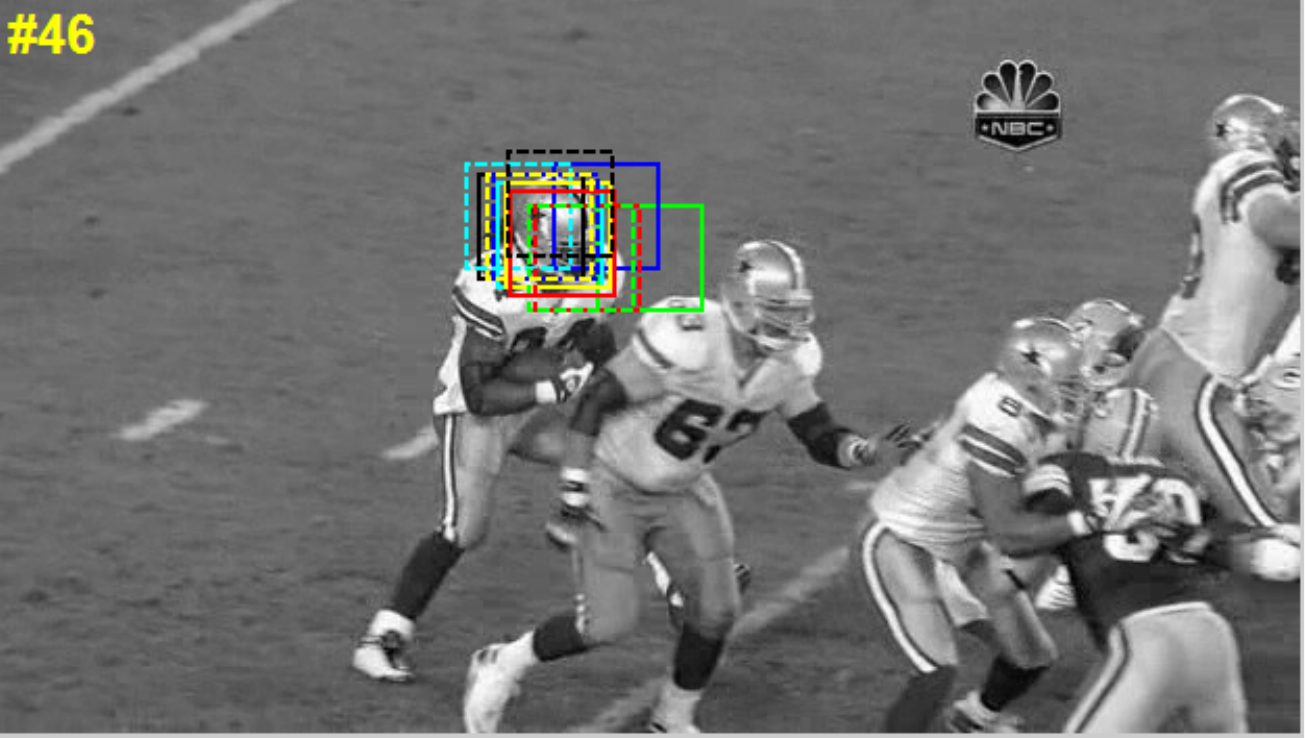}\end{subfigure}&
			\begin{subfigure}{0.2\textwidth}\centering\includegraphics[height=2.8cm, width=2.8cm]{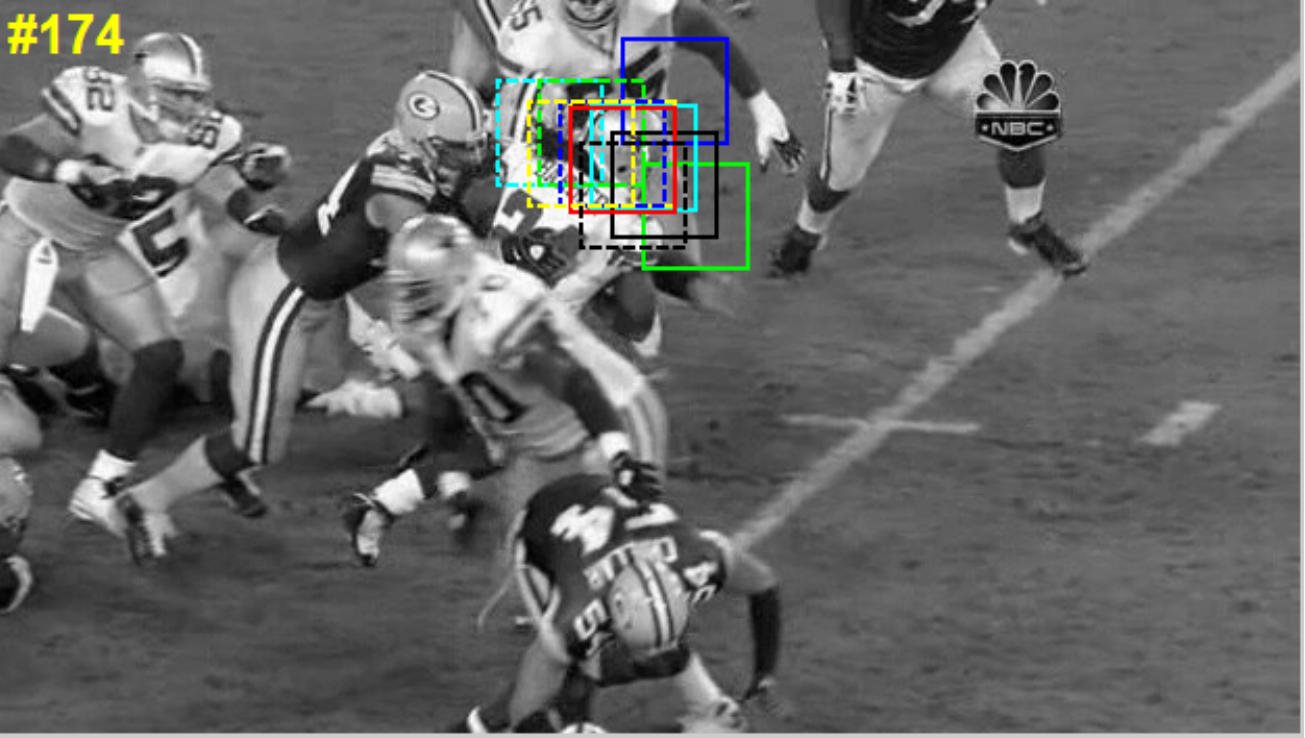}\end{subfigure}&
			\begin{subfigure}{0.2\textwidth}\centering\includegraphics[height=2.8cm, width=2.8cm]{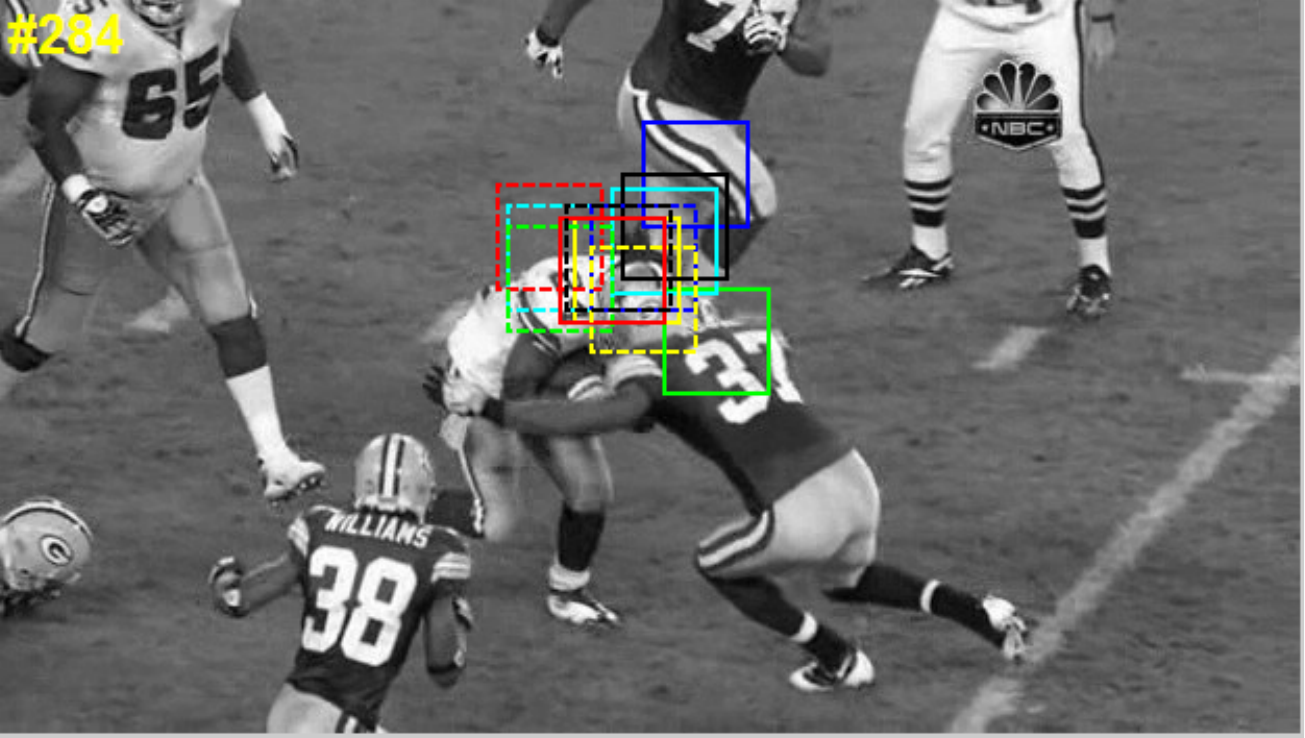}\end{subfigure} &
			\begin{subfigure}{0.2\textwidth}\centering\includegraphics[height=2.8cm, width=2.8cm]{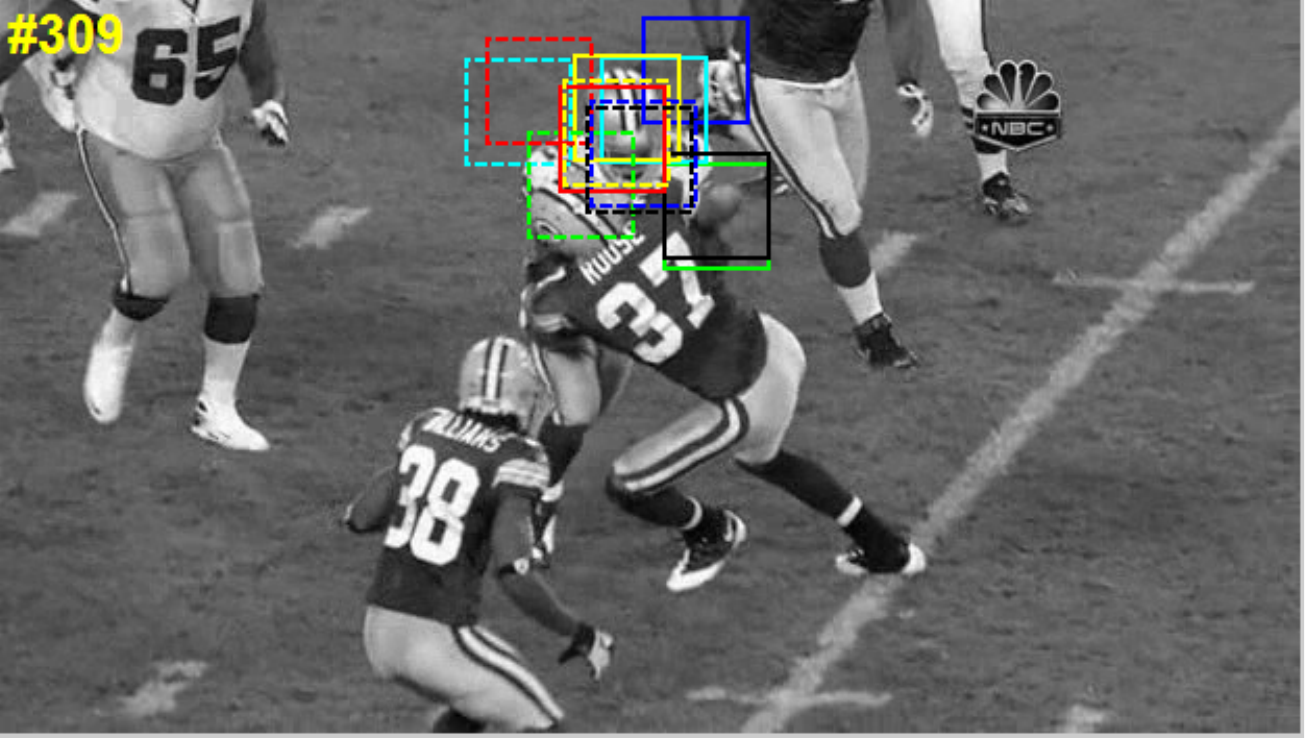}\end{subfigure} &
			\begin{subfigure}{0.2\textwidth}\centering\includegraphics[height=2.8cm, width=2.8cm]{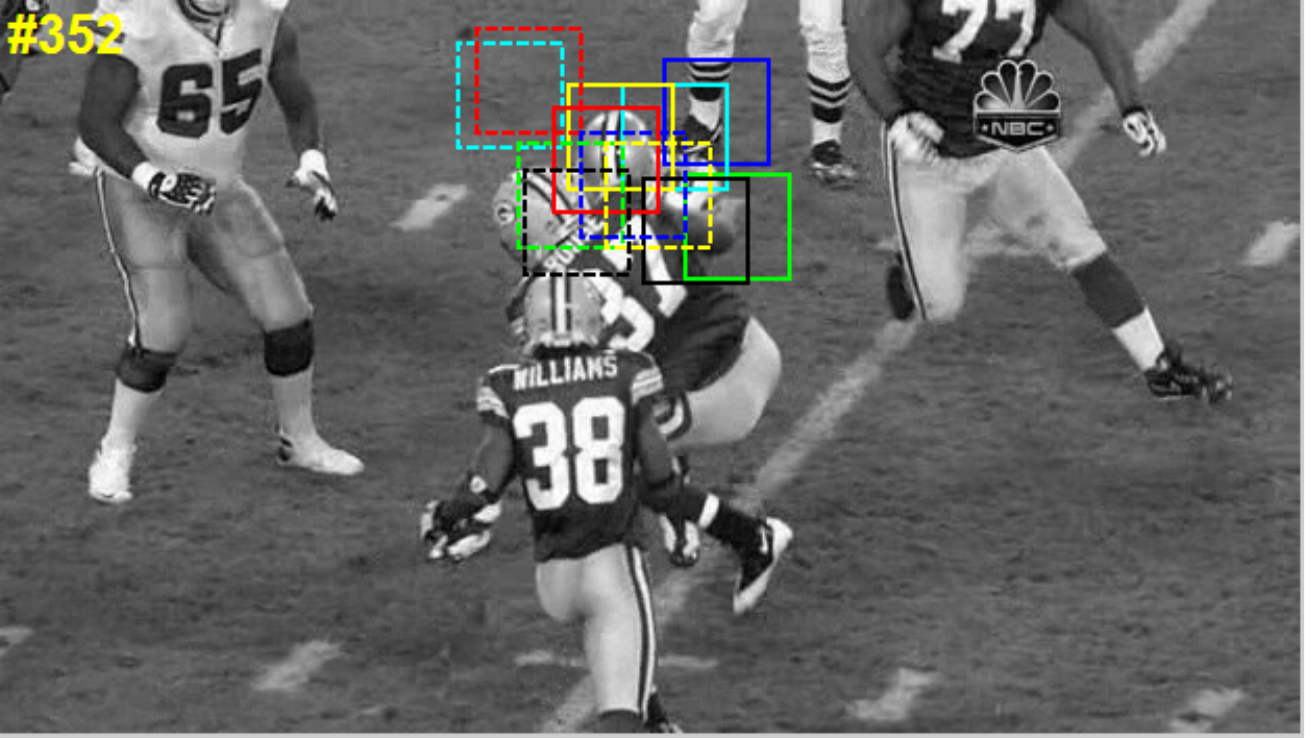}\end{subfigure} 	\\
			\multicolumn{5}{c}{(b) Football} \\
			\begin{subfigure}{0.2\textwidth}\centering\includegraphics[height=2.8cm, width=2.8cm]{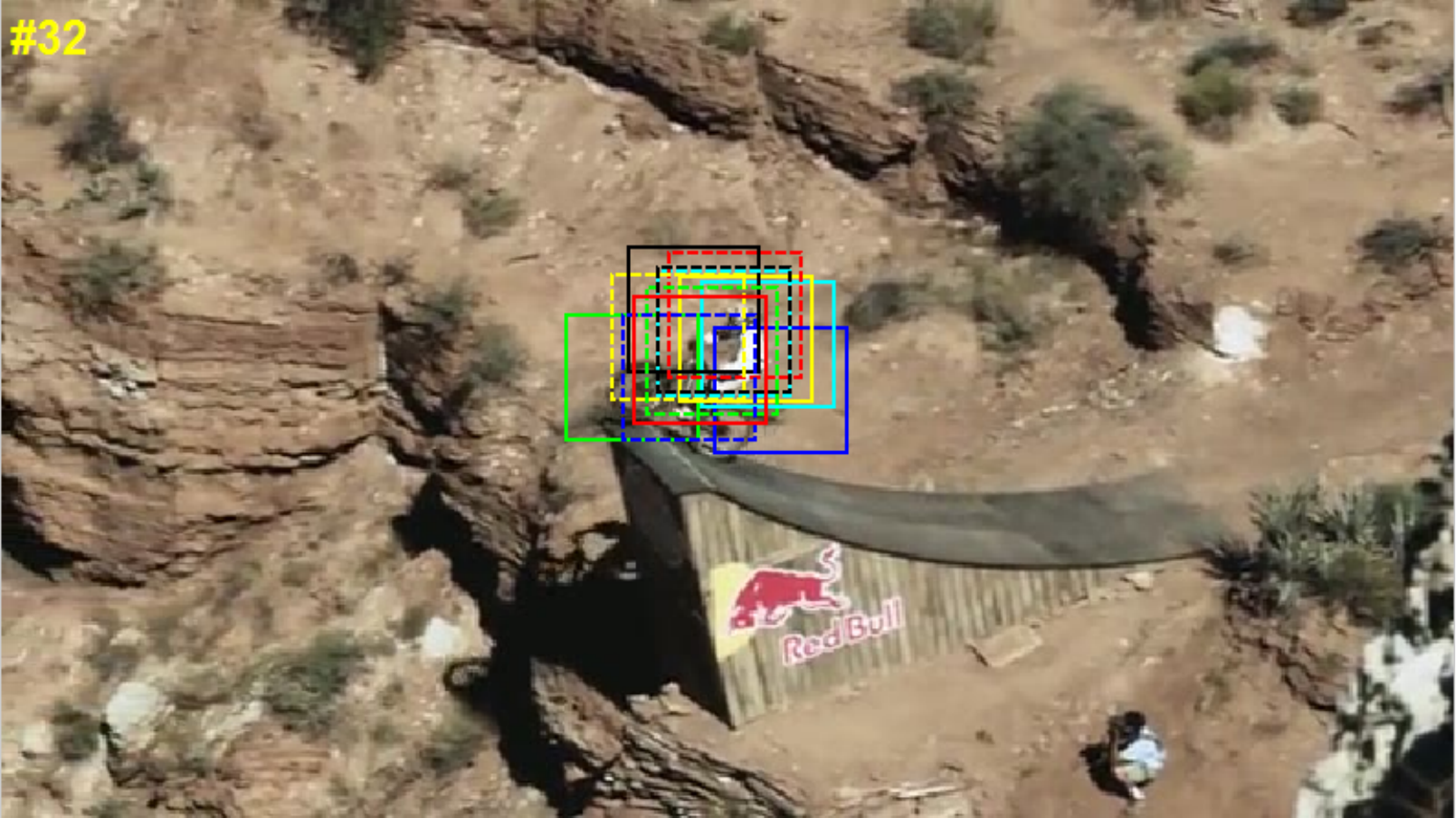}\end{subfigure}&
			\begin{subfigure}{0.2\textwidth}\centering\includegraphics[height=2.8cm, width=2.8cm]{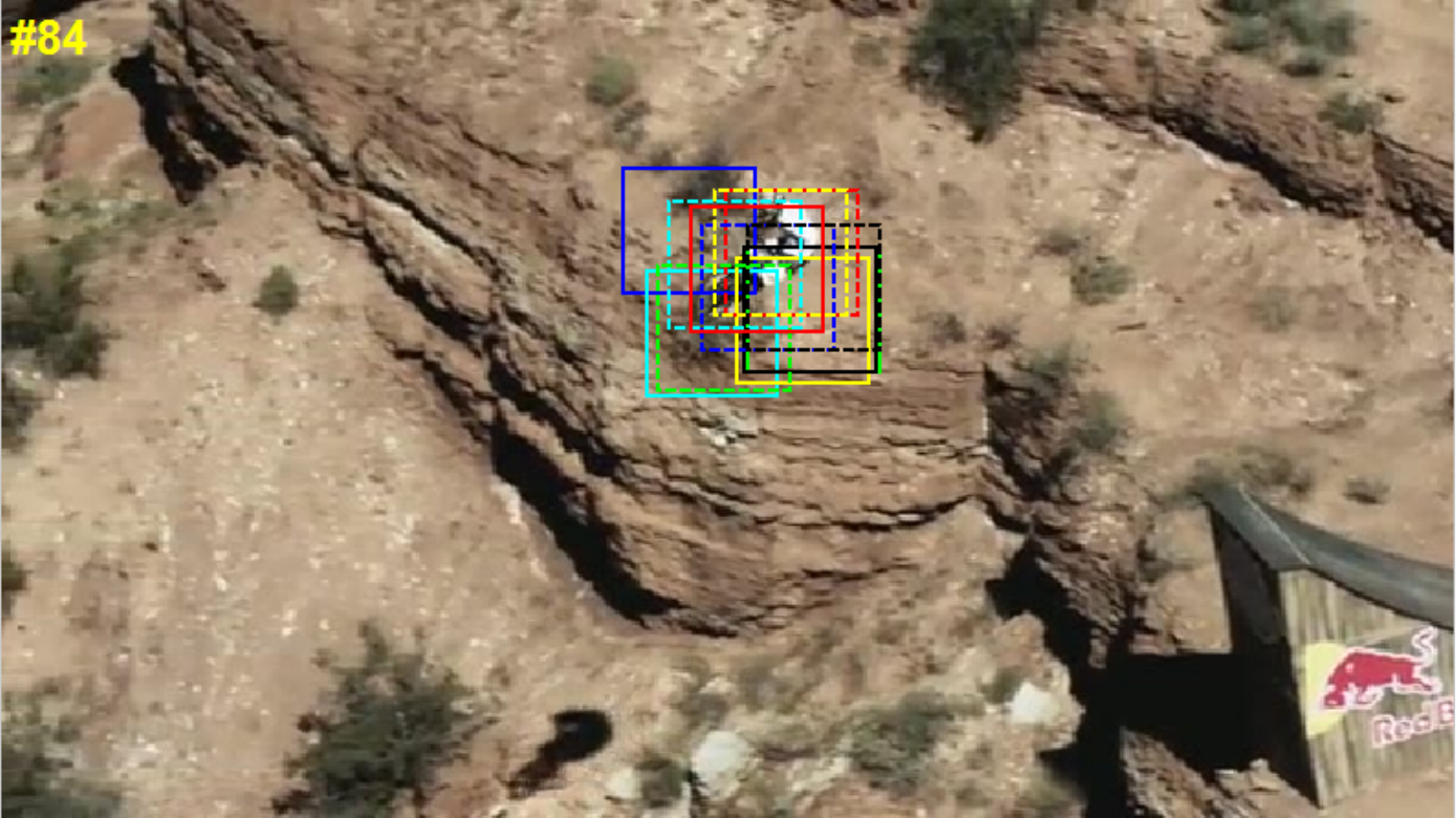}\end{subfigure}&
			\begin{subfigure}{0.2\textwidth}\centering\includegraphics[height=2.8cm, width=2.8cm]{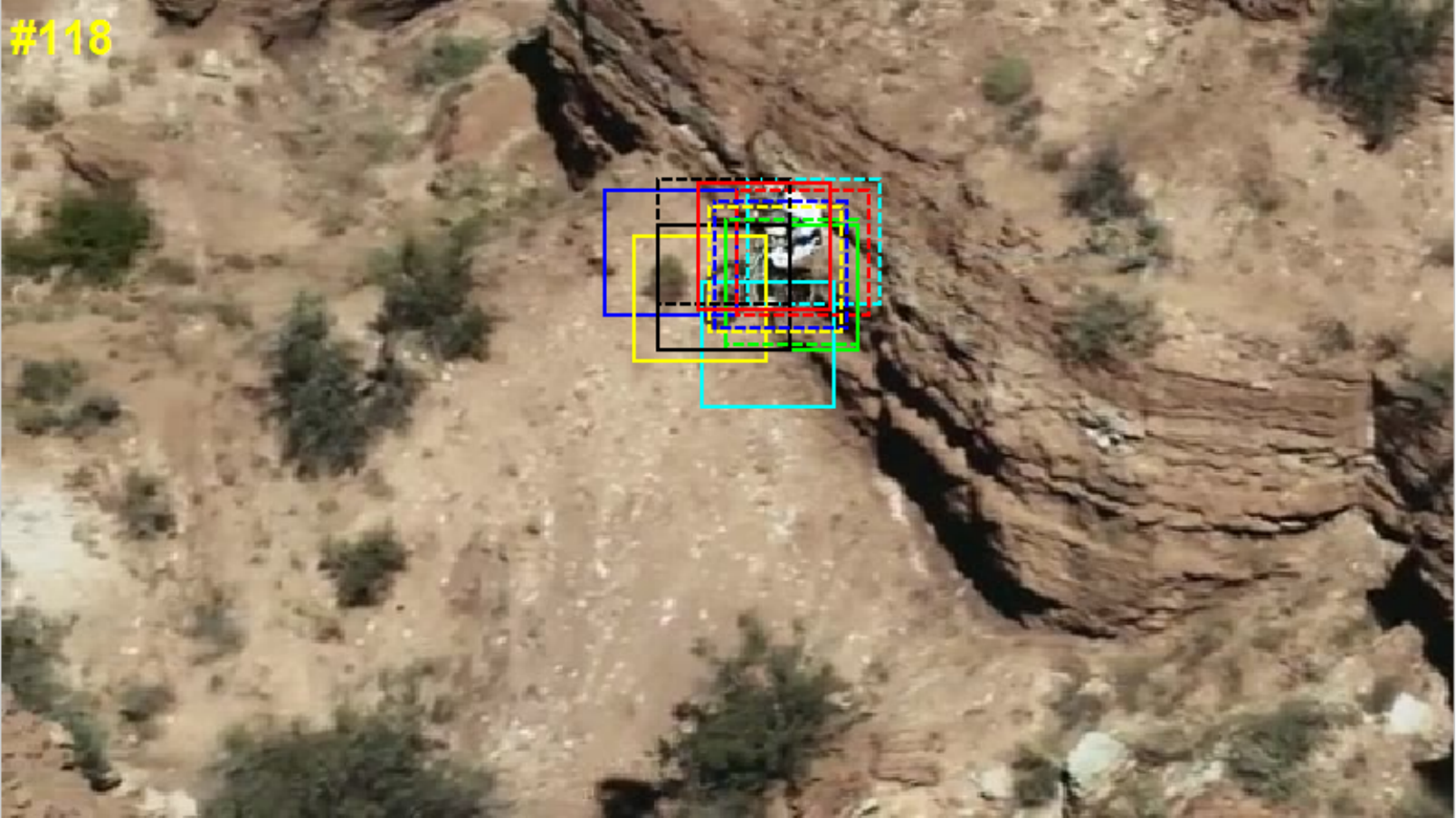}\end{subfigure} &
			\begin{subfigure}{0.2\textwidth}\centering\includegraphics[height=2.8cm, width=2.8cm]{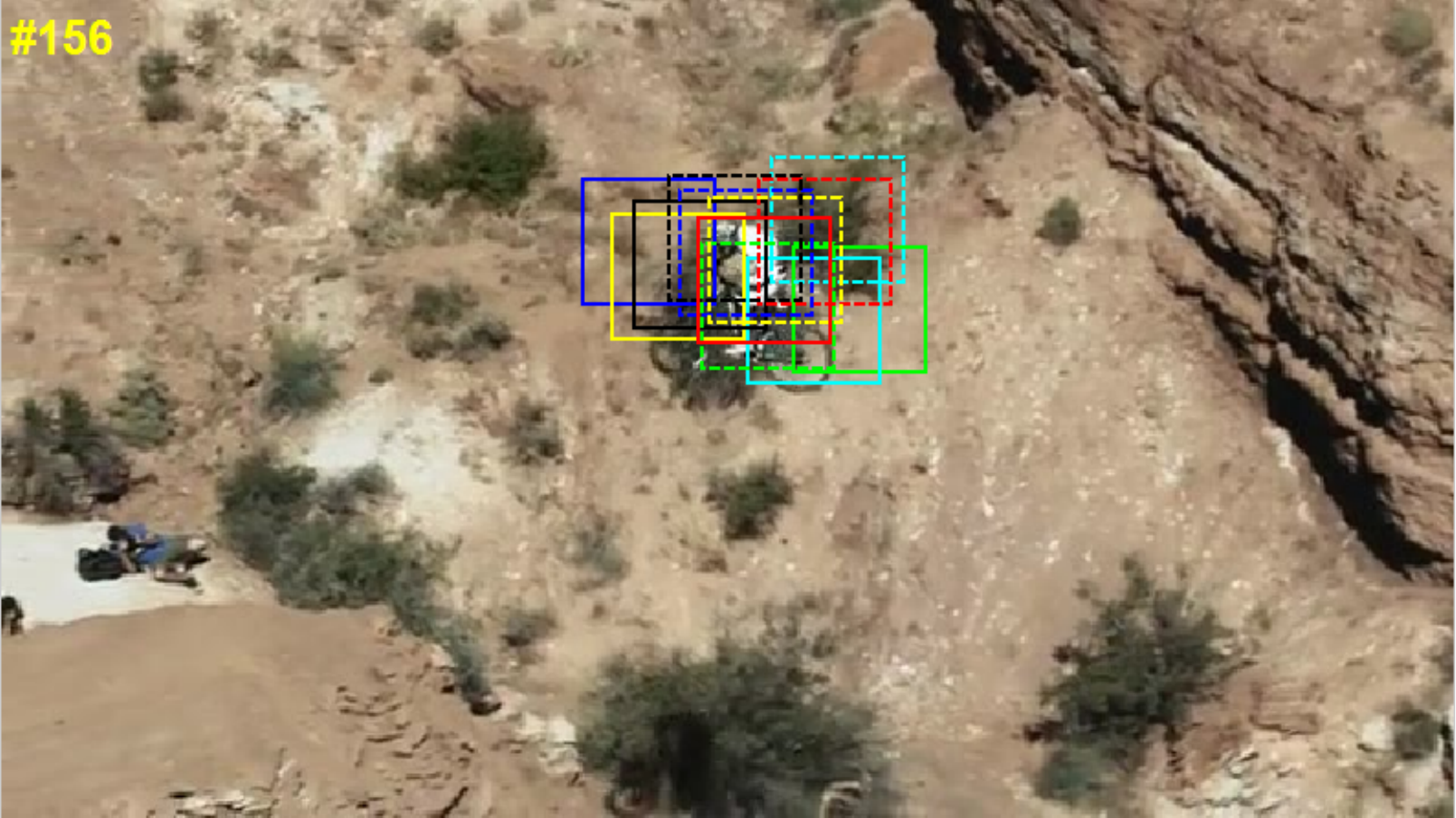}\end{subfigure} &
			\begin{subfigure}{0.2\textwidth}\centering\includegraphics[height=2.8cm, width=2.8cm]{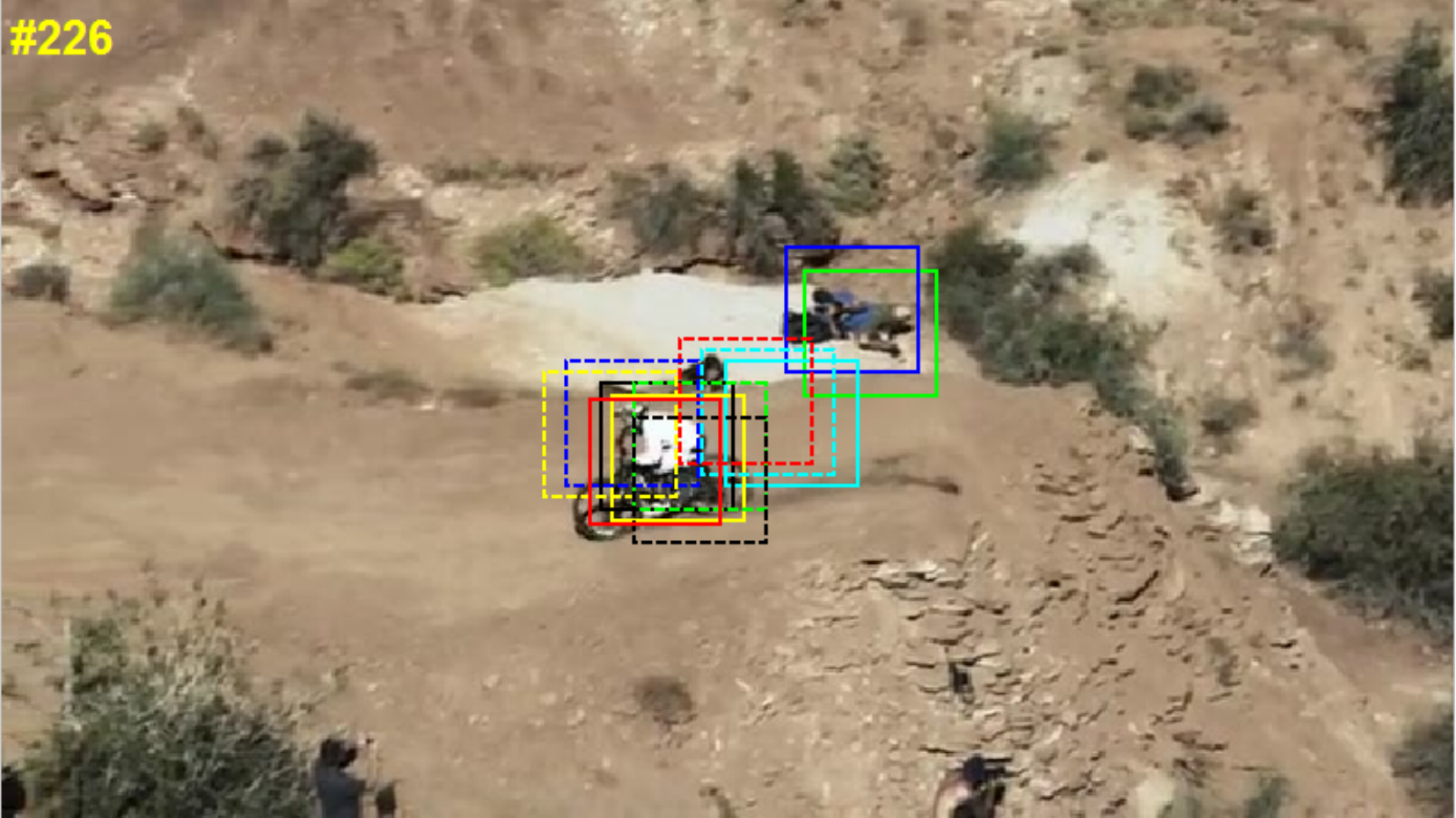}\end{subfigure} \\
			\multicolumn{5}{c}{(c) Mountain-bike} \\
			\begin{subfigure}{0.2\textwidth}\centering\includegraphics[height=2.8cm, width=2.8cm]{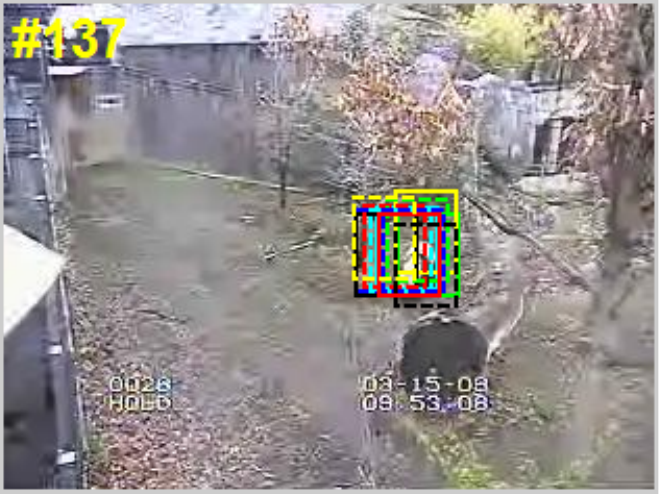}\end{subfigure}&
			\begin{subfigure}{0.2\textwidth}\centering\includegraphics[height=2.8cm, width=2.8cm]{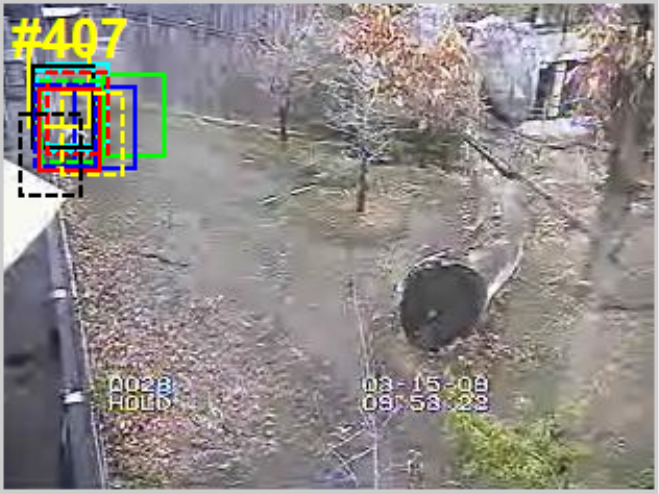}\end{subfigure}&
			\begin{subfigure}{0.2\textwidth}\centering\includegraphics[height=2.8cm, width=2.8cm]{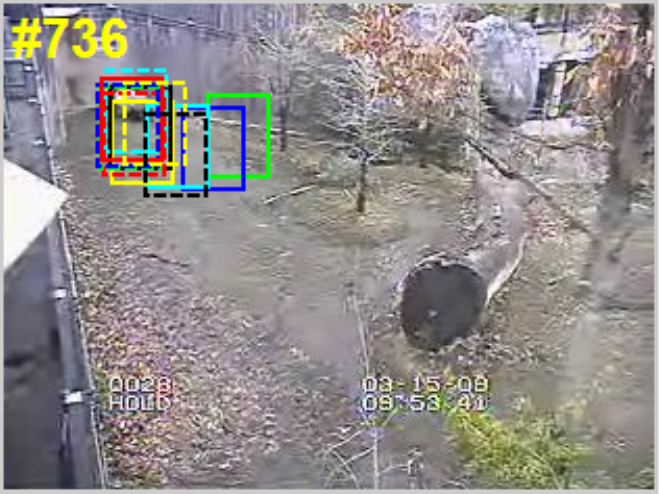}\end{subfigure} &
			\begin{subfigure}{0.2\textwidth}\centering\includegraphics[height=2.8cm, width=2.8cm]{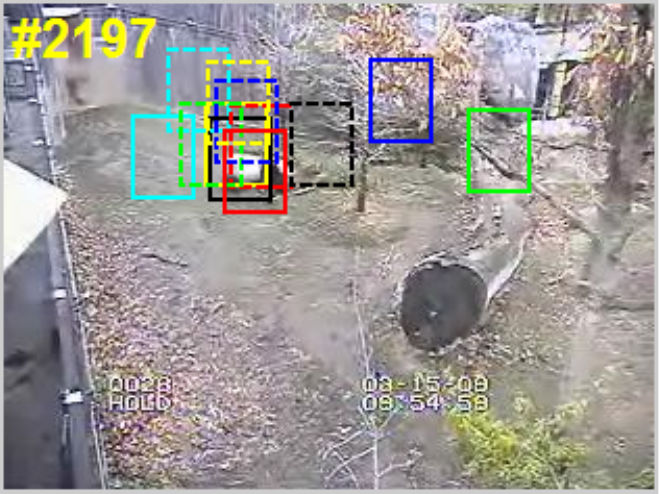}\end{subfigure} &
			\begin{subfigure}{0.2\textwidth}\centering\includegraphics[height=2.8cm, width=2.8cm]{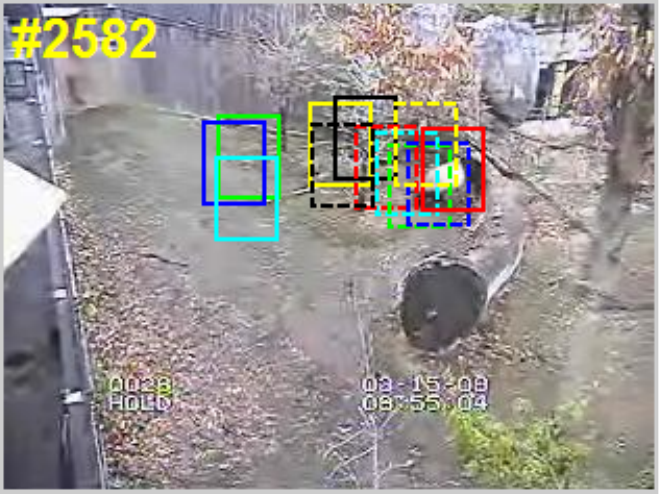}\end{subfigure} \\
			\multicolumn{5}{c}{(d) Panda} \\
			\begin{subfigure}{0.2\textwidth}\centering\includegraphics[height=2.8cm, width=2.8cm]{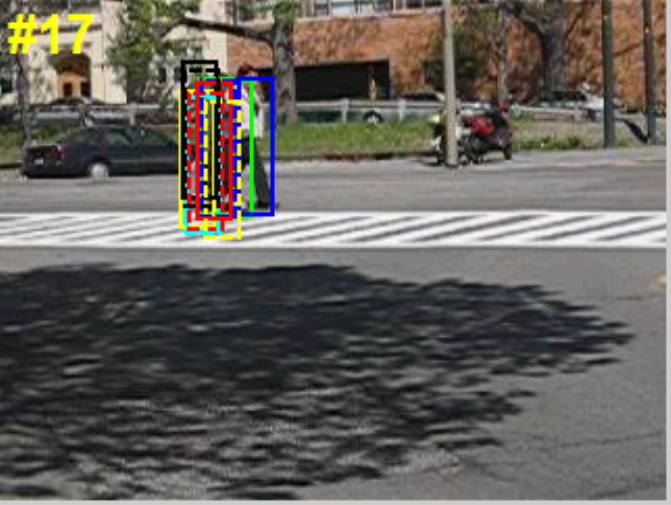}\end{subfigure}&
			\begin{subfigure}{0.2\textwidth}\centering\includegraphics[height=2.8cm, width=2.8cm]{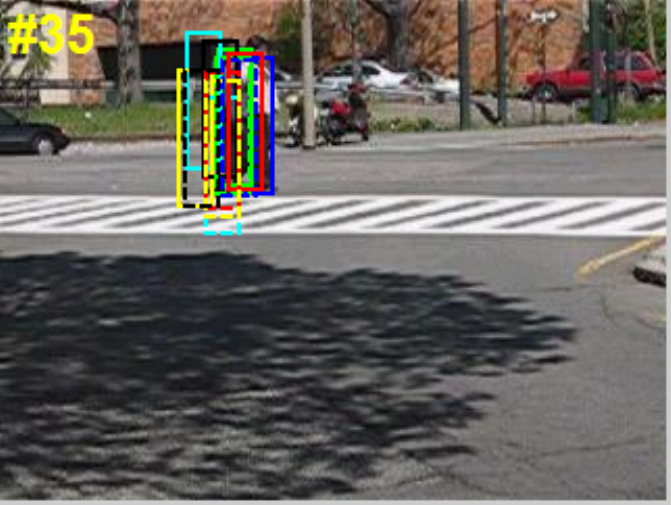}\end{subfigure}&
			\begin{subfigure}{0.2\textwidth}\centering\includegraphics[height=2.8cm, width=2.8cm]{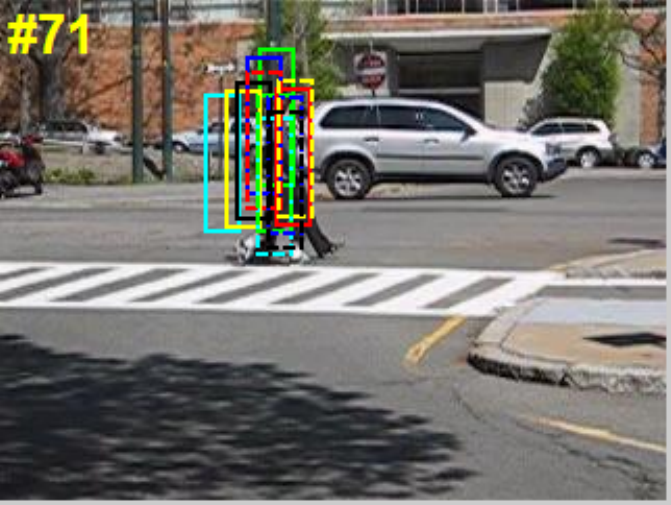}\end{subfigure} &
			\begin{subfigure}{0.2\textwidth}\centering\includegraphics[height=2.8cm, width=2.8cm]{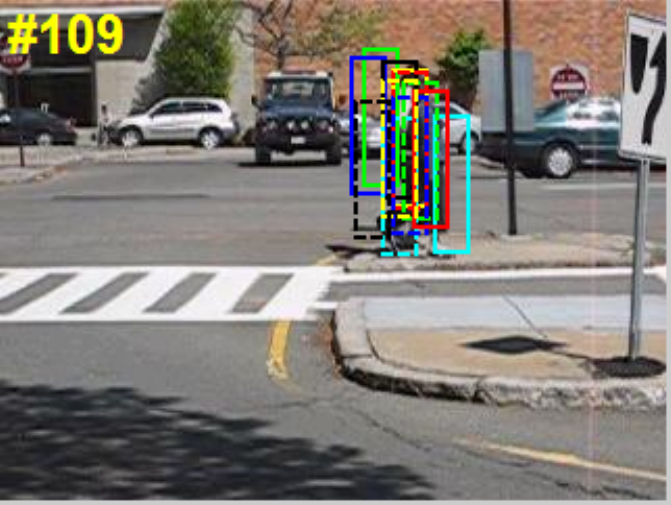}\end{subfigure} &
			\begin{subfigure}{0.2\textwidth}\centering\includegraphics[height=2.8cm, width=2.8cm]{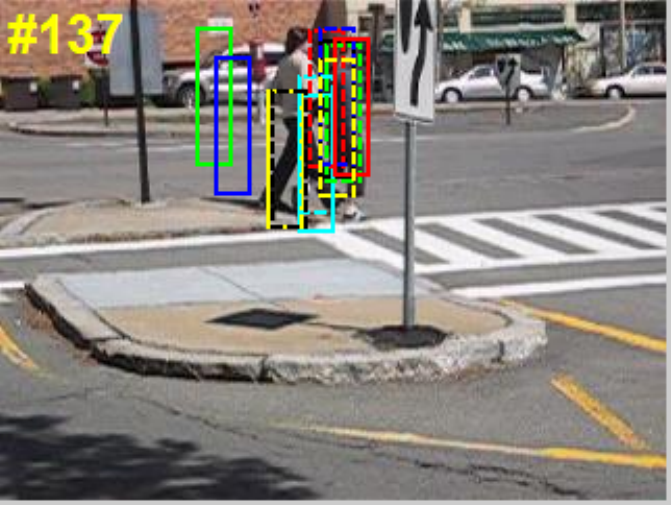}\end{subfigure} \\
			\multicolumn{5}{c}{(e) Pedestrian3} \\
			\begin{subfigure}{0.2\textwidth}\centering\includegraphics[height=2.8cm, width=2.8cm]{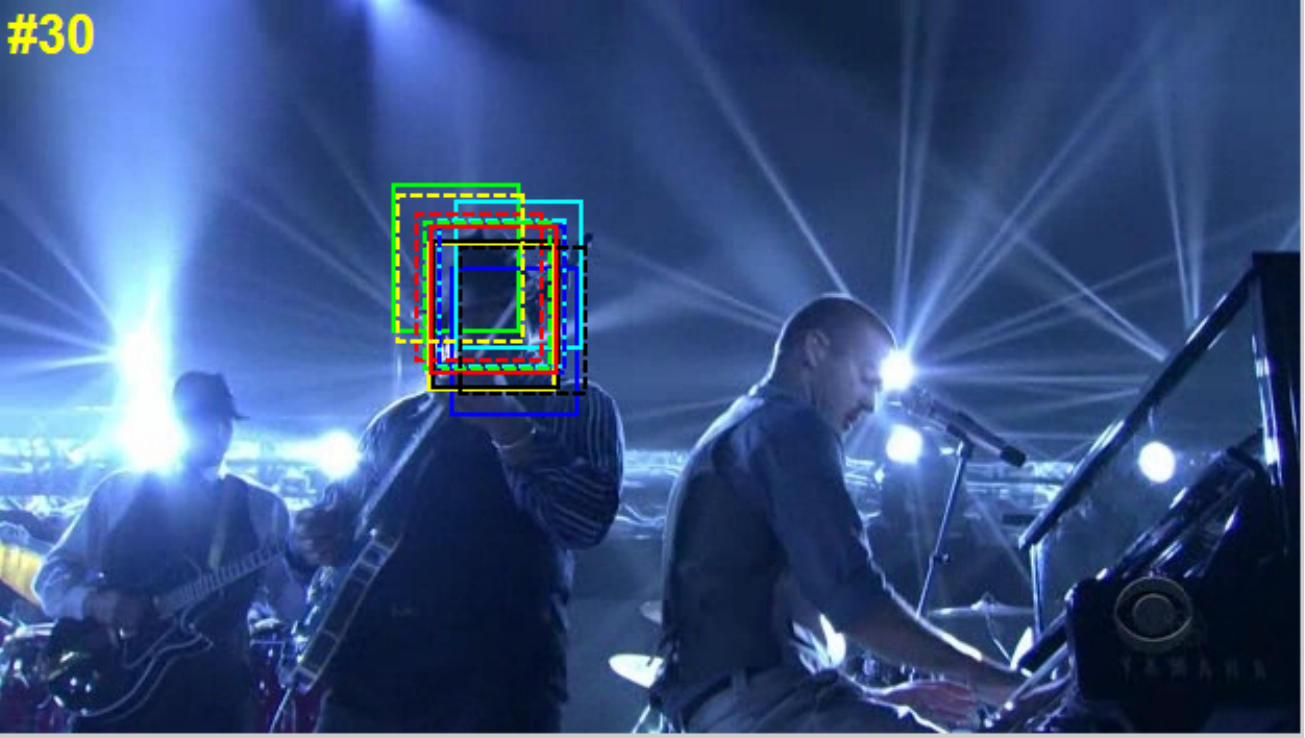}\end{subfigure}&
			\begin{subfigure}{0.2\textwidth}\centering\includegraphics[height=2.8cm, width=2.8cm]{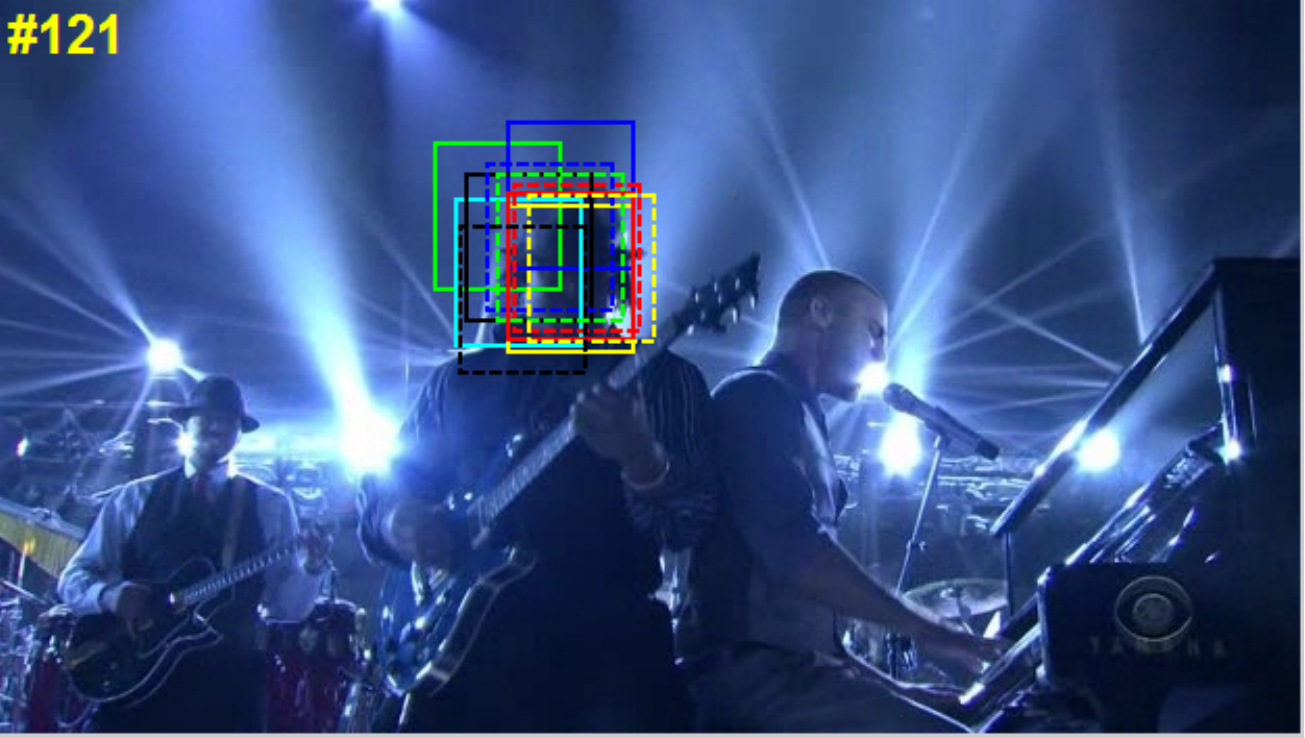}\end{subfigure}&
			\begin{subfigure}{0.2\textwidth}\centering\includegraphics[height=2.8cm, width=2.8cm]{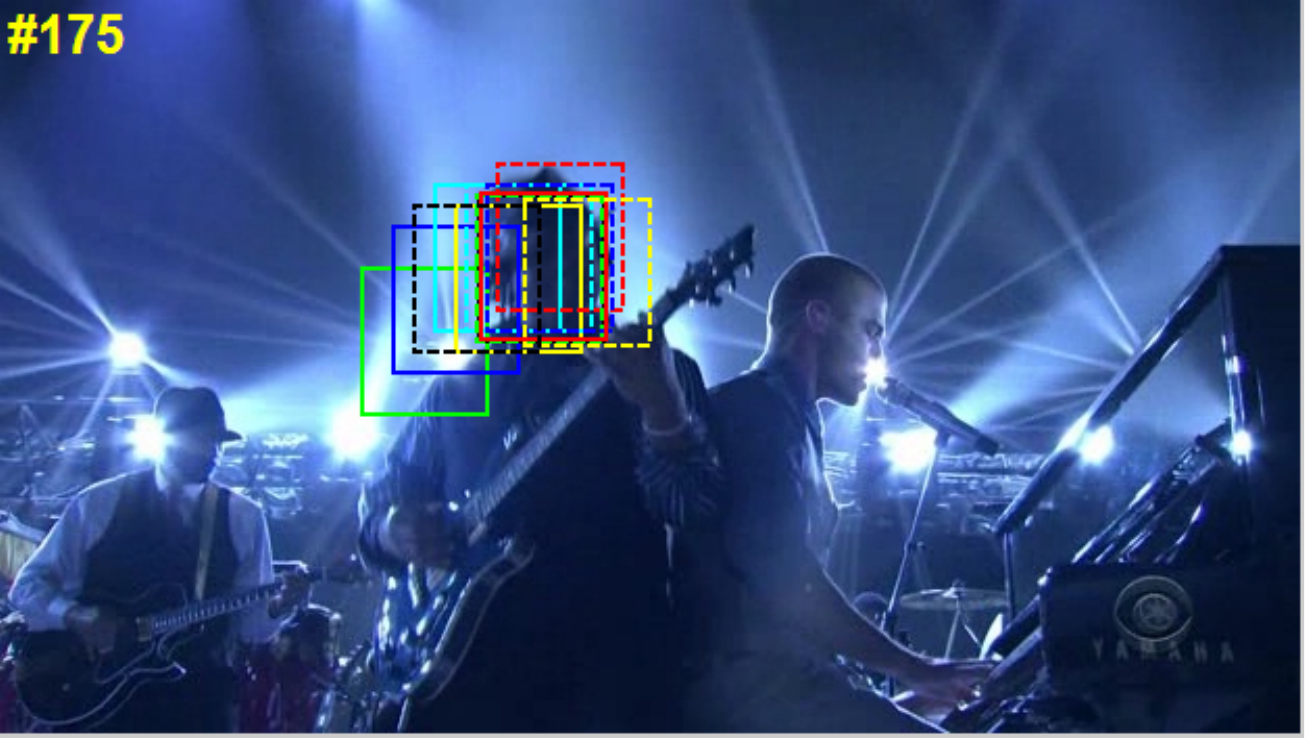}\end{subfigure} &
			\begin{subfigure}{0.2\textwidth}\centering\includegraphics[height=2.8cm, width=2.8cm]{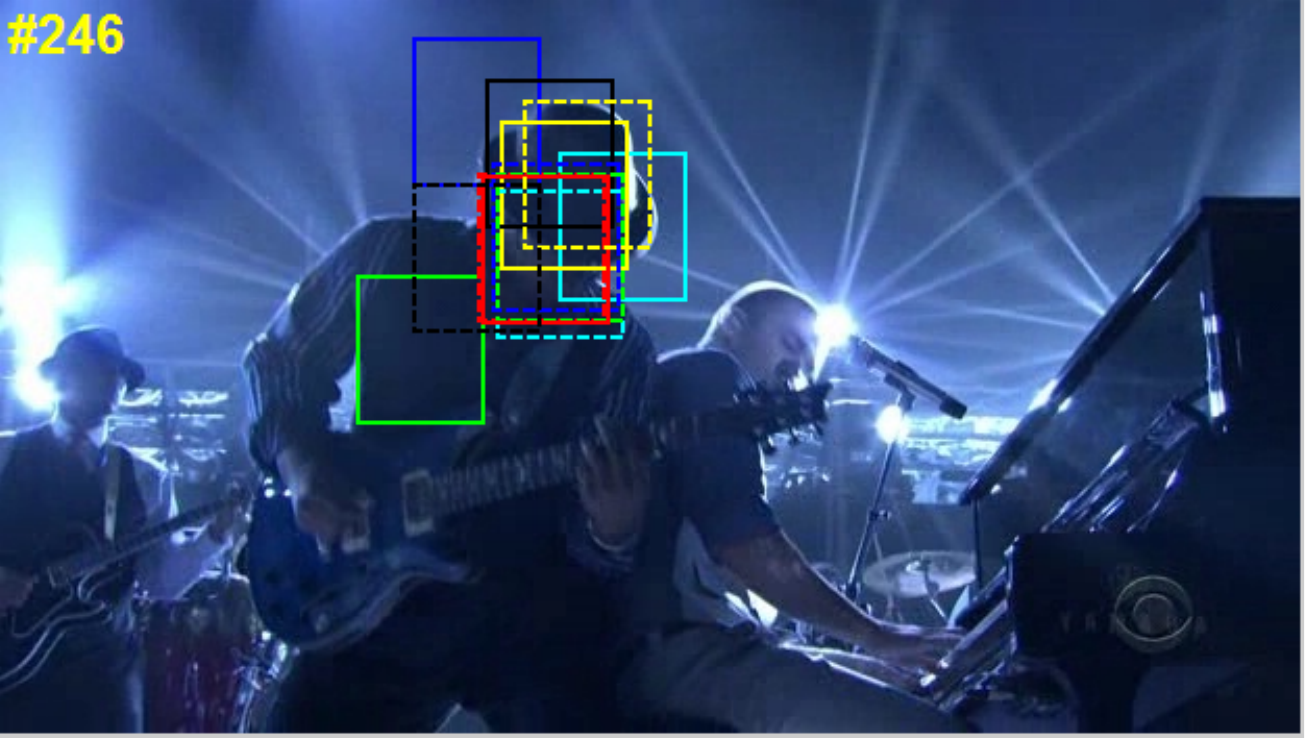}\end{subfigure} &
			\begin{subfigure}{0.2\textwidth}\centering\includegraphics[height=2.8cm, width=2.8cm]{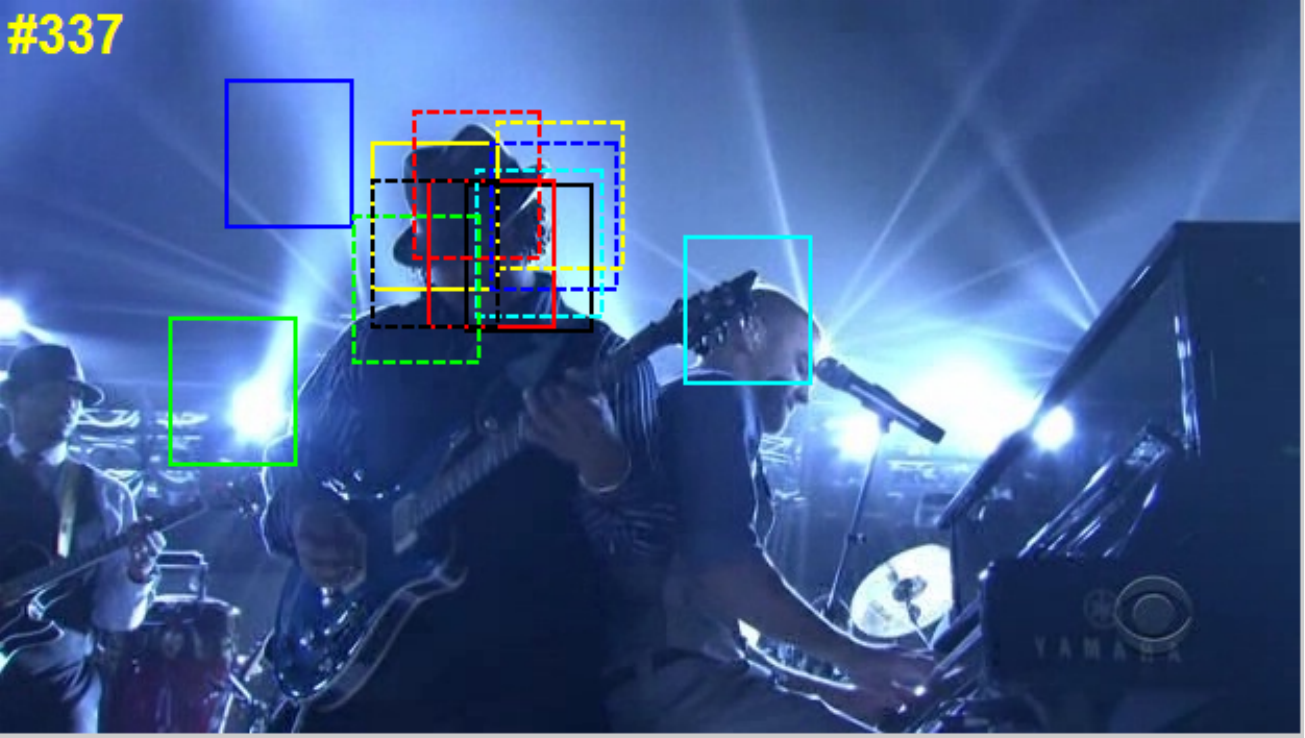}\end{subfigure} \\
			\multicolumn{5}{c}{(f) Shaking} \\
		\end{tabular}
	}
	\begin{subfigure}{1.0\textwidth}\hspace{.15cm}\centering\includegraphics[height=1cm, width=11.5cm]{legend.pdf}\end{subfigure}
	\caption{
		\label{output2}
		Screenshots of some sample tracking results, from left to right and top to bottom. For clarity, we only draw the tracking results of \textcolor{blue}{12} high performing trackers.} 
\end{figure*}

\begin{figure*}
	\resizebox{.987\textwidth}{!}{
		\begin{tabular}{ccccc}
			\begin{subfigure}{0.2\textwidth}\centering\includegraphics[height=2.8cm, width=2.8cm]{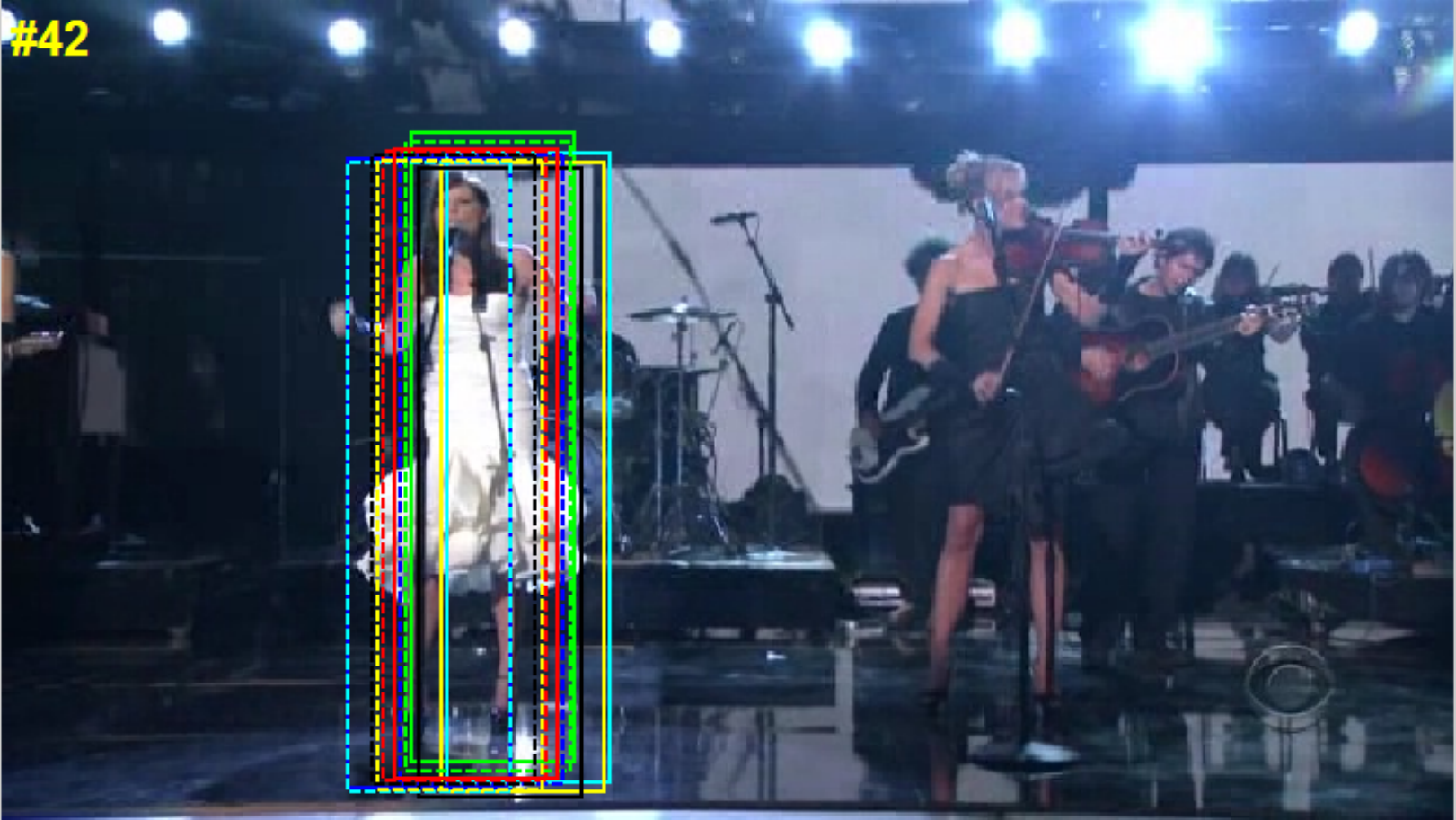}\end{subfigure}&
			\begin{subfigure}{0.2\textwidth}\centering\includegraphics[height=2.8cm, width=2.8cm]{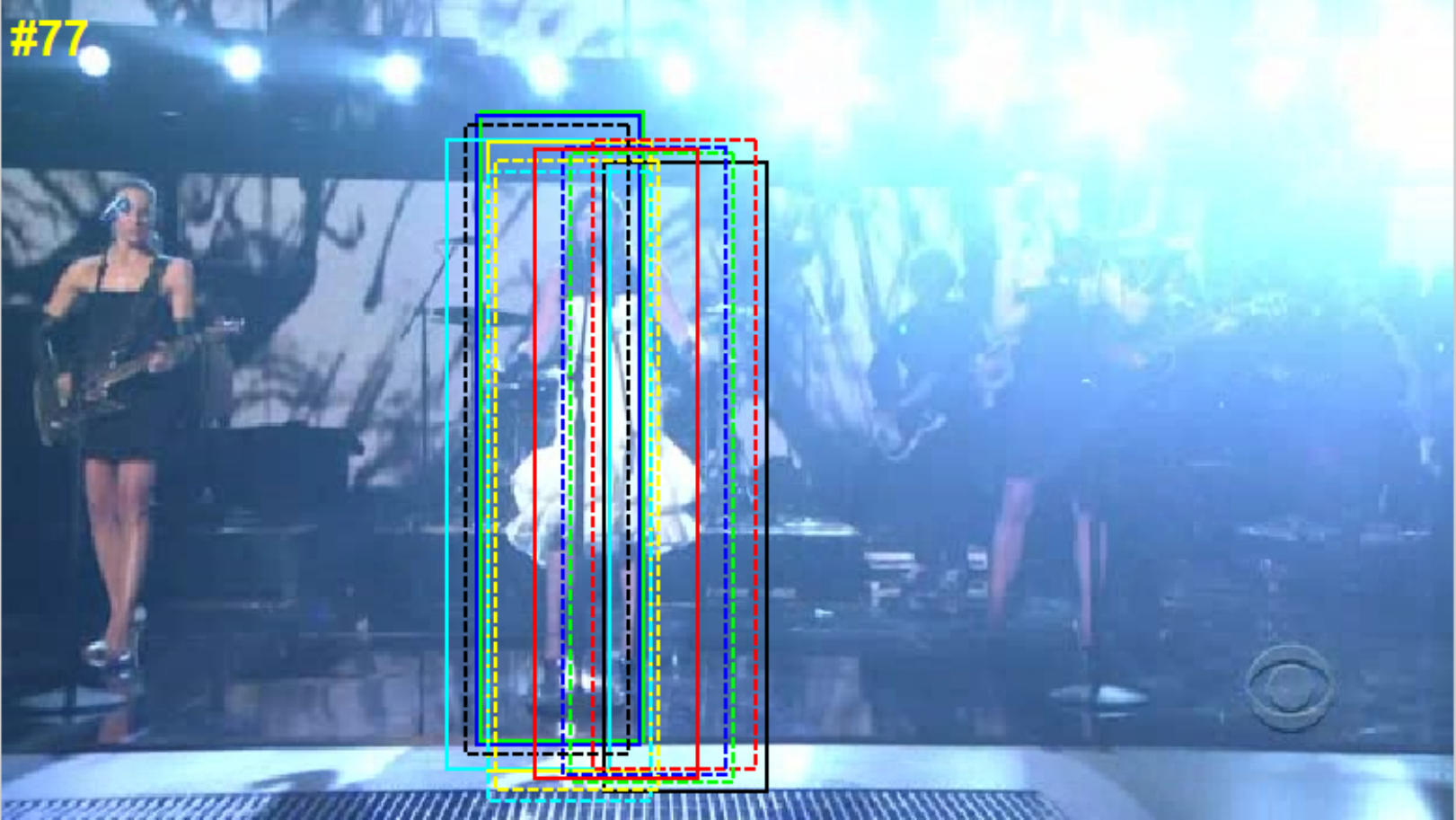}\end{subfigure}&
			\begin{subfigure}{0.2\textwidth}\centering\includegraphics[height=2.8cm, width=2.8cm]{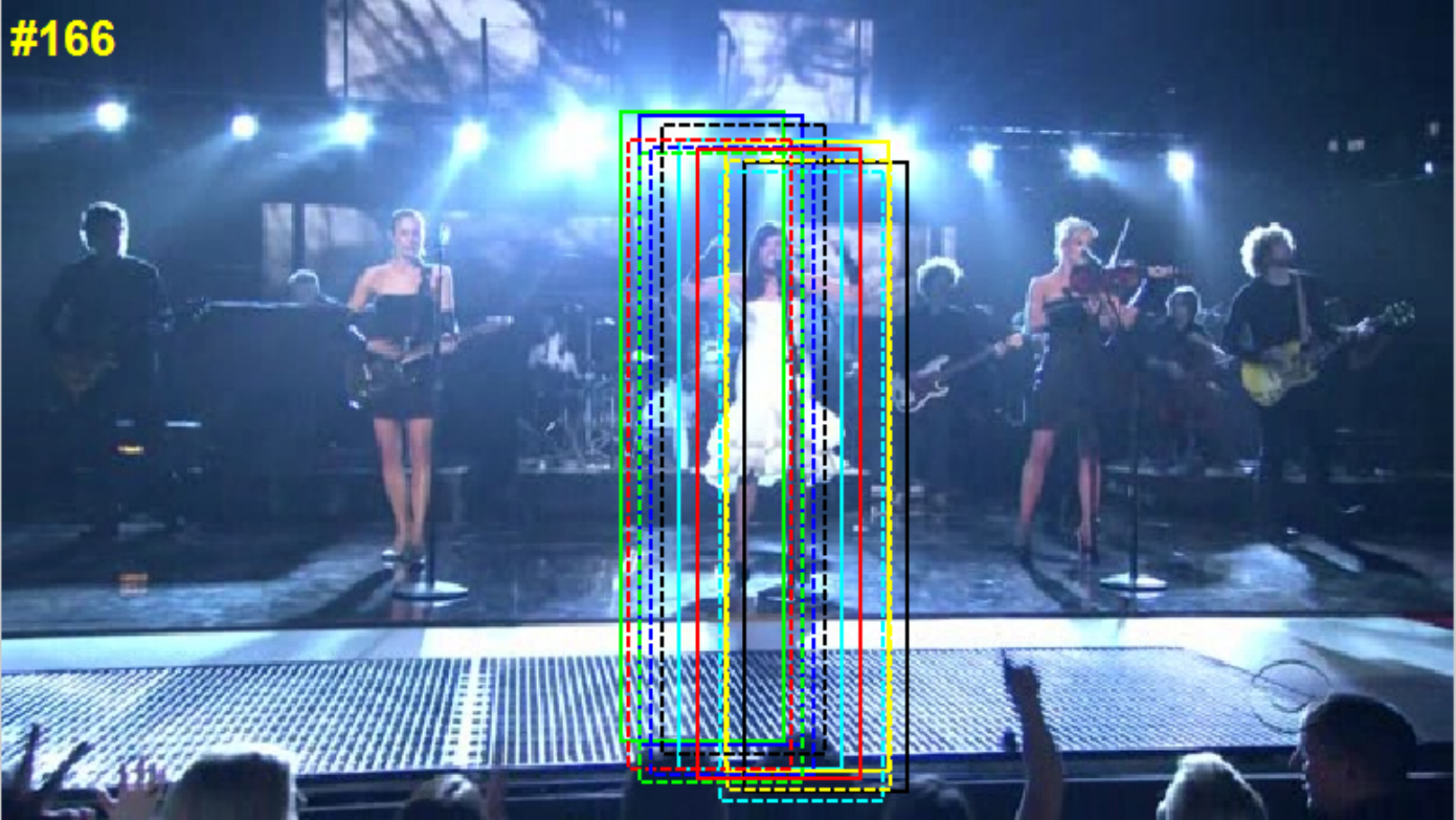}\end{subfigure} &
			\begin{subfigure}{0.2\textwidth}\centering\includegraphics[height=2.8cm, width=2.8cm]{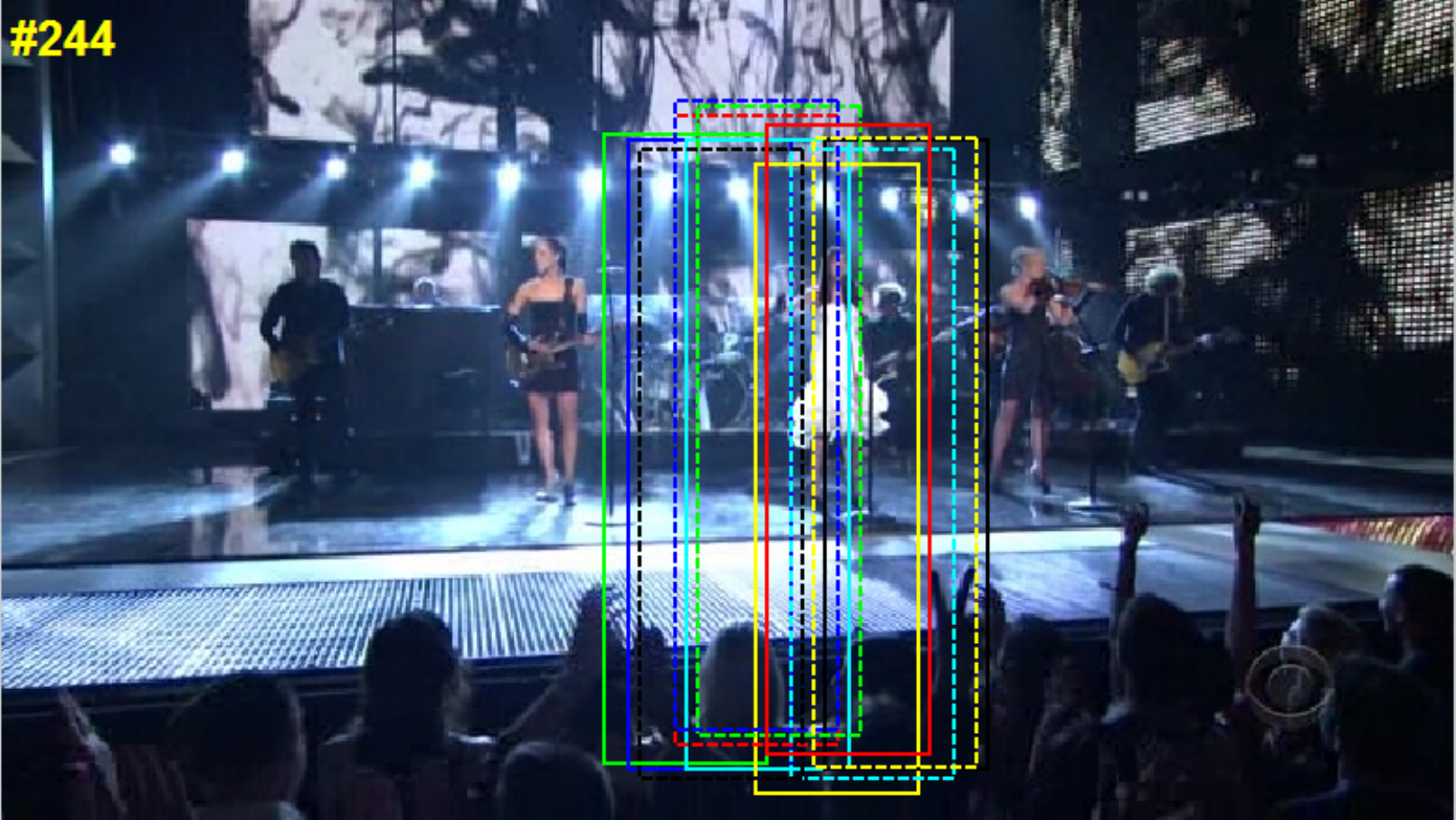}\end{subfigure} &
			\begin{subfigure}{0.2\textwidth}\centering\includegraphics[height=2.8cm, width=2.8cm]{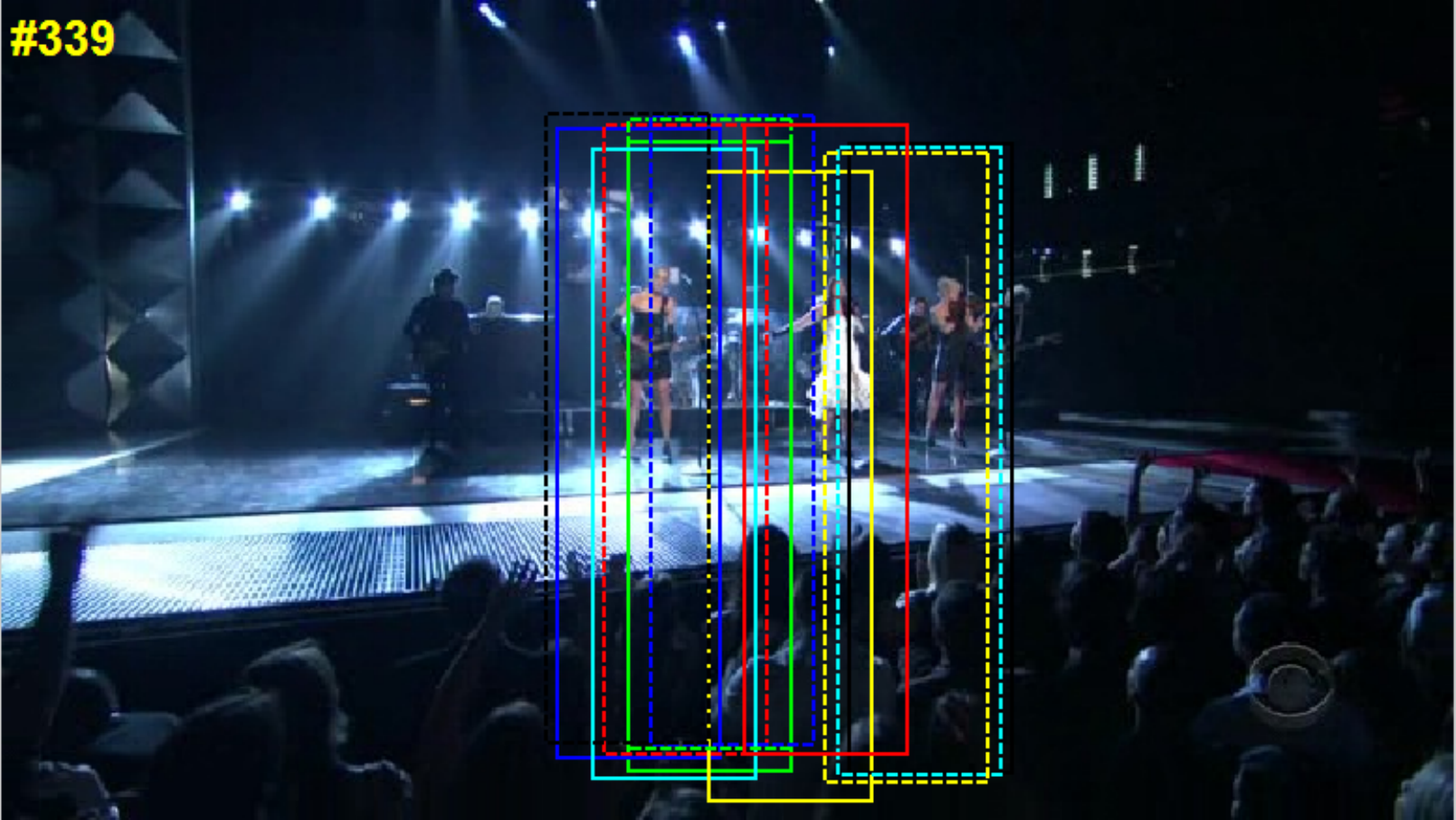}\end{subfigure} \\
			\multicolumn{5}{c}{(a) \textcolor{blue}{Singer1}} \\
			\begin{subfigure}{0.2\textwidth}\centering\includegraphics[height=2.8cm, width=2.8cm]{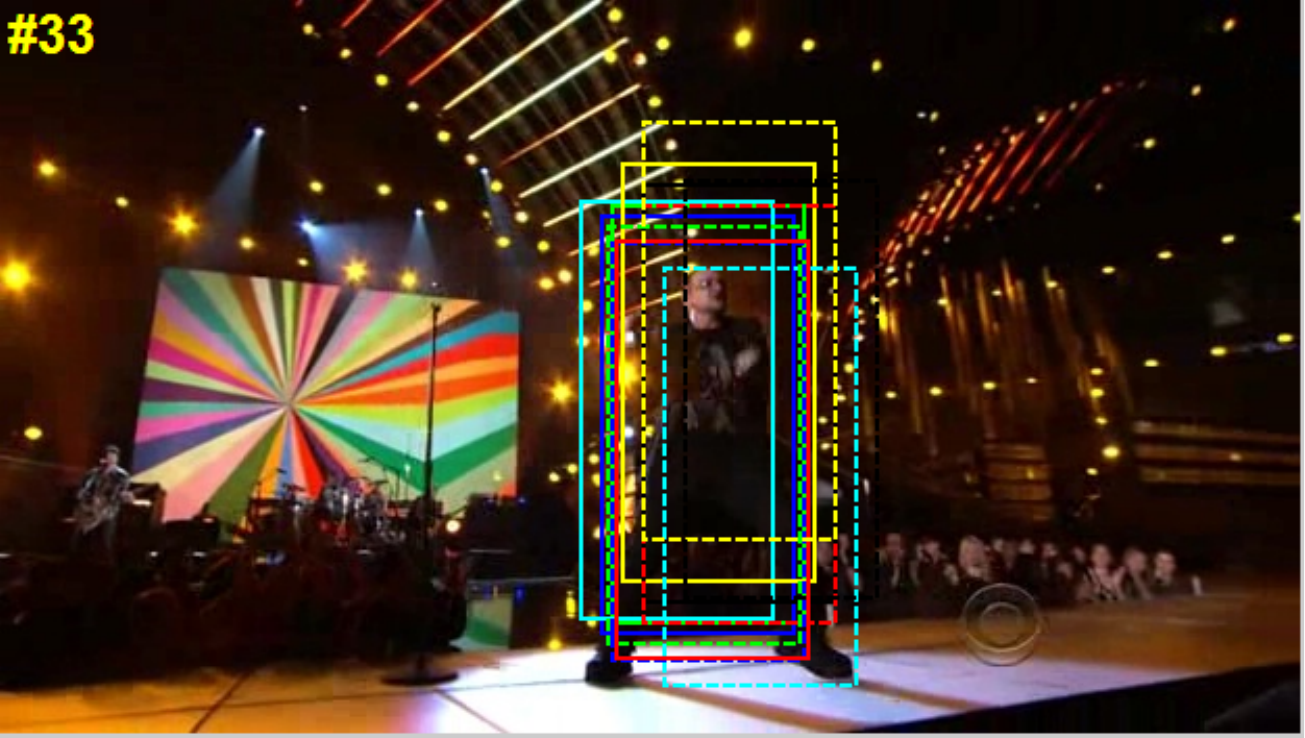}\end{subfigure}&
			\begin{subfigure}{0.2\textwidth}\centering\includegraphics[height=2.8cm, width=2.8cm]{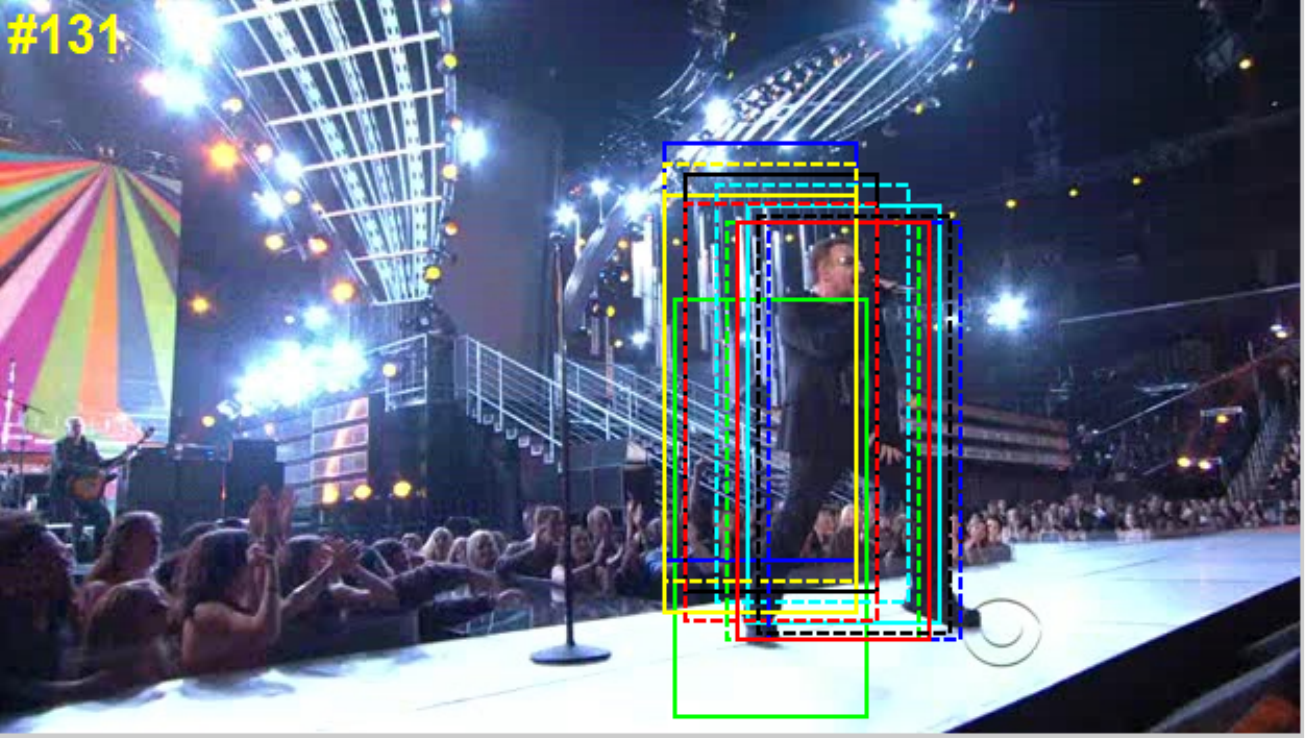}\end{subfigure}&
			\begin{subfigure}{0.2\textwidth}\centering\includegraphics[height=2.8cm, width=2.8cm]{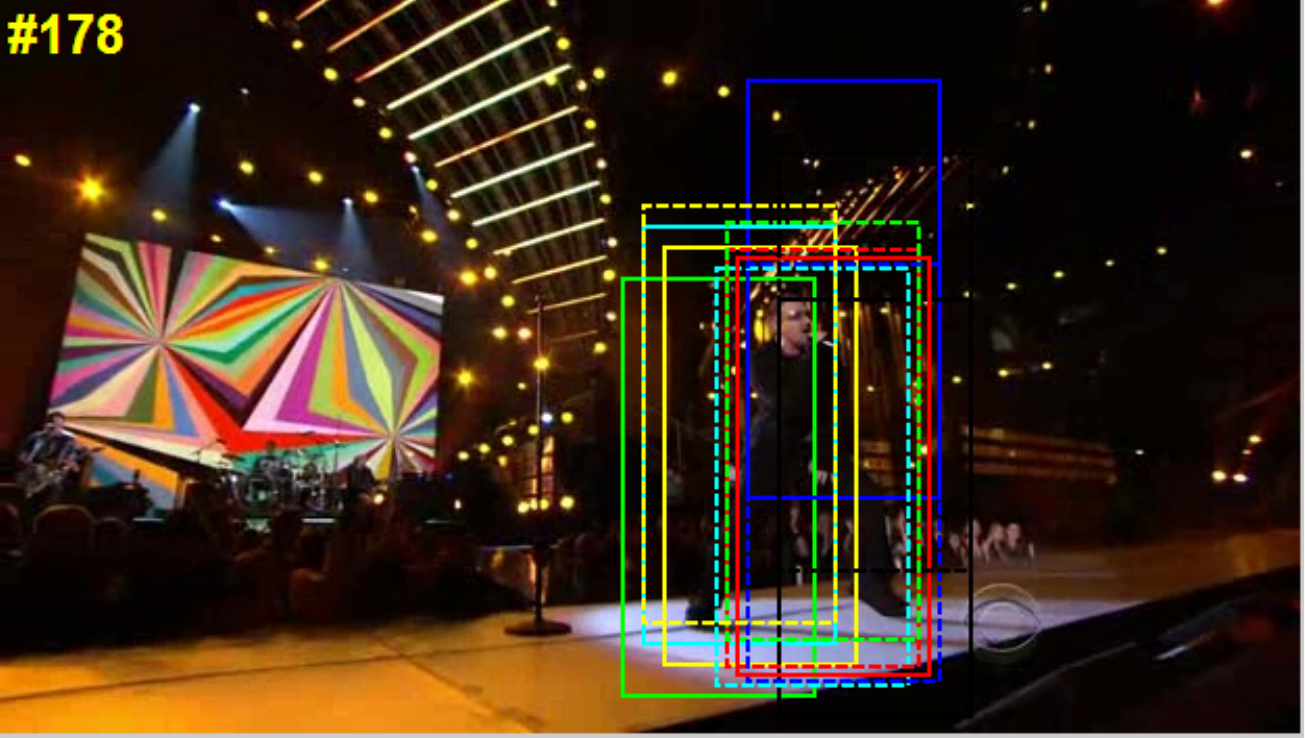}\end{subfigure} &
			\begin{subfigure}{0.2\textwidth}\centering\includegraphics[height=2.8cm, width=2.8cm]{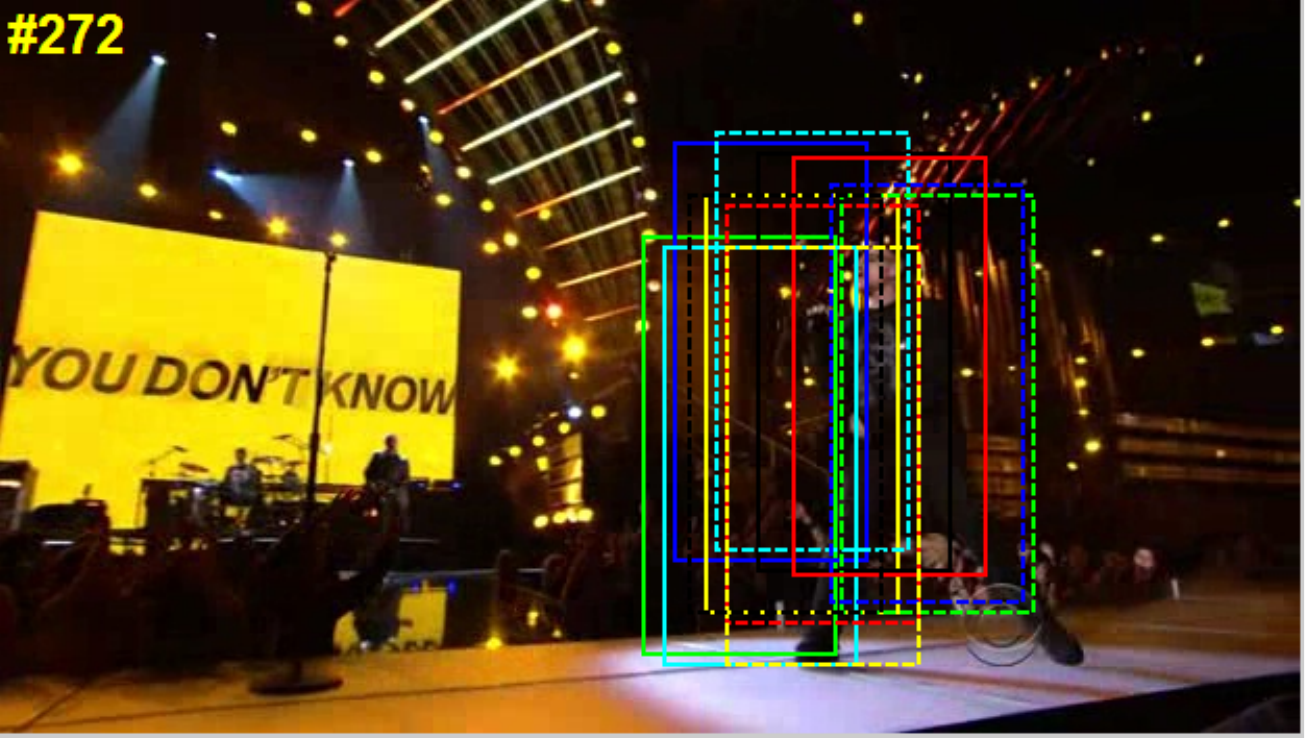}\end{subfigure} &
			\begin{subfigure}{0.2\textwidth}\centering\includegraphics[height=2.8cm, width=2.8cm]{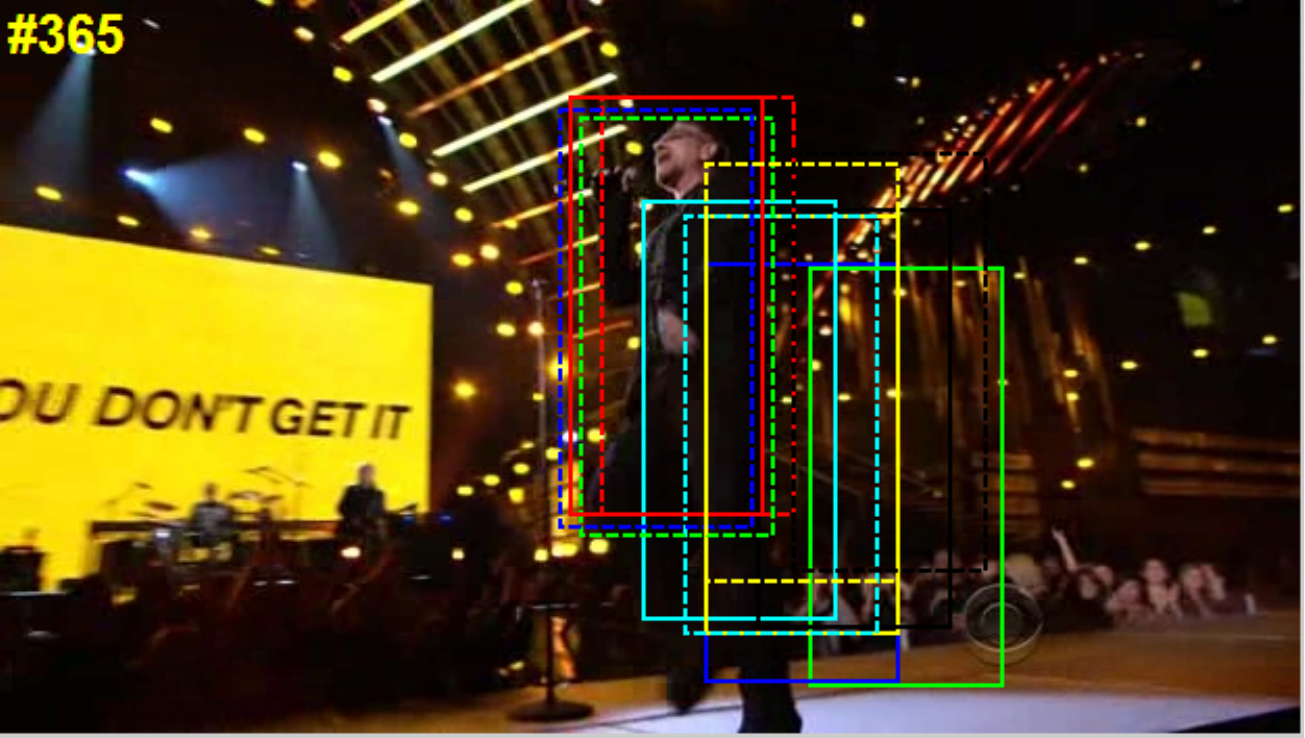}\end{subfigure} \\
			\multicolumn{5}{c}{(b) Singer2} \\
			\begin{subfigure}{0.2\textwidth}\centering\includegraphics[height=2.8cm, width=2.8cm]{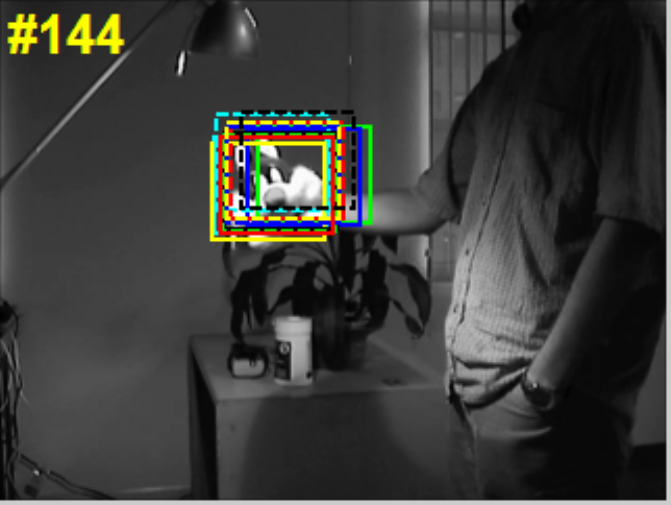}\end{subfigure}&
			\begin{subfigure}{0.2\textwidth}\centering\includegraphics[height=2.8cm, width=2.8cm]{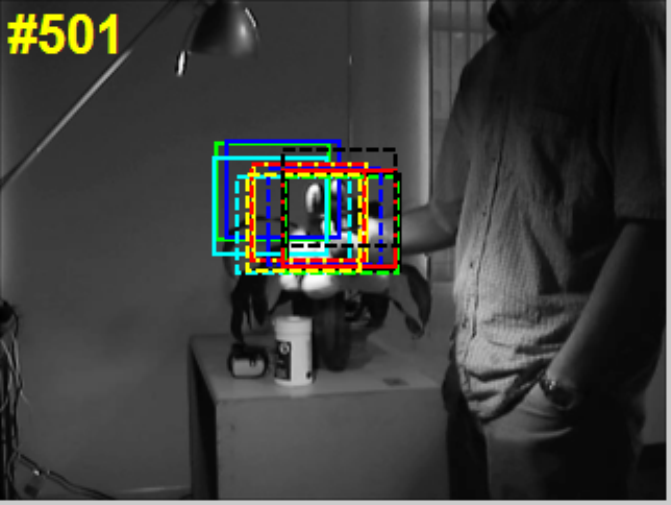}\end{subfigure}&
			\begin{subfigure}{0.2\textwidth}\centering\includegraphics[height=2.8cm, width=2.8cm]{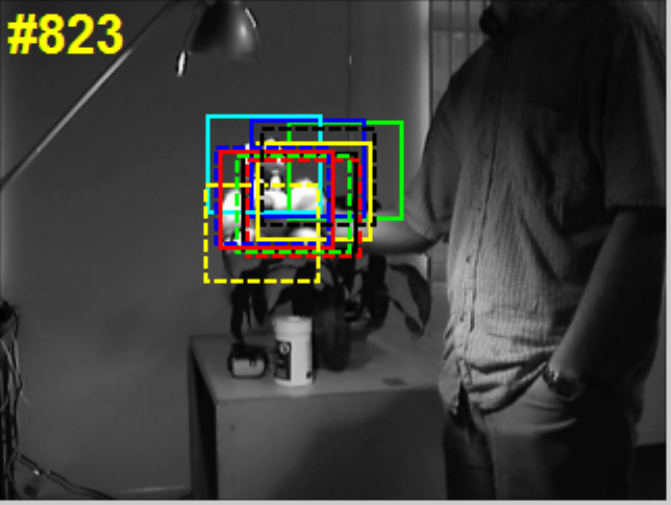}\end{subfigure} &
			\begin{subfigure}{0.2\textwidth}\centering\includegraphics[height=2.8cm, width=2.8cm]{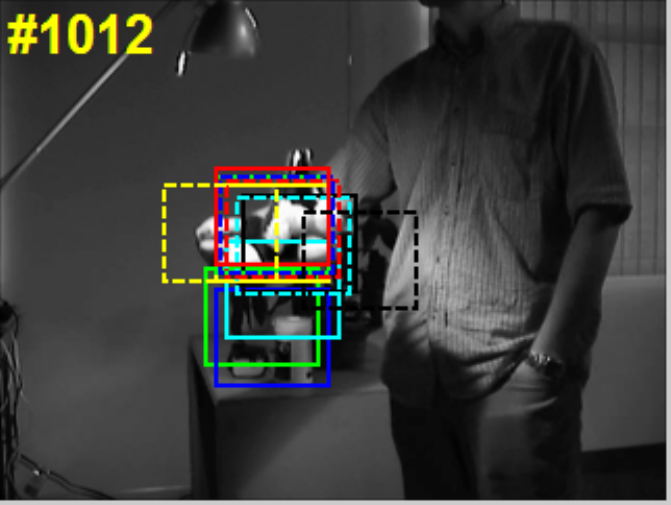}\end{subfigure} &
			\begin{subfigure}{0.2\textwidth}\centering\includegraphics[height=2.8cm, width=2.8cm]{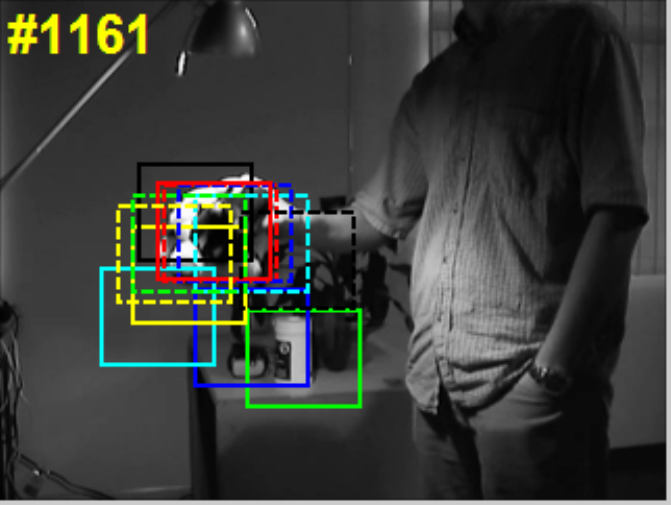}\end{subfigure} 	\\
			\multicolumn{5}{c}{(c) Sylvester } \\
			\begin{subfigure}{0.2\textwidth}\centering\includegraphics[height=2.8cm, width=2.8cm]{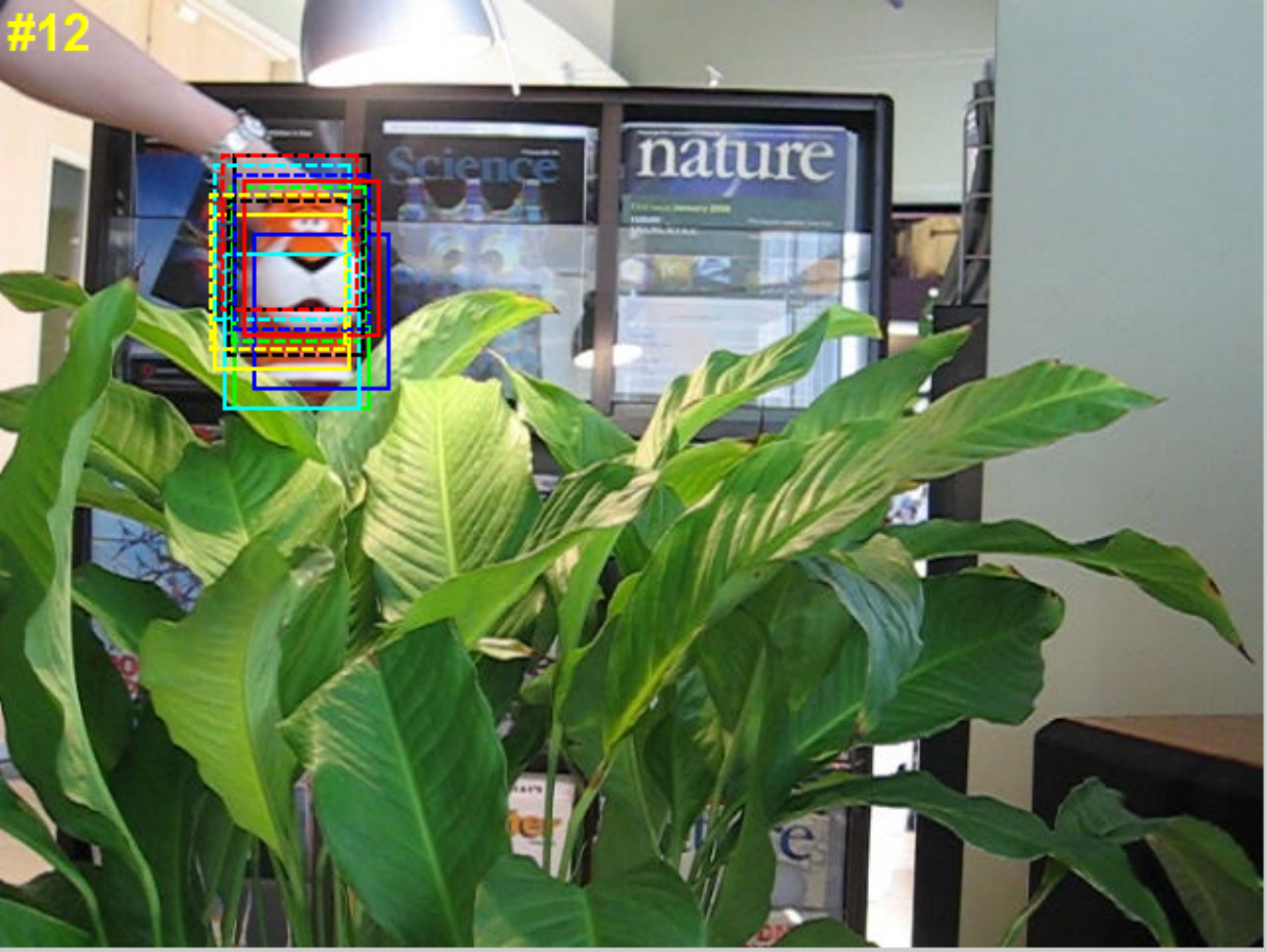}\end{subfigure}&
			\begin{subfigure}{0.2\textwidth}\centering\includegraphics[height=2.8cm, width=2.8cm]{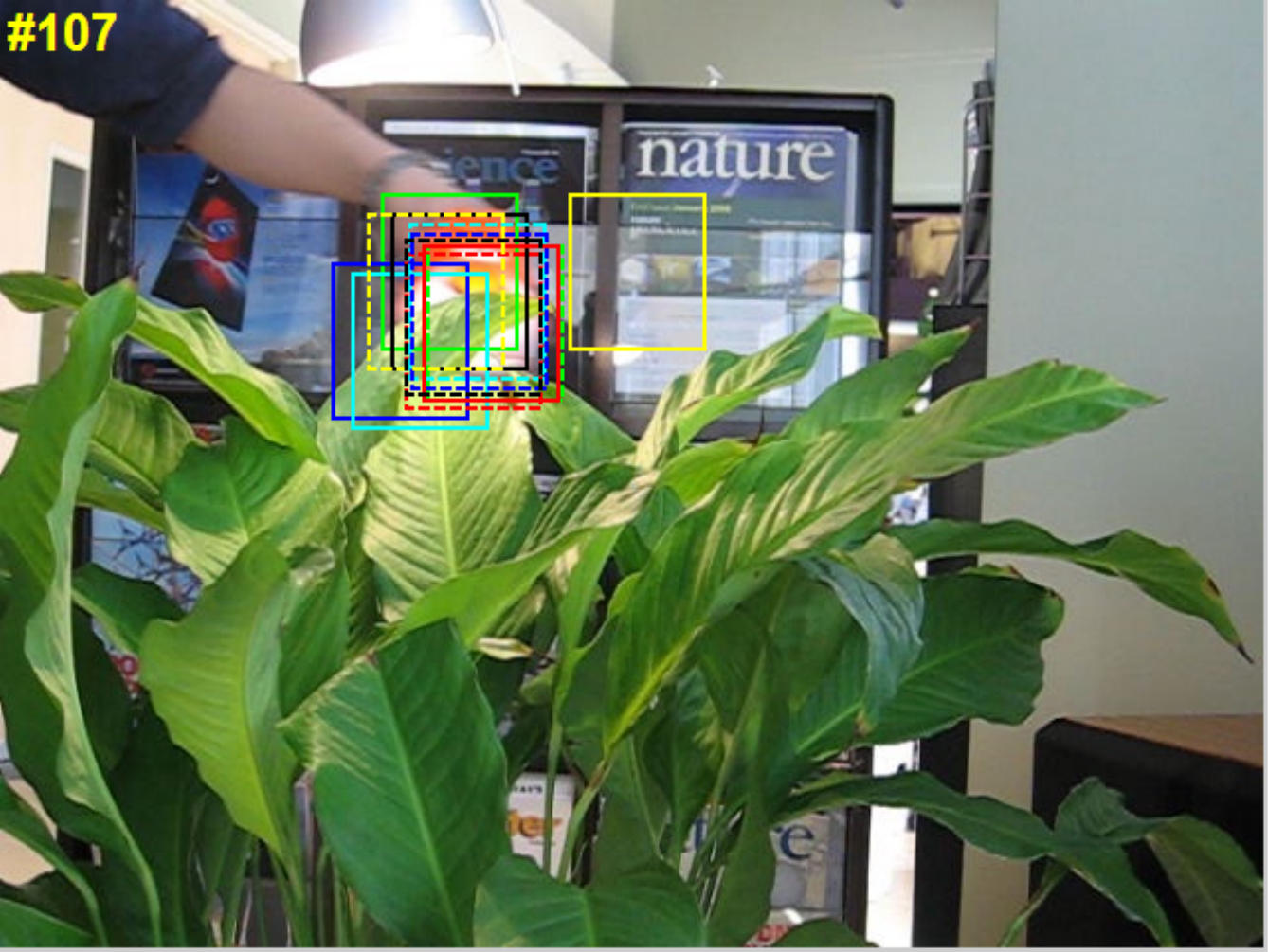}\end{subfigure}&
			\begin{subfigure}{0.2\textwidth}\centering\includegraphics[height=2.8cm, width=2.8cm]{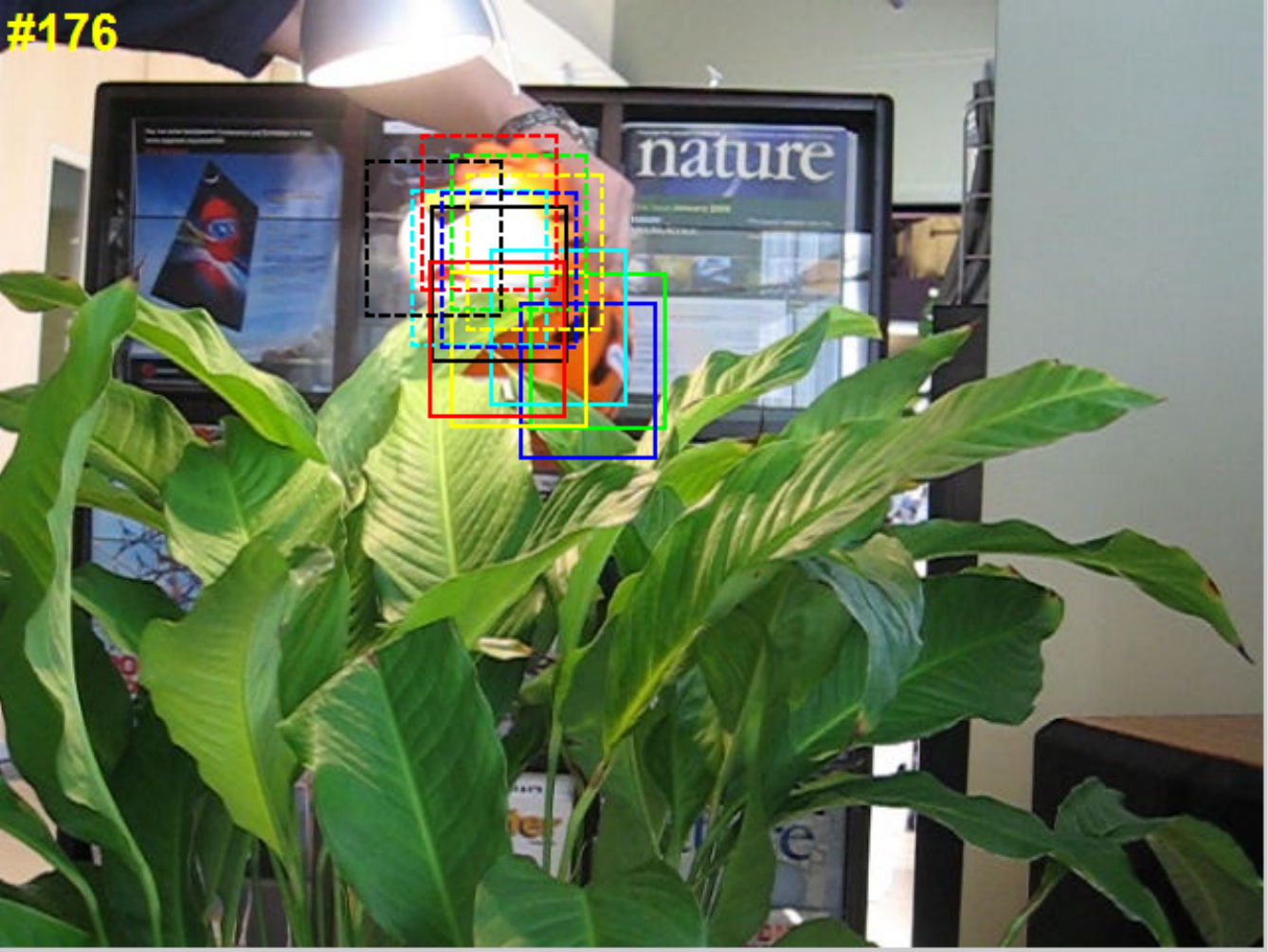}\end{subfigure} &
			\begin{subfigure}{0.2\textwidth}\centering\includegraphics[height=2.8cm, width=2.8cm]{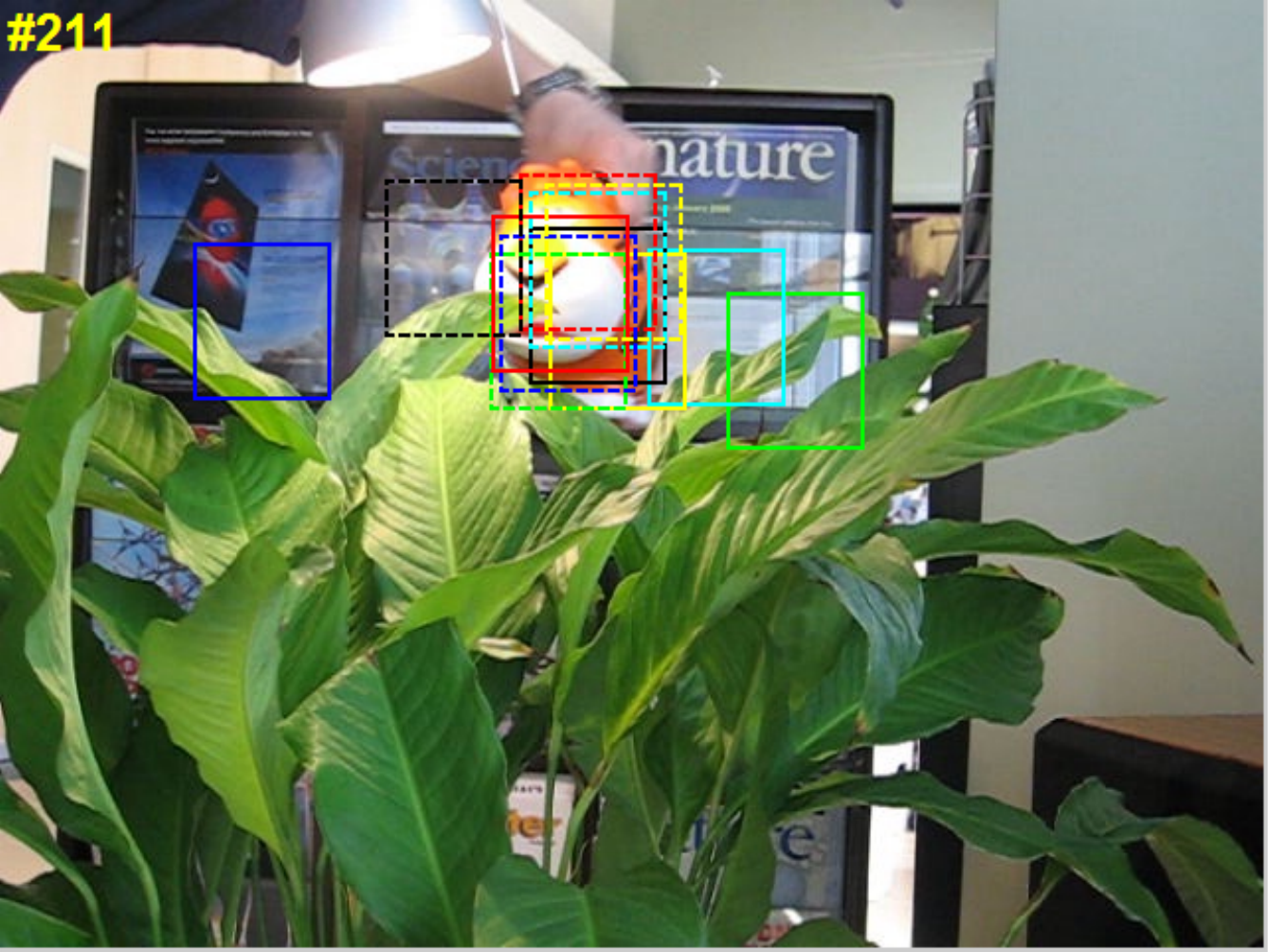}\end{subfigure} &
			\begin{subfigure}{0.2\textwidth}\centering\includegraphics[height=2.8cm, width=2.8cm]{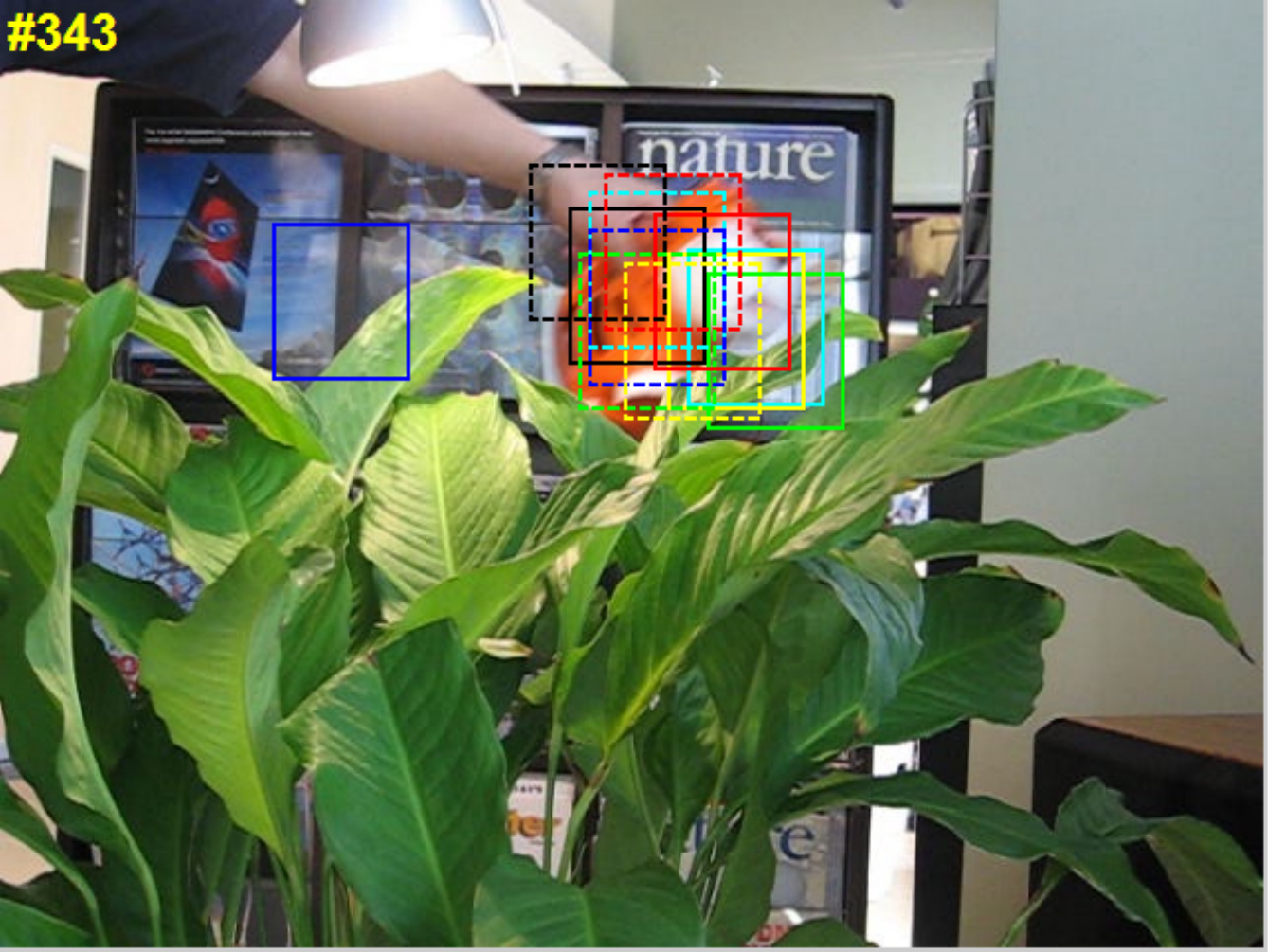}\end{subfigure} \\
			\multicolumn{5}{c}{(d) Tiger2} \\
			\begin{subfigure}{0.2\textwidth}\centering\includegraphics[height=2.8cm, width=2.8cm]{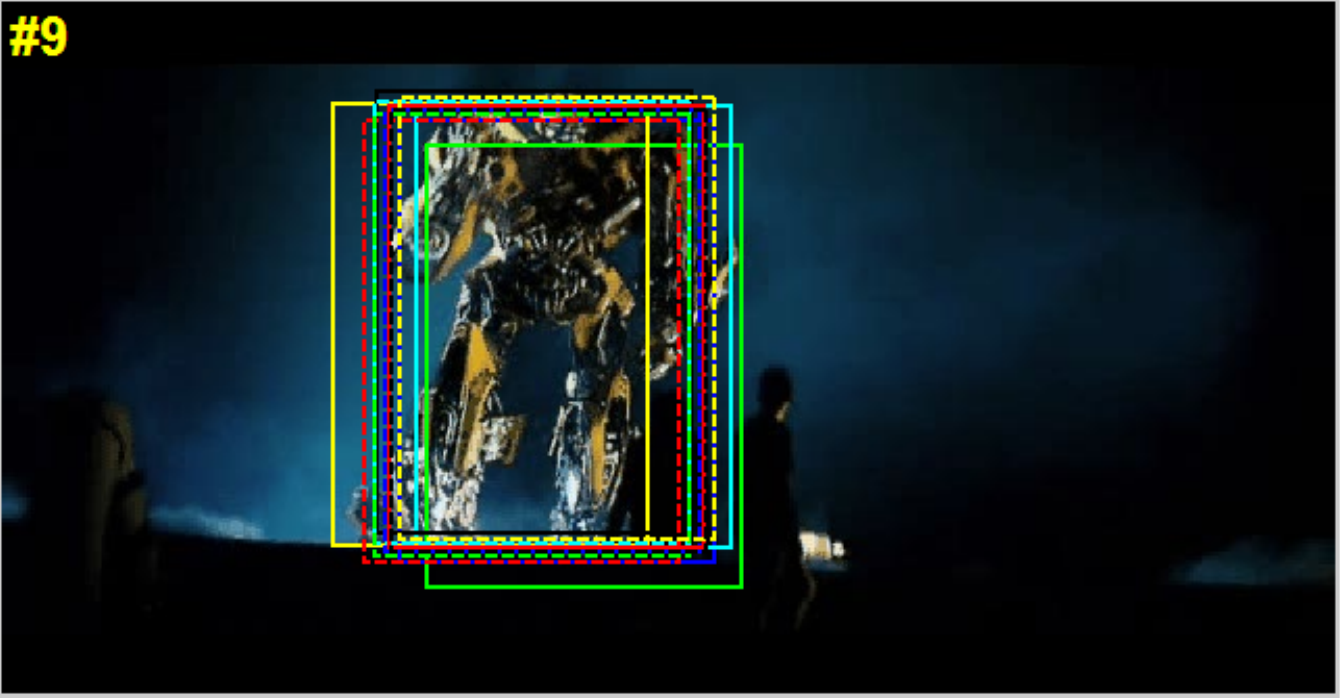}\end{subfigure}&
			\begin{subfigure}{0.2\textwidth}\centering\includegraphics[height=2.8cm, width=2.8cm]{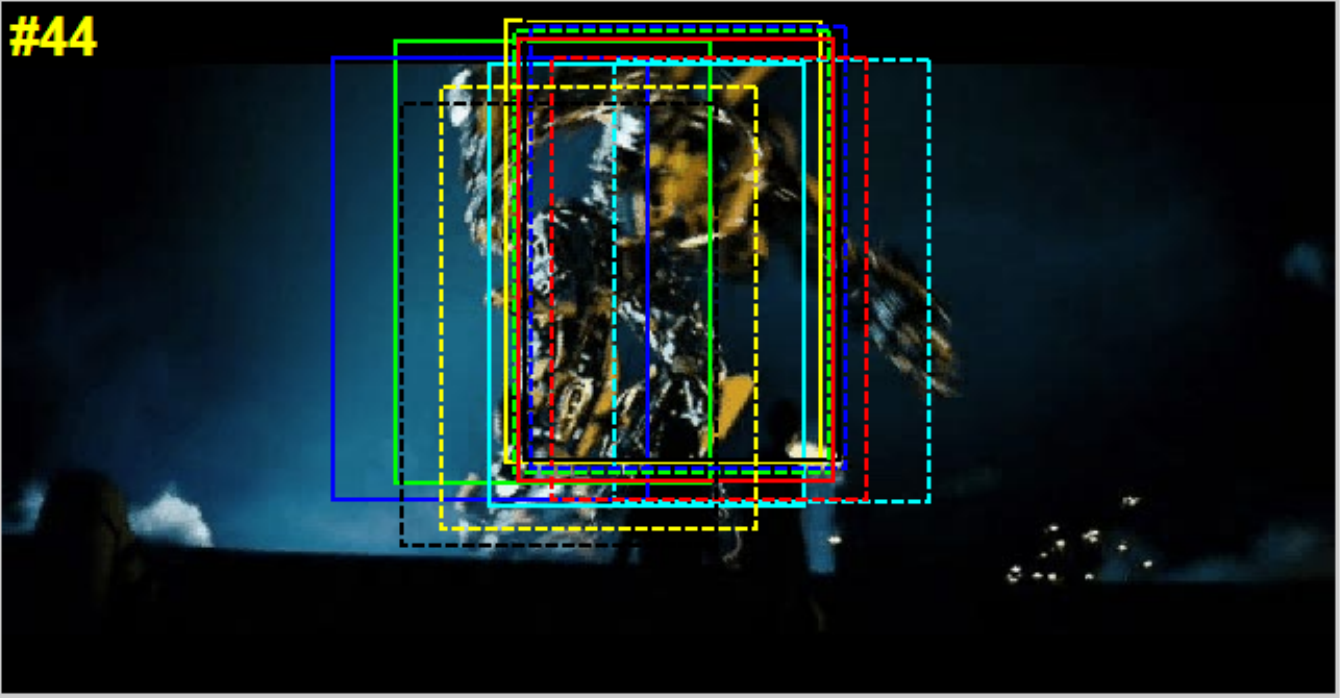}\end{subfigure}&
			\begin{subfigure}{0.2\textwidth}\centering\includegraphics[height=2.8cm, width=2.8cm]{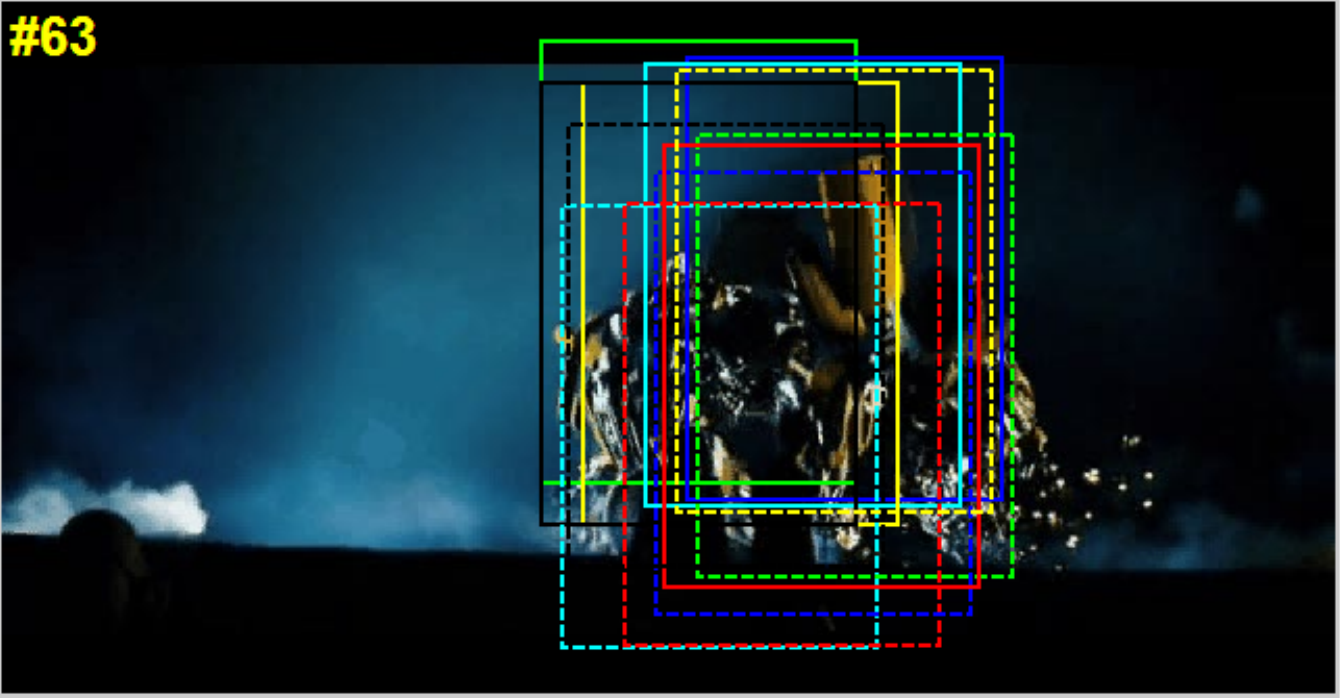}\end{subfigure} &
			\begin{subfigure}{0.2\textwidth}\centering\includegraphics[height=2.8cm, width=2.8cm]{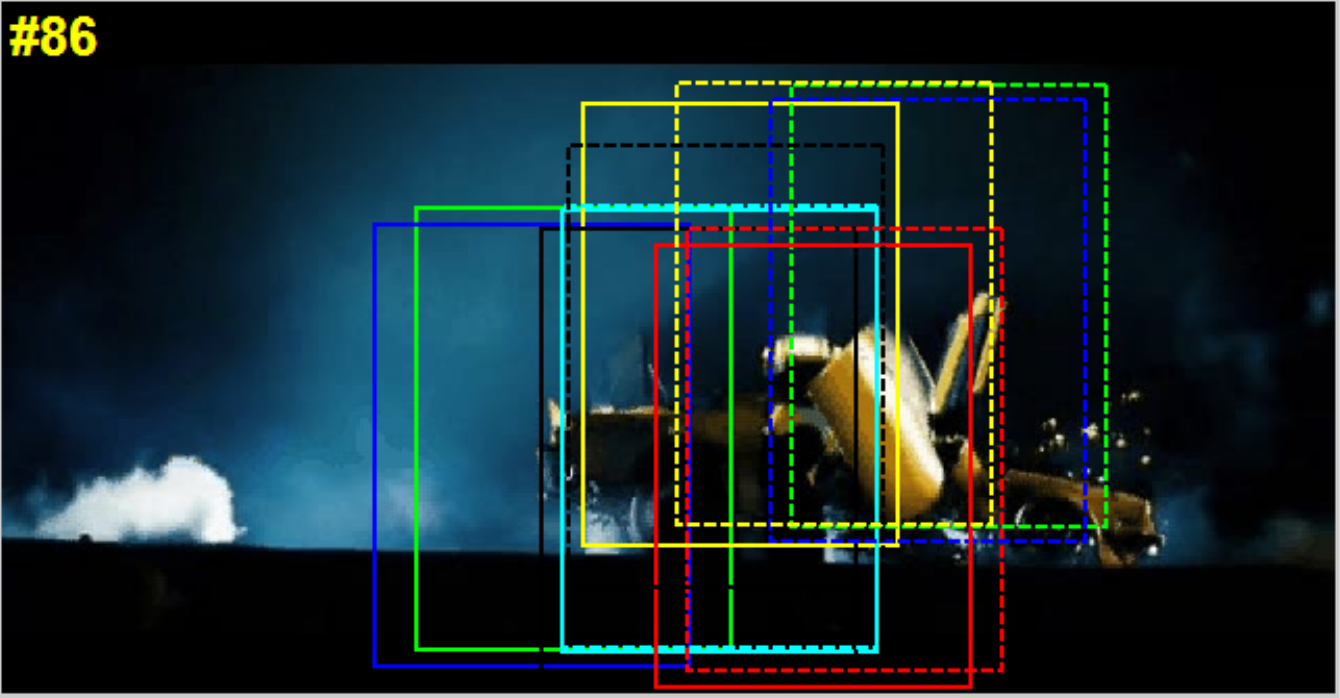}\end{subfigure} &
			\begin{subfigure}{0.2\textwidth}\centering\includegraphics[height=2.8cm, width=2.8cm]{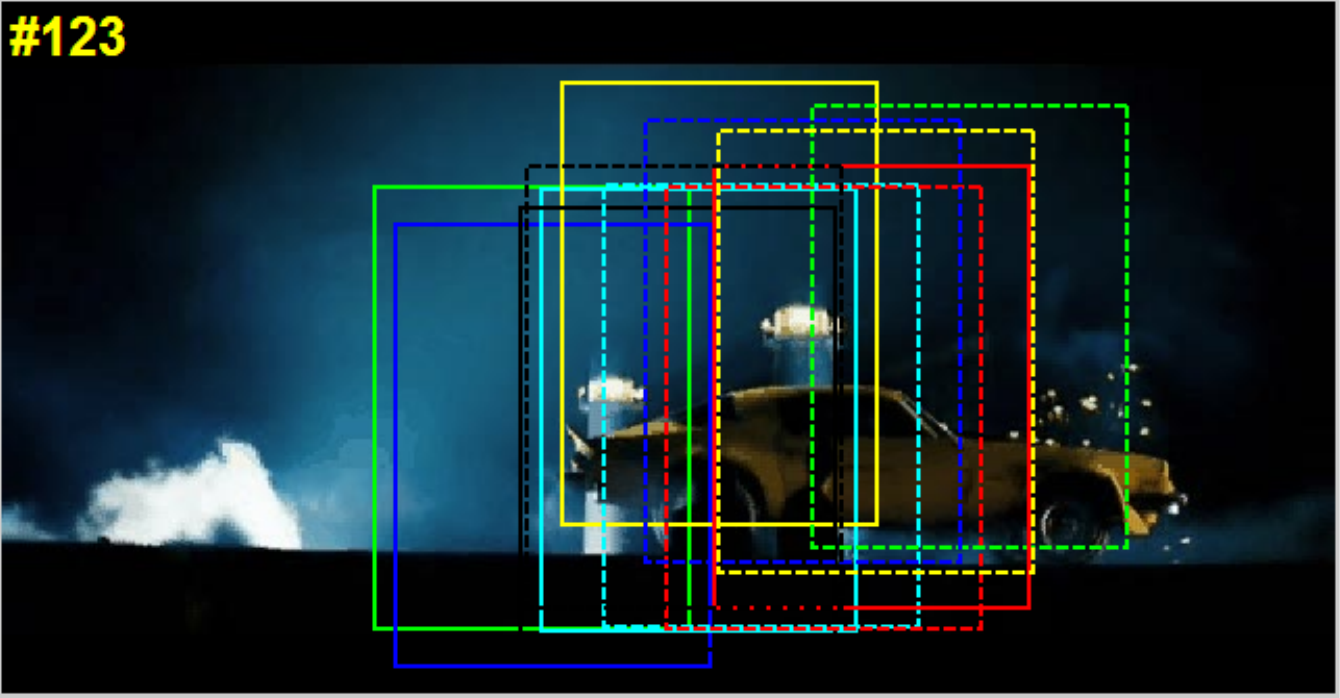}\end{subfigure} \\
			\multicolumn{5}{c}{(e) Trans} \\
			\begin{subfigure}{0.2\textwidth}\centering\includegraphics[height=2.8cm, width=2.8cm]{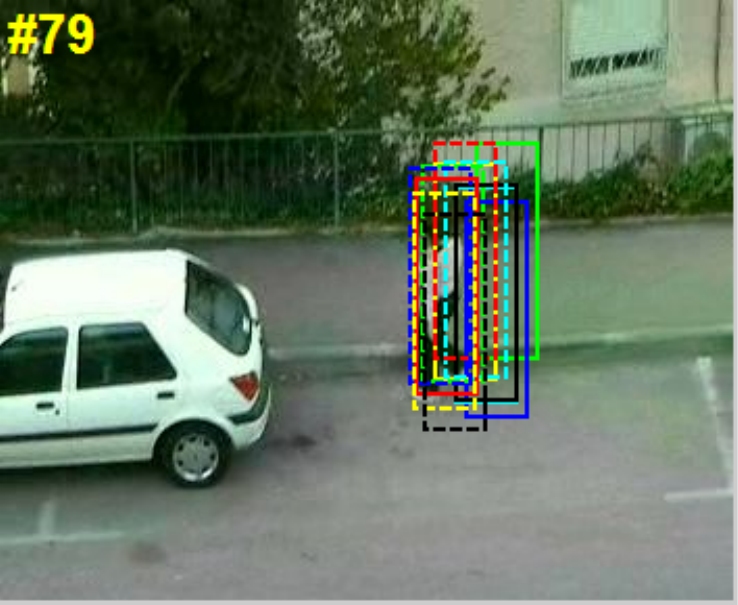}\end{subfigure}&
			\begin{subfigure}{0.2\textwidth}\centering\includegraphics[height=2.8cm, width=2.8cm]{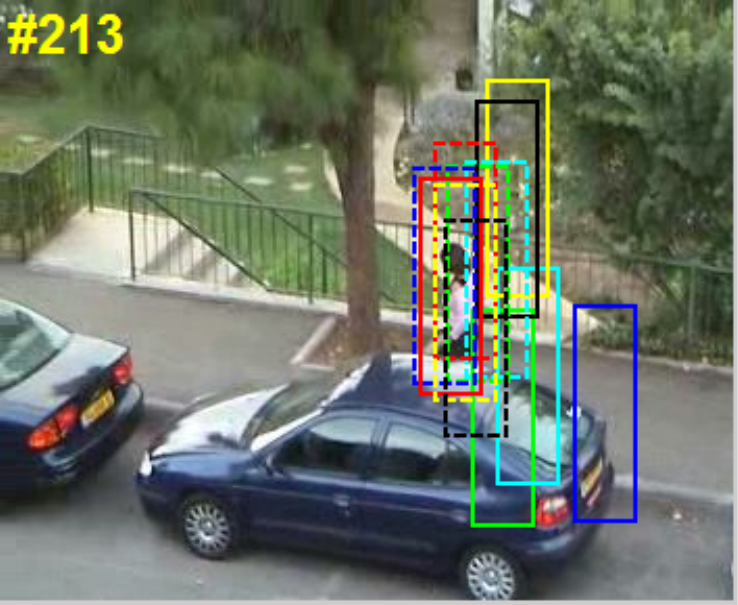}\end{subfigure}&
			\begin{subfigure}{0.2\textwidth}\centering\includegraphics[height=2.8cm, width=2.8cm]{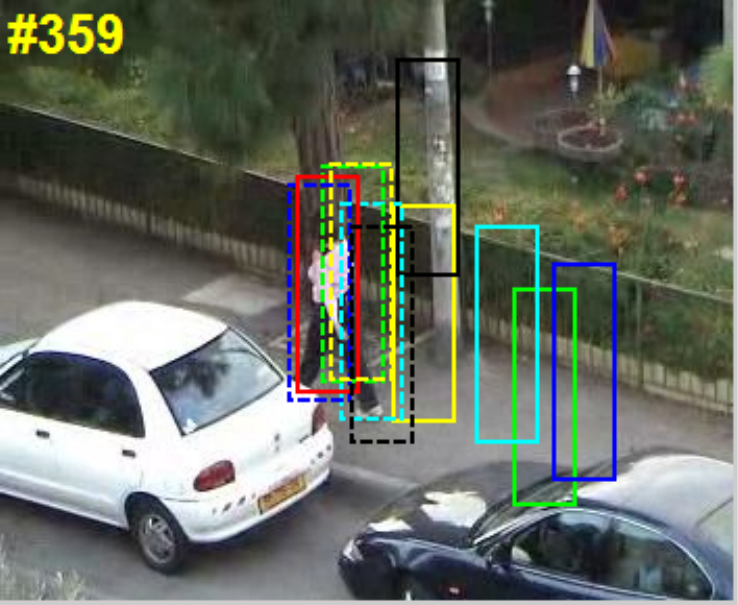}\end{subfigure} &
			\begin{subfigure}{0.2\textwidth}\centering\includegraphics[height=2.8cm, width=2.8cm]{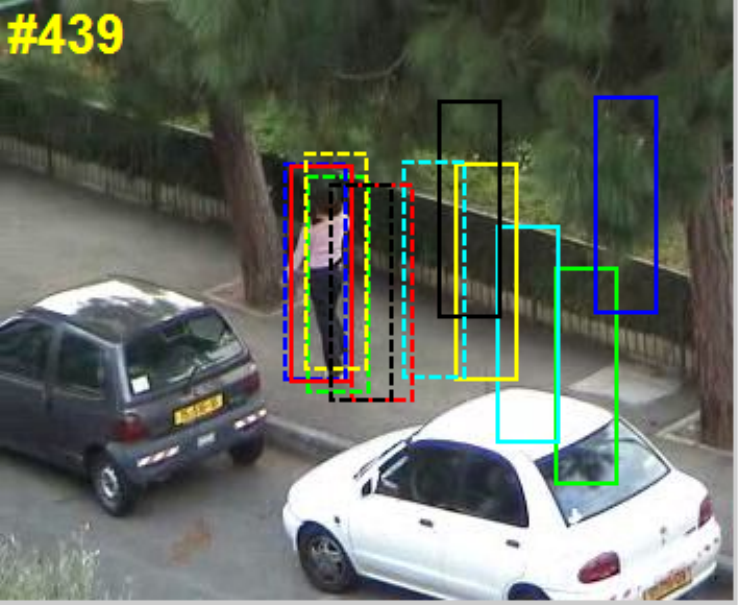}\end{subfigure} &
			\begin{subfigure}{0.2\textwidth}\centering\includegraphics[height=2.8cm, width=2.8cm]{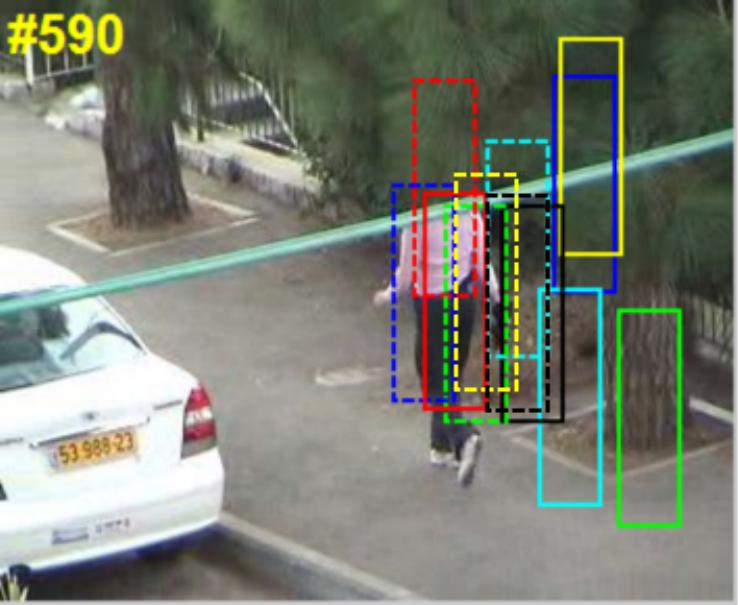}\end{subfigure} \\
			\multicolumn{5}{c}{(f) Woman}
		\end{tabular}
	}

	\begin{subfigure}{1.0\textwidth}\hspace{.15cm}\centering\includegraphics[height=1cm, width=11.5cm]{legend.pdf}\end{subfigure}
	\caption{
		\label{output3}
		Screenshots of some sample tracking results, from left to right and top to bottom. For clarity, we only draw the tracking results of \textcolor{blue}{12} high performing trackers.} 
\end{figure*}

For clearly visible bounding boxes, we show the sample tracking results of only top \textcolor{blue}{twelve} performing approaches namely  OSIL, DWCM, Struck, FCT, DCT, VTD, WMIL, CT, TLD, \textcolor{blue}{KCF, MEEM} and the proposed tracker for qualitative comparison (as displayed in Figs.~\ref{output1}--\ref{output3}). In this section, we discuss the tracking results of \textcolor{blue}{some of the randomly  chosen  tested videos of OTB100 and VOT2015 datasets} based on the different challenging attributes. 
\par \textbf{Background clutter:} The texture or color information of the object in the \textcolor{blue}{Motocross1,} Mountain-bike, Shaking, Singer2 and Football sequences is very similar to the background (Figs.~\ref{output1}--\ref{output3}). As the VTD, L1T, IVT methods employ generative appearance model that do not utilize the background information, it is not easy to accurately track the target object. Due to \textcolor{blue}{the influence} of the surrounding background, the WMIL, CT and TLD trackers suffer from drifting in all the aforementioned sequences. \textcolor{blue}{In the Motocross1 sequence, the appearance of the target object change  (see the frames \textcolor{blue}{\#19} and \textcolor{blue}{\#51} in \textcolor{blue}{Fig.~\ref{output1}(a))}, due to this the VTD, FCT, and KCF trackers are distracted to accurately track the object (see the frame \#87, \#124 and \#158 in Fig.~\ref{output1}(a)).} The \textcolor{blue}{MEEM,} Struck, DWCM, OSIL methods can keep tracking the object, but our approach is \textcolor{blue}{comparatively} more accurate than these.  
\par In the Shaking, Football and Singer2 sequences, the \textcolor{blue}{KCF,} Struck, DCT, FCT and VTD tracker start drifting the bounding box from the frame \#121 in Shaking, \textcolor{blue}{\#174} in Football and \#131 in Singer2 due to the similar color and texture with the background. The \textcolor{blue}{MEEM}, OSIL and DWCM algorithms perform well for Shaking sequence, but not better than our method (see \#246 and \#337 in Fig.~\ref{output2}(f)). However, The DWCM, OSIL and our method work well for Singer2 sequence (see in Fig.~\ref{output3}(b)). Due to the background clutter, the DCT, VTD, WMIL, CT and TLD trackers drift away from the target object after frame \#32 in the Mountain-bike sequence (see Fig.~\ref{output2}(c)). Furthermore, our method tracks the object more accurately than the \textcolor{blue}{MEEM,} OSIL, DWCM and Struck approach in these sequences also. 
\par \textbf{Deformation:}  In the Singer2, Tiger2 and Trans sequences, the tracking object suffers from large changes as the object moves from their places. Figs.~\ref{output1}(d), \ref{output2}(d), \ref{output3}(d), \ref{output3}(f)  show that \textcolor{blue}{KCF,} VTD, WMIL, CT and TLD methods drift after suffering from the appearance variance and occlusion in David, Panda,  Tiger2, and Woman Sequences. In the Woman Sequence, the \textcolor{blue}{MEEM,} DWCM, Struck, FCT and DCT methods also drift away from the objects (see \#439 and \#590 in Fig.~\ref{output3}(f)). In the Panda Sequence, all the algorithms except our method are not able to track the target accurately (see \#2197 and \#2582 in Fig.~\ref{output2}(d)). 
\par The Trans sequence suffers from appearance variation, and scale changes when the object moves to transform (see the frame \#63 in Fig.~\ref{output3}(e)), due to this several trackers drift away from the target. However, \textcolor{blue}{MEEM,} DWCM, OSIL, Struck and \textcolor{blue}{the} proposed method work well for this sequence.  In the David sequence, all the trackers except ours and OSIL are not able to accurately track the target in all the frames (see the frame \#345 and \#447 in Fig.~\ref{output1}(d)) and our method outperforms the OSIL approach. Our algorithm can deal with the deformation well due to its selection of Haar-like features from the un-occluded sub-regions and the coarse-to-fine search strategy. 
\par \textbf{Occlusion, fast motion, motion blur, and rotation:}
Figs.~\textcolor{blue}{\ref{output1}(b)}, \ref{output1}(c), \ref{output1}(f), \ref{output2}(a), \textcolor{blue}{\ref{output3}(a)} \ref{output3}(d), \ref{output3}(f) display the performance of trackers when the target object suffers with occlusion. Due to the heavy occlusion, in-plane as well as out-of-plane rotation, and fast motion in the Coke1 sequence, the \textcolor{blue}{KCF, MEEM,} DWCM, FCT, DCT, VTD, WMIL, CT and TLD algorithms drift away from the target (see after frame \#68 in Fig.~\ref{output1}(c)). All the tested trackers except OSIL and ours do not track the target accurately in Occluded face1 and Occluded face2 sequence due to rotation and occlusion (see \#699 in \ref{output1}(f) \#593 and \#736 in \ref{output2}(a)).

\par In the \textcolor{blue}{Butterfly}, Tiger2, and Woman sequences displayed in Fig.~\ref{output1}(b), \ref{output3}(d),\ref{output3}(f), there are partial occlusion (\textcolor{blue}{\#2 of Butterfly,} \#107 and \#176 of tiger2, \#213, \#359 and \#590 of Woman), motion blur (\#107 and \#343 of Tiger2, \#79 of Woman), out-of-plane rotation (\#79 of Woman, and \#343 of Tiger2), and in-plane rotation (\#176 and \#211 of Tiger2), which makes it very difficult for stable results. The DWCM, Struck, \textcolor{blue}{KCF,} FCT, DCT, VTD, WMIL, CT and TLD trackers do not produce good results for Woman sequence as illustrated by \#359 \#439 \#590 in Fig.~\ref{output3}(f) and the \textcolor{blue}{KCF,}, VTD, CT and TLD do not perform well for Tiger2 sequence. While the CT, VTD and TLD techniques fail for most of the frames in the Woman and Tiger2 sequences. Only the OSIL and our approach perform well on these sequences. Due to the partial occlusion from traffic light as well as running woman and motion blur in the Pedestrian3 sequence (Fig.~\ref{output2}(e)), most of the trackers are not able to successfully track the object.  
\par In the David and Panda sequences, the target undergoes occlusion, in-plane rotation and out-of-plane rotation, as shown in Figs.~\ref{output1}(d) and ~\ref{output2}(d), no approach performs well except ours, DWCM and FCT for Panda. \textcolor{blue}{On the other hand MEEM,} DWCM, FCT, OSIL, Struck and our method for David sequence \textcolor{blue}{perform better}. However FCT is fails to track at \#2582 of Panda sequence. 
\par There is in-plane and out-of-plane rotation in the \textcolor{blue}{Motocross1,} Football, Mountain bike, Shaking, Singer2, Sylvester sequences and overall only our tracker performs favorably to deal with these challenges (see Figs.~\ref{output1}-\ref{output3}). 
\par \textbf{Illumination and low resolution:}
In the David, Shaking, \textcolor{blue}{Singer1,} and Trans sequences, the CT, TLD, \textcolor{blue}{KCF} and WMIL tracker drift the bounding box to another place due to the heavy illumination changes in \#51 of Fig.~\ref{output1}(d), \#121 of Fig.~\ref{output2}(f), \textcolor{blue}{\#77 of Fig.~\ref{output3}(a),} and \#44 of Fig.~\ref{output3}(d). In the Shaking, the DCT, VTD, DWCM, \textcolor{blue}{MEEM,}  and our tracker proves to be efficient in dealing with significant illumination changes. The Struck, FCT, DCT, TLD, CT, WMIL, \textcolor{blue}{KCF}, and VTD tracker drift the bounding box to another place in Woman Sequence due to significant illumination changes and partial occlusion (see \#439 and \#590 in Fig.~\ref{output3}(f)). Due to the same reason, the above trackers except our method, Struck, FCT and DCT do not perform well with pedestrian3 sequence (see \#137 in Fig.~\ref{output2}(e)). The TLD, Struck, OSIL, DWCM and our method work comparably well with Coke1 sequence. The Panda sequence also suffers \textcolor{blue}{from} drifting problem with the \textcolor{blue}{KCF}, \textcolor{blue}{MEEM}, FCT, VTD, WMIL, CT and TLD trackers due to the illumination variations and low resolution. In the Football sequence, there is the problem of low resolution and background clutter  which makes it very complicated for robust tracking (see Fig.~\ref{output2}(b)). However VTD, OSIL and our method show reasonably better results for this sequence. There are illumination changes problem in the \textcolor{blue}{Butterfly, Motocross1, Singer1}, Occluded face2, Sylvester  and Tiger2 sequences, but our tracker is reasonably better than others for these sequences too.  
\par In summary, from the above discussion, the presented approach is able to correctly track the targets in all the tested sequences. Our approach outperforms the others because it extracts the discriminative features from the un-occluded regions.  The coarse-to-fine search \textcolor{blue}{strategy} and weight for positive samples mechanism employed in our method \textcolor{blue}{are} able to rectify the drifting problem.

\section{Conclusion}
\label{conclusion}
In this paper, we have proposed a robust visual object tracking algorithm via online weighted multiple instance learning under the coarse-fine-search strategy based sparse representing framework. Here we have considered the spatial information into account by assigning weights to the important features, by which, it can efficiently discriminate the target samples in the different challenging environments. In addition to this, we have extracted the stable Haar-like random rectangular features from the un-occluded sub-regions to develop a strong classifier. Extensive experimental results on different attribute based challenging sequences demonstrate that our tracker outperforms the state-of-the-art algorithms in terms of stability and accuracy. 

\section*{References}
\bibliography{mybibfile}
\end{document}